\documentclass[sigconf]{acmart}
\pdfpageattr {/Group << /S /Transparency /I true /CS /DeviceRGB>>}

\usepackage{amsmath}
\usepackage{amsfonts}
\usepackage{enumitem}
\usepackage{multirow}
\usepackage{booktabs}
\usepackage{caption}
\usepackage{array}
\usepackage{makecell}
\usepackage{color}

\usepackage{subcaption}
\usepackage{algorithm}
\usepackage{algpseudocode}
\usepackage{xcolor}
\usepackage{multirow}

\usepackage{threeparttable}
\usepackage{pifont}
\usepackage{pdfpages}
\usepackage{pdfcomment}
\usepackage{tikz}
\usepackage{pifont}
\usepackage{float}
\usepackage{bm}
\usepackage{listings} 
\usepackage{xcolor} 
\usepackage{minted}
\usepackage{placeins}
\newcommand{\fst}[1]{\textcolor[RGB]{255,0,185}{#1}}
\newcommand{\snd}[1]{\textcolor[RGB]{0,153,76}{#1}}
\newcommand{\trd}[1]{\textcolor[RGB]{0,128,255}{#1}}
\definecolor{deepgreen}{rgb}{0,0.5,0} 
\DeclareMathOperator*{\argmax}{argmax}
\DeclareMathOperator*{\argmin}{argmin}

\algrenewcommand{\algorithmiccomment}[1]{\hfill \textcolor{gray}{\textit{// #1}}}
\usepackage{graphicx} 
\usepackage{multirow} 
\usepackage{booktabs} 
\usepackage{array}    
\usepackage{colortbl} 
\usepackage{xcolor}   
\usepackage{arydshln} 
\AtBeginDocument{%
  \providecommand\BibTeX{{%
    \normalfont B\kern-0.5em{\scshape i\kern-0.25em b}\kern-0.8em\TeX}}}

\setcopyright{acmlicensed}
\copyrightyear{2026}
\acmYear{2026}
\acmConference[KDD '26]{KDD '26: Proceedings of the 32nd ACM SIGKDD Conference on Knowledge Discovery and Data Mining}{August 9-13, 2026}{Jeju, Korea}
\acmBooktitle{Proceedings of the 32nd ACM SIGKDD Conference on Knowledge Discovery and Data Mining, August 9-13, 2026, Jeju, Korea}

\renewcommand\footnotetextcopyrightpermission[1]{} 

\renewcommand\footnotetextcopyrightpermission[1]{} 

\begin{document}
\title{FineFT: Efficient and Risk-Aware Ensemble Reinforcement Learning for Futures Trading}

\author{Molei Qin}
\affiliation{%
  \institution{Nanyang Technological University}
  \country{Singapore}}
\email{molei001@e.ntu.edu.sg}

\author{Xinyu Cai}
\affiliation{%
  \institution{Nanyang Technological University}
  \country{Singapore}}
\email{xinyu009@e.ntu.edu.sg}

\author{Yewen Li}
\affiliation{%
  \institution{Nanyang Technological University}
  \country{Singapore}}
\email{yewen001@e.ntu.edu.sg}

\author{Haochong Xia}
\affiliation{%
  \institution{Nanyang Technological University}
  \country{Singapore}}
\email{haochong001@e.ntu.edu.sg}

\author{Chuqiao Zong}
\affiliation{%
  \institution{Nanyang Technological University}
  \country{Singapore}}
\email{zong0005@e.ntu.edu.sg}

\author{Shuo Sun}
\affiliation{%
  \institution{Hong Kong University of Science and Technology (Guangzhou)}
  \country{China}}
\email{shuosun@hkust-gz.edu.cn}

\author{Xinrun Wang}
\authornote{Corresponding authors.}
\affiliation{%
  \institution{Singapore Management University}
  \country{Singapore}}
\email{xrwang@smu.edu.sg}

\author{Bo An}
\affiliation{%
  \institution{Nanyang Technological University}
  \country{Singapore}}
\email{boan@ntu.edu.sg}

\renewcommand{\shortauthors}{Qin et al.}



\begin{abstract}
Futures are contracts obligating the exchange of an asset at a predetermined date and price, notable for their high leverage (e.g., 5-fold) and liquidity (e.g., trillions of dollars) and, therefore, thrive in the Crypto market.
Reinforcement learning (RL) has been widely applied in various quantitative tasks. However, most methods focus on the spot (e.g., stock) and could not be directly applied to the futures market with high leverage because of 2 key challenges. 
First, high leverage amplifies reward fluctuations, making RL training highly stochastic and difficult to converge.
Second, prior works lacked self-awareness of capability boundaries, exposing them to the risk of significant capital loss when encountering previously unseen market state representations (e.g., during a black swan event like COVID-19).
To tackle these challenges, we propose the e\textbf{F}ficient and r\textbf{I}sk-aware e\textbf{N}semble r\textbf{E}inforcement learning for \textbf{F}utures \textbf{T}rading (FineFT), a novel three-stage ensemble RL framework with stable training and proper risk management. 
In stage I, ensemble Q learners are selectively updated by ensemble temporal difference (TD) errors, i.e., TD errors across different learners, to improve convergence and performance.
In stage II, we filter the Q-learners based on their profitabilities under different market dynamics and train variational autoencoders (VAEs) on market representations of each dynamic
to identify the capability boundaries of the filtered learners.
In stage III, we dynamically choose from the filtered ensemble and a conservative policy, guided by trained VAEs, to maintain profitability and mitigate risk with new market states.
Through extensive experiments on crypto futures in a high-frequency trading environment with high fidelity and 5$\times$ leverage, we demonstrate that FineFT significantly outperforms 12 state-of-the-art baselines in 6 widely-used financial metrics, reducing risk by more than 40\% while achieving superior profitability compared to the runner-up. Visualization of the selective update mechanism shows that different agents specialize in distinct market dynamics, and ablation studies certify routing with VAEs reduces maximum drawdown effectively, and selective update improves convergence and performance.
\end{abstract}



\maketitle

\section{Introduction}

Futures trading, as standardized financial contracts obligating the purchase or sale of an asset at a predetermined price and date, accounts for over 60\% of the trading volume in the cryptocurrency market~\cite{aleti2021bitcoin}. This dominance is driven by its flexibility to take long or short positions and its high leverage, which attract risk-seeking traders and foster a highly liquid market. With a market capacity of more than 60 trillion dollars, the cryptocurrency futures market represents a crucial domain for quantitative trading.

\begin{figure}[!t]
    \centering
    \includegraphics[width=\linewidth]{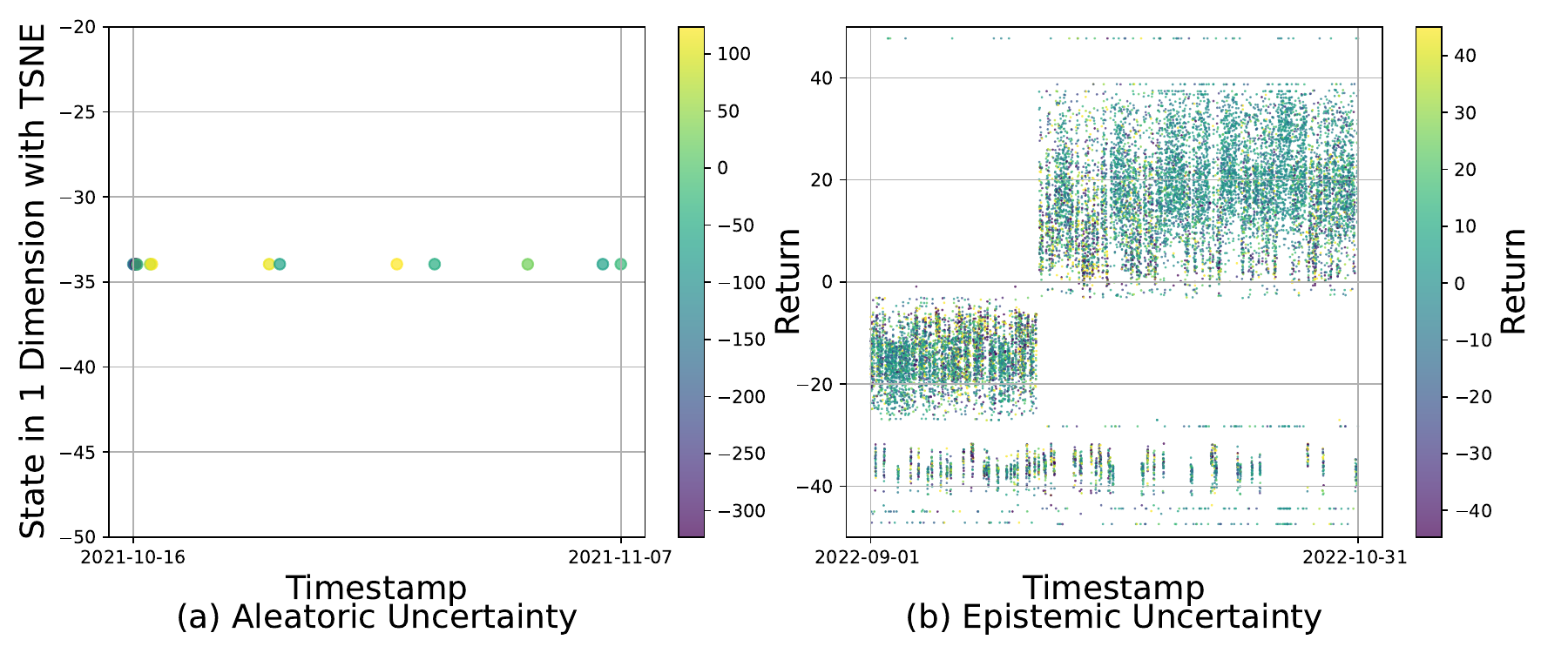}
    \vspace{-0.65cm}
\caption{Two challenges RL faces in future trading. The X-axis is the timestamp; the y-axis represents the 1D t-SNE~\cite{van2008visualizing} embedding of market state (technical indicators) $y_t$, and the colour represents the future 5-min return rate. A detailed description of its generation process is in Appendix~\ref{sec:app_motivation_visualization}. }
    \label{fig: motivation}
    \vspace{-0.65cm}
\end{figure}

As quantitative trading evolves, deep learning (DL) has become popular for modeling complex dependencies using large-scale historical data. In high-frequency trading, reinforcement learning (RL) stands out among DL approaches for optimizing sequential decision-making processes. Unlike supervised learning (SL), which assumes independent and identically distributed (IID) data, RL can handle scenarios where changes in current positions affect trading costs in subsequent decisions due to transaction costs. Although RL has achieved significant success in various quantitative trading tasks and markets~\cite{xia2024market,qin2024earnhft,zhang2024reinforcement,zong2024macrohft}, most studies focus on low-leverage markets such as stock trading, and are less applicable to the high-leverage futures market due to 2 challenges as shown in Figure~\ref{fig: motivation}:
1) High leverage (e.g., 5$\times$) amplifies the inherent stochasticity~\cite{sun2023mastering} of financial markets, meaning that even in similar states, identical actions can lead to vastly different rewards and subsequent states. This phenomenon is illustrated in Figure~\ref{fig: motivation} (a), where the future price movements, represented by the color coding, show significant variability even for nearly identical market state values along the y-axis (1-dimensional market state $y_t$) at different timestamps (x-axis). Such differences highlight the aleatoric uncertainty inherent in financial data, which undermines the convergence and stability of RL algorithms, reduces data efficiency, and can lead to performance decay~\cite{stulp2012model}.
2) Previous works have overlooked the importance of RL agents' self-awareness of their capability boundaries, i.e., the market states it can handle. Due to RL's tendency to overfit the training dataset, the policy becomes unreliable when confronted with unseen market states during training, exposing it to substantial capital loss~\cite{lan2022generalization}. For example, a black swan event in late 2022, triggered by the US Federal Reserve's aggressive interest rate hikes, caused significant shifts in market states. As shown in Figure~\ref{fig: motivation} (b), a cluster near -20 in the 1-dimensional state representation suddenly jumped to 20. Such dramatic changes increase the risk of liquidation for RL agents which lack awareness of capability boundaries.

To tackle these challenges, we propose an e\textbf{F}ficient and r\textbf{I}sk-aware e\textbf{N}semble r\textbf{E}inforcement learning for \textbf{F}utures \textbf{T}rading (FineFT)
In stage I, we initiate $N$ deep Q-networks~\cite{mnih2015human} (DQN) as ensemble learners and pre-train them with transitions from demonstrations. Then, an ensemble temporal difference (TD) error matrix, i.e., TD errors across the learners for each transition, is computed. This matrix selectively updates the learner with the smallest TD error with its neighbors, enhancing training stability, performance, and convergence. In Stage II, we back-test each ensemble element on various market dynamics and then filter the ensemble based on their profitability on each dynamic to reduce the size of the strategy pool. Then, variational autoencoders (VAEs) for state representations from each market dynamic are also trained to capture the capability boundary for each learner in the ensemble and detect out-of-distribution (OOD) state representations during the test phase. In stage III, guided by VAE losses of a rolling window of recent market states representations, we dynamically select the optimal agent from the filtered ensemble for in-distribution (ID) market states to maximize profit while using a conservative policy for OOD market states to mitigate risk. 
The main contributions are 4-fold: 
\vspace{-0.05 cm}
\begin{itemize}[left=0em]
    \item We introduce selective updates with neighbors, which enables agents to segment non-stationary environments into stationary sub-environments and learn optimal policies for each, significantly improving training convergence and overall performance.
    \item To the best of our knowledge, FineFT is the first to use VAEs for out-of-distribution (OOD) detection in RL for quantitative trading to reduce risks from unseen market state representations.
    \item Through extensive experiments on 4 Crypto with diverse market dynamics, evaluated at a minute-level scale, FineFT demonstrates high profitability, significantly reduces risk, and outperforms 12 state-of-the-art baseline methods. The ablation studies show that the selective update segments the market into meaningful financial dynamics reduces the convergence steps, and boosts the performance of the ensemble learners. VAEs effectively route learners and significantly reduce the maximum drawdown.
    \item We develop a high-fidelity trading environment with adjustable leverage, transaction costs, slippage and funding fees. The code is at \url{https://github.com/qinmoelei/FineFT_code_space}.
\end{itemize}
\section{Related Works}
We first introduce technical analysis and traditional routing methods, followed by the application of reinforcement learning (RL) in quantitative trading, with a focus on their respective limitations.
\subsection{Technical Analysis \& Traditional Routing}
Technical analysis has been a fundamental tool in financial markets, relying on historical data to predict price movements~\cite{edwards2018technical}. For instance, imbalance volume (IV)~\cite{chordia2002order}, indicating current buying and selling pressures in the limit order book, is useful for short-term price movement prediction, while moving average convergence divergence (MACD)~\cite{krug2022enforcing} which shows price trends in multiple time windows, is commonly used in a single-dynamic market. Routing methods are commonly used in Fintech to combine results from multiple predictors, allowing for more robust decision-making. These methods dynamically select the best-performing model based on historical performance. For instance, the Winnow algorithm~\cite{zhang2000regularized} adjusts weights iteratively, giving more importance to predictors with better performance and improving decision accuracy.

However, technical analysis often experiences significant performance volatility in non-stationary markets like Crypto, which has been repeatedly criticized in recent studies~\cite{qin2024earnhft,liu2020adaptive}. Moreover, because Winnow relies on historical performance to select models, it tends to perform poorly in sudden trend-reversal situations.
\subsection{RL for Quantitative Trading}
RL has demonstrated stellar performance under multiple quantitative trading tasks. EHFT~\cite{qin2024earnhft} and MHFT~\cite{zong2024macrohft} use a hierarchical framework and optimal action value supervisor to train efficient RL on multiple dynamics for high-frequency trading. DRA~\cite{briola2021deep} utilizes LSTM to capture trends with different time window lengths and PPO~\cite{schulman2017proximal} in high-frequency trading. CRP~\cite{zhu2022quantitative} incorporates random perturbation to stabilize the training of a convolutional DQN~\cite{mnih2015human}. RAQR~\cite{bernhard2019addressing} trains a QRDQN to handle the aleatoric uncertainty in the RL and utilizes a risk-averse policy to avoid the worst case. EDQN~\cite{carta2021multi} utilizes different DQNs to handle different markets. SUNRISE~\cite{lee2021sunrise} views the deviation of learners as the aleatoric uncertainty and decreases the learning rate of the uncertain samples.

However, these studies focus on non-leveraged markets, overlooking risk management and convergence issues. Consequently, they lack sensitivity to black swan events, rendering them unreliable when applied directly to leveraged futures in Crypto.
\begin{figure*}[th]
\begin{center}
\includegraphics[width=\textwidth]{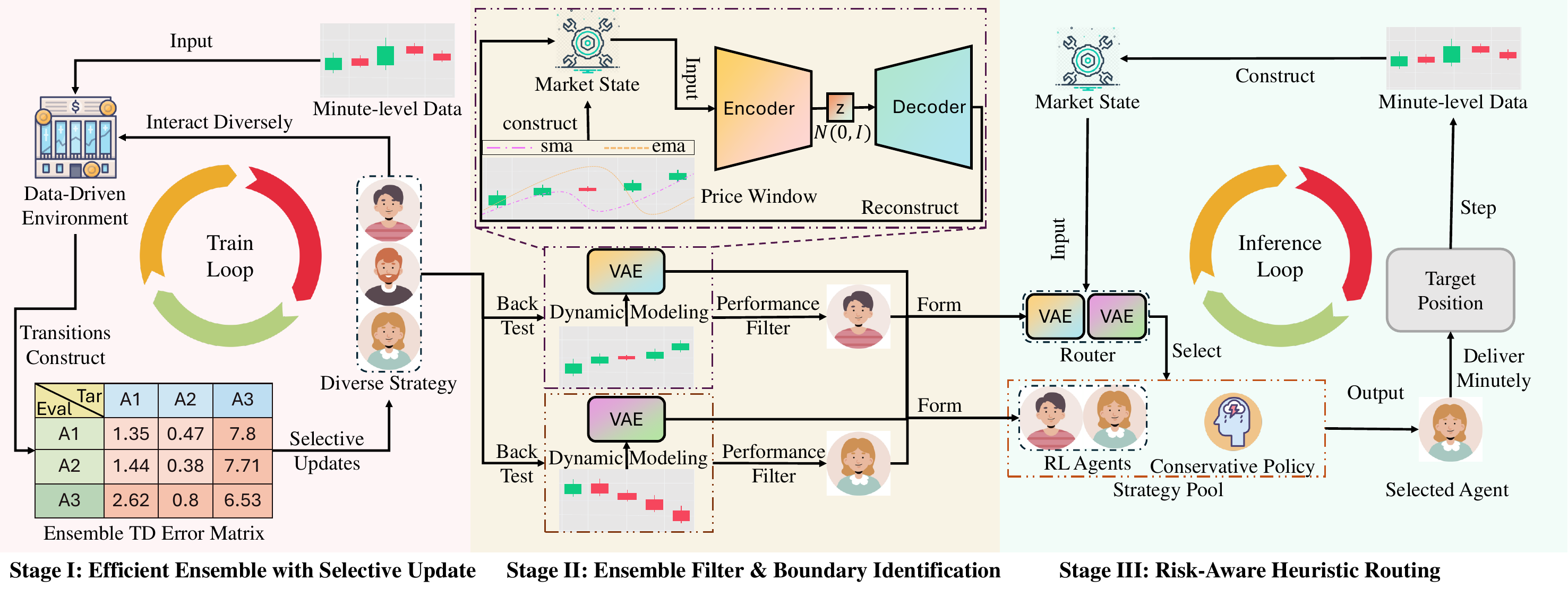}
\end{center}
\caption{The overview of FineFT. First, we compute ensemble TD errors for selective updates to enhance efficiency. Then, we filter the ensemble based on performance across different dynamics and train VAEs for each dynamic to identify the ensemble's capability boundaries. Finally, VAEs route the ensemble and a conservative policy for stable performance and risk reduction.}
\label{fig:pipline}
\end{figure*}
\section{Problem Formulation}
\label{sec:problem_formulation}
This section first presents several finance concepts used to simulate the trading process and then models this futures trading process as a dynamic parametric Markov decision process (DP-MDP).
\subsection{Financial Foundations for Futures Trading}
We first introduce some fundamental financial concepts used to describe state, reward, and action in the DP-MDP framework and present the objective of futures trading using these concepts.

\noindent\textbf{Limit Order book} (LOB) refers to unfilled orders aggregated by price and side, reflecting the micro-structure~\cite{madhavan2000market}. An m-level LOB at time $t$ is denoted as $B_t=(p_t^{b_1},p_t^{a_1},q_t^{b_1},q_t^{a_1},...,p_t^{b_m},p_t^{a_m},q_t^{b_m},q_t^{a_m})$, where $p_t^{b_i}$  ($p_t^{a_i}$)
is level $i$'s bid (ask) price, $q_t^{b_i}$ ($q_t^{a_i}$) is the quantity. 

\noindent\textbf{Trades} refers to executed orders, denoted as $T_{\tau_0:\tau_t}$ from time $\tau_0$ to time $\tau_t$, where an executed order $T_t=(p_t,q_t,c_t)$ at time $t$ is defined by the price $p_t$, quantity $q_t$ and side $c_t$ respectively.

\noindent\textbf{Mark Price} refers to the reference price used to calculate the fair value of the holding position. It is denoted as $M_t$ at timestamp $t$.

\noindent\textbf{Technical Indicators} indicate features calculated by a function of the past trades and LOB to predict future price movements, denoted as $y_t=\phi (B_t,M_t,..., B_{t-h},M_{t-h}, T_{t-h:t})$, where $\phi$ is a function.

\noindent\textbf{Position} is the amount of asset a trader holds, denoted as $H_t$ at time $t$. Relative to 0, it indicates having a long or short position.


\noindent\textbf{Leverage} refers to the ratio of the value of the wanted asset to the capital required (margin), denoted as $L_t$ at time $t$, and it magnifies the maximum position a trader can hold under the same capital.

\noindent\textbf{Market Order Loss} refers to the loss caused by the difference between the executed price and the mark price when executing market orders. It is denoted as $O_t$ and calculated as 
$O_t=c_t(\sum_{i} (p_t^{c_i} \times \min(q_t^{c_i}, \Delta_{i-1}))\times (1+\kappa)-QM_t)$, where $c_t$ is 1/ask for buying and -1/bid for selling, $p_t^{c_i},q_t^{c_i}$ refer to the price and quantity in level $i$ LOB $B_t$, $Q$ is the total trading quantity, $\Delta_{i-1}$ is the remaining quantity after level $i$ in $B_t$, $\kappa$ refers to the commission rate.

\noindent\textbf{Funding fee} refers to interests required to borrow capital for leveraging a position. Denoted as $F_{ft}=F_{rt}\times H_t$, it is determined by a funding rate $F_{rt}$, which is given by the exchange and determined by the price difference between spot and futures, and the position.

\noindent\textbf{Margin Balance} $V_t$ refers to the sum of cash (wallet balance) and position value, where the realized PnL, such as transaction cost or funding fee, is directly deducted from the cash, and the unrealized PnL, such as position value increments due to non-zero position and mark price fluctuations, is included in the total position value. 

\noindent\textbf{Our Objective} is to maximize the margin balance via market orders with leverage on a single asset at a minute-level time scale.
\subsection{Dynamic Parametric MDP Framework}
In this subsection, we formulate futures trading as a DP-MDP~\cite{xie2020deep}. A DP-MDP is defined by the tuple: ($S$, $A$, $P$, $r$, $\gamma$, $T$,$Z$,$P_z$), where $S$ is the state space and $A$ is the action space. 
$Z$ is the dynamic space, which determines the transition function $P: S\times Z\times A\times S \rightarrow [0, 1]$. $Z$ also has its transition function $P_z: Z\times Z \rightarrow [0, 1]$, where the dynamics evolve much slower than state transitions. This indicates that within a short time horizon, the experienced transitions can be viewed as resulting from a static MDP.
$r: S\times A\times S\rightarrow R$ is the reward function, $\gamma \in (0, 1]$ is the discount factor and $T$ is the time horizon. In a DP-MDP, the agent receives the current state $s_t \in S$ from the environment, performs an action $a_{t} \in A$, and moves to the next state $s_{t+1} \in S$ with a reward $r_t$, but the current dynamic $z$ is not explicitly revealed to the agent. An agent’s policy is defined by $\pi_{\theta}$: $S \times A \rightarrow [0, 1]$, which is parameterized by $\theta$. A router's policy is $\pi_{\theta_r}$: $S \times A \rightarrow Z$, parameterized by $\theta_r$.
Our goal is to learn a set of agent policies $\Pi_A=\begin{pmatrix}
    \pi_{\theta_1}, \cdots, \pi_{\theta_m}
\end{pmatrix}$, and a router $\pi_{\theta_r}$ to composite an optimal strategy $\pi^*=\arg\max_{A,\theta_r} E_{\Pi_A,\pi_{\theta_r}}[\sum_{t=0}^T \gamma^tr_t|S_0,Z_0]$,  where $S_0$ is the initial state and $Z_0$ is the initial dynamic. 

\noindent\textbf{State} $S_{t}$ consists of 2 parts: market state, consisting of around 300 technical indicators $y_{t}$ and personal state, which is the position $H_t$. 

\noindent\textbf{Dynamic} $Z_t$ refers to the dependence of market state and price movements, which cannot be reflected in the current market state.

\noindent\textbf{Reward} $r_t$ denotes margin balance change, including valuation and order loss~\cite{qin2024earnhft,zong2024macrohft},
and is computed in Equation~\ref{eq:reward_calculation}. 
\begin{equation}
\label{eq:reward_calculation}
    r_{t}=V_{t+1}-V_{t}=H_t(M_{t+1}-M_t)-O_t
\end{equation}

\noindent\textbf{Action} $a_t$ at time $t$ refers to the target position with corresponding leverage choice from a finite pre-defined pool $A=\{ -H_{l_1},\cdots, H_{l_n} \}$, where $\{-H,\cdots, H\}$ represents the position pool and $\{l_1,\cdots, l_n\}$ is the leverage pool. We conduct the market order instantly to fill the gap between the current position and leverage and the target.

\section{Method}
\label{sec:method}
In this section, we demonstrate three stages of FineFT as shown in Figure~\ref{fig:pipline}. 
In stage I, we selectively update the ensemble deep Q-networks (EDQN) based on the ensemble TD errors to enhance training efficiency. In stage II, we filter the ensemble based on profitability under various dynamics to reduce the strategy pool size and train VAEs for each dynamic to identify the filtered learners' capability boundaries regarding market state representation. In stage III, we dynamically choose from the filtered ensemble and a conservative policy, guided by VAEs' losses, to ensure stable performance in the non-stationary environment and mitigate risk when confronting unseen market state representations.
\begin{algorithm}[tb]
\raggedright
\caption{Constrained Optimal Action Value
\\ ({\small \textcolor{gray}{Line 4: Dynamic Programming with Eq~\ref{eq:reward_calculation}; Line 6: Constraint Masking})}}
\label{alg:constraint_optimal_action_value}
\noindent \textbf{Input}: LOB Series $B$, Markprice Series $M$ with Length $N$, Action Space $A$, Mask Punishment $P$, Capital $C$.\\
\noindent \textbf{Output}: Tabular $Q^*$ of Optimal Action Value at Time $t$, Capital $C$, Position $p$, and Action $a$.\\
\begin{algorithmic}[1] 
\State Initialize $Q^*$ with shape $(N,|A|,|A|)$ and all elements $0$.
\For{$t \gets N-1$ to $1$} 
    \For{$p,a \gets 1$ to $|A|$}
        \State $Q^*[t,p,a]\gets\mathop{\max}\limits_{a'} Q^*[t+1,a,a']+r_t$ \Comment{Future LOB}
        \If{$V_t \leq C$}
            \State $Q^*[t,p,a] \gets Q^*[t,p,a] - P$ \Comment{Apply Masking}
        \EndIf
    \EndFor \Comment{Rollout All Position-Action Pairs}
\EndFor \Comment{Rollout Each Timestamp}
\State \textbf{return} $Q^*$
\end{algorithmic}
\end{algorithm}

\subsection{Efficient Ensemble with Selective Update}

While a mixture of experts (MoE) has shown exceptional results in quantitative trading tasks~\cite{sun2023mastering}, it comes with a significantly increased computational burden. Unlike previous approaches aiming to improve each learner's training efficiency~\cite{qin2024earnhft}, our method first equally pre-trains each learner to establish common knowledge, then employs a selective update strategy that prioritizes updates for the learner with the best estimation, i.e., the smallest TD error, of the current dynamic transitions, along with its neighbors to reduce the computational cost. Our goal in this stage is to train an ensemble DQN where each individual agent within the ensemble learns an optimal policy tailored to a specific dynamic $Z$ in the DP-MDP.

First, each learner in the ensemble is initialized with the identical multilayer perception (MLP) structure but a different random seed. Then, we individually assign a unique index from 1 to N to each learner, where N is the number of learners, before equally pre-training the ensemble and selectively updating with neighbors.

\noindent\textbf{Pretrain with Demonstration.}
In real trading companies, most new traders are trained together before trading separately. Inspired by this, we first train all the agents with the same transitions equally before the selective update. Inspired by~\cite{qin2024earnhft}, a constrained optimal action value is constructed as shown in Algorithm~\ref{alg:constraint_optimal_action_value} to create demonstration transitions as an optimal actor by choosing the action with the largest action value given the state and, as an optimal value supervisor, to help update the EDQN. The computational complexity of this process is $O(T)$, where $T$ denotes the number of timestamps. We further utilize 3 additional policies to generate more diverse transitions for better generalization, which include 1) emptying the position, 2) longing the maximum position, and 3) shorting the maximum position. We update each agent using Huber TD error to avoid the influence from outliers and the optimal value supervisor to boost the learning of advantage value across different actions for the same state as shown in Equation~\eqref{eq:loss_function_pretrain},
\begin{equation}
\label{eq:loss_function_pretrain}
L(\theta_i) =  L_{td_i} +\alpha KL(Q(s, \cdot; \theta_i)||Q^*(s, \cdot)) 
\end{equation}
where $KL(Q(s, \cdot; \theta_i)||Q^*(s, \cdot)) $ is the KL-divergence of the agent with index $i$'s action value and the optimal action value. $L_{td_i}$ is the Huber temporal difference error and is defined as 
\begin{equation}
 L_{td_i} = \mathcal{H}\left( r + \gamma \max Q(s, \cdot; \theta_i') - Q(s,a;\theta_i) \right)
\end{equation} $\gamma$ is the decay coefficient, and $\mathcal{H}(x)$ is the Huber loss~\cite{huber1992robust}, preventing influence from outliers by using mean absolute error for large error and mean square error for small scale errors. It is defined as
\begin{equation}
\label{eq:hubber_loss}
\mathcal{H}(x) =
\begin{cases} 
\frac{1}{2}x^2 & \text{if } |x| \le \delta \\
\delta(|x| - \frac{1}{2}\delta) & \text{otherwise}
\end{cases}.
\end{equation}
During pre-training, each learner is updated equally regarding each transition, and the total loss for all learners could be described as  
\begin{equation}
\label{eq:loss_pretrain_calculation}
 L(E)_{prt} = \sum_{i}^N L(\theta_i).
\end{equation} 
Pretraining with demonstration enables the ensemble to rapidly acquire fundamental trading principles (e.g., not trading too fast or recklessly), thereby expediting the convergence process.

\begin{algorithm}[!t]
\raggedright
\caption{Construction of Weight Matrix
\\ ({\small \textcolor{gray}{Line 3-5: Diagonal Element; Line 6-9: Non-Diagonal Element})}}
\label{alg:weight}
\noindent \textbf{Input}: ETD Error Matrix $\mathcal{L}$, Neighbor Number $m$\\
\noindent \textbf{Output}: $\mathcal{W}$ indicating the Weights assigned to Learners\\
\begin{algorithmic}[1]
\State Initialize $\mathcal{W}$ with shape $(N, N)$ and all elements $0$
\State $i^* \gets \arg\min_{i} L_{ii}$; $i_{\min} \gets \max(i^* - m, 1)$; $i_{\max} \gets \min(i^* + m, N)$ \Comment{Confirm Update Agent Index Range}

\For{$i = i_{\min}$ to $i_{\max}$}
    \State $\mathcal{W}_{ii} \gets 1 - \frac{|i - i^*|}{i_{\max} - i_{\min}}$ \Comment{Diagonal Element Decay}
\EndFor

\For{$i, j \gets 1$ to $N$}
    \If{$i \neq j$}
        \State $\mathcal{W}_{ij} \gets \min(\mathcal{W}_{ii}, \mathcal{W}_{jj}) \times \left(1 - \frac{|i - j|}{i_{\max} - i_{\min}}\right)$ 
    \EndIf \Comment{Non-Diagonal Decay}
\EndFor

\end{algorithmic}
\end{algorithm}


\noindent\textbf{Selective Update with Neighbors.}
Though various dynamics share some fundamental trading principles, the most profitable strategies under different market dynamics often conflict with each other, highlighting the importance of a diverse strategy pool. Previous works~\cite{qin2024earnhft,sun2023mastering} use preference sampling to update the learners independently for the ensemble's diversity, causing a huge computational burden. To address this problem, we demonstrate a more efficient way of training ensembles: assigning transitions with different weights to different agents. Specifically, we first collect enough transitions generated by $\epsilon$-greedy policy from each learner and the optimal actor from the optimal action value. Then, for each transition, we assign larger weights to the learner with a more accurate Q-value estimation and its neighbors (learners whose index is close to the accurate learner's index), i.e., we prioritize updating the agents already with strong performance and their neighbors in the given dynamic while keeping those with poor performance untouched, because those learners with poor performance under this given dynamic may perform well under a different dynamic and using the experience under the given dynamic will poison their performance on the dynamic where they excel. This method is effective because the agent’s updates are tied to transitions from a specific dynamic, preventing random updates and ensuring more focused learning. When the agent is updated with transitions from a particular dynamic, its Q-value estimates improve within that context. As these estimates become more accurate, the TD error from the same dynamic decreases, making it more likely that future transitions from the same dynamic will be used for subsequent updates for the same agent. This creates a positive feedback loop, progressively reinforcing the agent’s specialization within that dynamic.

To determine the specific weights assigned to different learners for each transition, an ensemble temporal difference (ETD) error matrix for each transition is firstly calculated in Equation~\ref{eq:def_ETDMatrix}. 
\begin{equation}
\label{eq:def_ETDMatrix}
L = \begin{pmatrix}
L_{11}  & \cdots & L_{1N} \\
\vdots & \ddots  & \vdots \\
L_{N1} & \cdots & L_{NN}
\end{pmatrix}
\end{equation}
$L_{ij}$ is the Huber TD error across learner $i$ and $j$, defined as Equation~\eqref{eq:df_TDMatrix_element}, where $L_{ii}$ is just the Huber TD error of agent $i$. 
\begin{equation}
\label{eq:df_TDMatrix_element}
L_{ij} = \mathcal{H}\left( r + \gamma \max Q(s', \cdot; \theta_j') - Q(s,a;\theta_i) \right)
\end{equation}
Then, we compute the weight matrix, whose shape is the same as the ETD error matrix, using Algorithm~\ref{alg:weight}, where we first find out the index of the learner with the least Huber TD error $i^{*}$, then expand the index range according to the predefined neighbor size $m$. Then, we compute the weights for the diagonal elements, where $i^{*th}$ element's weight is 1, and the rest of the weights decay to $\frac{1}{2}$ for the boundary element $i_{min}$ and $i_{max}$. Those indices not covered in the range remain 0. Finally, we calculate the non-diagonal element as the minimum diagonal value at its row and column times a decay based on the absolute value of the subtraction between its row and column indices. After constructing the weight matrix for the given transition $(s, a,r,s')$ with $O(N+m^2)$ complexity, where $N$ is the ensemble size and $m$ is the neighbor size, we further take its diagonal elements to form a weight vector $
\mathbf{d}_{\mathcal{W}} = \begin{pmatrix}
\mathcal{W}_{11} & \cdots & \mathcal{W}_{NN}
\end{pmatrix}
$ as a weight for each learner with the optimal supervisor. 
The total loss for selective update given the transition $(s,a,r,s')$ is defined as
\begin{equation}
\label{eq:final_loss_func}
    L=\sum_{i=1}^{n} \sum_{j=1}^{n} \mathcal{W}_{ij} L_{ij}+\mathbf{d}_{\mathcal{W}} O_{\mathrm{KL}}^{T}
\end{equation}
where $O_{\mathrm{KL}} = \begin{pmatrix}
\mathrm{KL}_{1} & \cdots & \mathrm{KL}_{N}
\end{pmatrix}$ is the optimal supervisor loss vector and $\mathrm{KL}_{i}=\mathrm{KL}(Q(s, \cdot; \theta_i)||Q^*(s, \cdot))$ is the optimal supervisor's loss for agent $i$. The final loss could be viewed as a weighted version of Equation~\eqref{eq:loss_pretrain_calculation} with some cross-learner regulation for similar dynamics, improving the training efficiency and performance upper bond by sharing transitions among learners for similar dynamics and preventing experience poisoning from extremely different dynamics. Additional convergence analysis is in Appendix~\ref{Appendx:sec:Proof}.
\subsection{Ensemble Filter \& Boundary Identification}
To effectively route the ensemble during the testing phase, it is crucial to identify the market conditions under which each learner performs well or poorly. While previous research has back-tested low-level RL agents under different dynamics to minimize the strategy pool size for improving routing efficiency, we are the first to propose not only filtering the ensemble based on their profitability under various dynamics to reduce the ensemble size but also training VAEs for each market dynamic to capture market state representations, thereby gaining a deeper understanding of the capability boundaries within which learners perform well.

\noindent\textbf{Ensemble Filter based on Profitability Performance.}
After training an ensemble with diverse strategies, it is essential to identify the specific market dynamics for each strategy best suited. Many empirical methods divide a long market time series into various dynamics. Here, we follow the market dynamics division based on slope~\cite{qin2024earnhft}, where the sequence is first passed through a low-pass filter to remove high-frequency noise, and local extrema points are identified to segment the sequence into smaller chunks. If the slope differences between adjacent chunks are insignificant, these chunks are merged, and we continue the merging process until there are no chunks left to be merged. By applying this algorithm, we can obtain $m$ distinct market dynamics, where each dynamic comprises multiple segments of continuous time series data. We then back-test each learner in our ensemble with different initial positions. We then pick the agents that gain the highest average profitability across all initial positions on each dynamic to form an efficient, diverse, profitable strategy pool fit for routing.

\begin{algorithm}[!t]
\raggedright
\caption{Risk-Aware Heuristic Routing
\\ ({\small \textcolor{gray}{Line 1-4: EMA Reference Score Computing; Line 5-9: Routing})}}
\label{alg:Routing}
\noindent \textbf{Input}: Selected Strategies $\Pi=(\pi^1 \cdots \pi^m)$, VAEs $M^{v}$, Reference Scores $R$, State Window $Y^u_T$, time decay $\gamma$, reject threshold $\tau$\\
\noindent \textbf{Output}: Current Strategy $\pi$\\
\begin{algorithmic}[1]
\For{Dynamic $i$ from $1$ to $m$}
    \State Obtain state score $R^i_t$ from VAE $M^{v_i}$, $Y^u_T$ and $R^i$ 
    \State Compute the EMA using Eq.~\ref{eq:EMA_calculation} \Comment{Compute EMA Score}
\EndFor
\State Compute $i=argmax_i Re_T^{iu}$
\If{$h \leq \tau$}
    \State Return conservative policy $\pi^c$ \Comment{Choose Conservative strategy}
\Else
    \State Return $\pi^i$ \Comment{Choose Heuristic Strategy}
\EndIf
\end{algorithmic}

\end{algorithm}

\noindent\textbf{Boundary Identification for State Representation.} 
Picking profitable learners under different market dynamics divided empirically is insufficient for safe routing. We also should know the capability boundary of each learner within which it can perform optimally. To better understand these dynamics, we train VAEs~\cite{kingma2013auto} for the market state representations from each market dynamic. A VAE consists of an encoder compressing the input into a low-dimensional distribution, regularized with a normal distribution, and a decoder reconstructing the original input based on the sample drawn from the low-dimensional distribution. Unlike standard autoencoders, VAEs provide a probabilistic measure of reconstruction that accounts for the variability in data distribution, making them a more principled and objective anomaly detection method and, therefore, is often used in OOD detection. Here, for a total of $m$ market dynamics, we first initialise $m$ VAEs with identical MLP structures and random seeds. Then, for each market dynamic $z_i$, the market states $y_t$ for $t$ belonging to dynamic $i$ are collected to make the dataset $Y_i=\begin{pmatrix}
    y_{i1} &  \cdots &  y_{in_{i}}
\end{pmatrix}$ to update VAE $i$. The loss function for each VAE $i$ is defined as
\begin{equation}
    L^i(y \in Y_i)=N_{LL}(\mu_d,0.5\log(\sigma^2_d),y)+KLD(\mu,\log(\sigma^2)),
\end{equation}
where the negative likelihood loss $N_{LL}(\mu,\log(\sigma^2),y)=\frac{1}{2} \left( \frac{(y - \mu)^2}{\exp(\log\sigma)^2} \right) + \log\sigma + \frac{1}{2} \log(2\pi)$ refers to the reconstruction loss and $KLD(\mu,\log(\sigma^2))=KL(N(\mu,\sigma^2)|N(0,I))$ refers to normal distribution regulation. After training 2000 epochs, we record $-L^i(y)$ of each VAE $M^{v_i}$ for every market from the corresponding dynamic $z_i$, denoted as reference scores $R^i=\begin{pmatrix}
-L^i(y_{i1}) & \cdots & -L^i(y_{in})
\end{pmatrix}$. So, in conclusion, we have a set of filtered policies $\Pi=\begin{pmatrix}\pi^1 & \cdots & \pi^m \end{pmatrix}$, a set of trained VAEs $M^v=\begin{pmatrix}M^{v_1} & \cdots & M^{v_m} \end{pmatrix}$, and their corresponding reference scores set $R=\begin{pmatrix}R^1 & \cdots & R^m \end{pmatrix}$, which will be used in routing.


\subsection{Risk-Aware Heuristic Routing}
In Stage III, we construct a conservative policy for our strategy pool to handle unseen markets in train or valid datasets. Then, we design a risk-aware heuristic routing mechanism to choose the proper agent to gain stable performance under a rapidly changing dynamic and avoid risks in the new market state representations.

\noindent\textbf{Conservative Policy Design.} We need a conservative strategy to handle market state representations we have never encountered during the training or valid phase. We design the conservative policy as maintaining the position if the maximum drawdown of this single trade is not over 5\% or closing the position otherwise. This conservative policy only focuses on closing the position instead of opening positions, reducing the frequency of position changes. It also ensures that the position is closed to mitigate further losses once a stop-loss is triggered under unseen market representations.

\noindent\textbf{Routing with VAEs' Losses.}
Though a few previous works~\cite{qin2024earnhft,zong2024macrohft} have constructed hierarchical MDP and let the high-level agent serve as the router, it is time-consuming to train such a meta-policy because of RL's poor data efficiency. Therefore, we utilize a heuristic method based on scores from recent market states and trained VAEs from stage II to determine which agent to conduct trading. Given a sequence of recent market states $Y^u_T=\begin{pmatrix}
    y_{T-u},\cdots,y_T
\end{pmatrix}$, each market state's score $R^i_t$ is computed via empirical cumulative distribution function (ecdf) from reference scores $R^i$, followed by an EMA smoothing, described in Equation~\eqref{eq:EMA_calculation},
\begin{equation}
\label{eq:EMA_calculation}
  R^i_t=F_{R^i}(-L^i(y_t)), Re_{T_t}^{iu}=\gamma Re_{T_{t-1}}^{iu}+R^{iu}_t,Re_T^{iu}=Re_{T_T}^{iu}
\end{equation}
where $Re_{T_{T-u}}^{iu}=R^i_{T-u}$ for each market dynamic $i$.
 We pick the dynamic $i=argmax_i Re_T^{iu}$ and compare it with a risk threshold $\tau$. If $Re_T^{iu}$ is smaller than $\tau$, we are unfamiliar with the recent market representations and choose the conservative policy for safe consideration. Otherwise, we choose the learner corresponding to dynamic $i$ to conduct trading, as concluded in Algorithm~\ref{alg:Routing}.

\begin{table}[!t]
\setlength\tabcolsep{2.8pt}
    \small
  \centering
  \renewcommand{\arraystretch}{1.2}
  \caption{Dataset statistics detailing market, dynamics, step size, and chronological period, where the dates are in the YY/MM/DD formats, 22/01/01 indicating Jan 1\textsuperscript{st} of 2022.}
    \label{tab:dataset}
    \vspace{-0.3cm}
    \begin{tabular}{lccccc}
    \toprule
    \textbf{Dataset} & \textbf{Test Dynamics}  & \textbf{Size} & \textbf{From} & \textbf{To} \\ 
    \midrule
    BNB/USDT & Bear  & 210241  & 22/01/01 & 24/01/01  \\
    BTC/USDT & Bull  & 210241  & 22/01/01 & 24/01/01 \\
    DOT/USDT & Sideways  & 210241  & 22/01/01 & 24/01/01  \\
    ETH/USDT & Sideways  & 210241  & 22/01/01 & 24/01/01  \\ 
    \bottomrule
    \end{tabular}
    \vspace{-0.6cm}
    
\end{table}

\begin{table*}[!thb]
\centering
\caption{Performance comparison on 4 Crypto markets with 12 baselines, including 4 plain, 3 hierarchical, 2 ensemble, 1 distributional RL, and 2 rule-based methods. \fst{Pink}, \snd{green}, and \trd{blue} results show the \fst{best}, \snd{second-best}, and \trd{third-best} results.}
\label{tab:performance_tex}
\vspace{-0.4cm}
\renewcommand{\arraystretch}{1.2}
  \resizebox{0.95\textwidth}{!}{
\begin{tabular}{cccccccccccccccc}
\toprule
\multicolumn{2}{ c  }{} & \multicolumn{1}{ c }{Profit} & \multicolumn{3}{ c }{Risk-Adjusted Profit} & \multicolumn{2}{ c }{Risk Metrics}    &\multicolumn{2}{ c  }{} & \multicolumn{1}{ c }{Profit} & \multicolumn{3}{ c  }{Risk-Adjusted Profit} & \multicolumn{2}{ c }{Risk Metrics}    
\\ 
\midrule
Market   & Model & TR(\%)$\uparrow$ & ASR$\uparrow$ & ACR$\uparrow$ & ASoR$\uparrow$ & AVOL(\%)$\downarrow$ & MDD(\%)$\downarrow$ &Market   & Model & TR(\%)$\uparrow$ & ASR$\uparrow$ & ACR$\uparrow$ & ASoR$\uparrow$ & AVOL(\%)$\downarrow$ & MDD(\%)$\downarrow$ \\ 
\midrule
{\multirow{13}{*}{\rotatebox[origin=c]{0}{BNB}}} & DRA & -64.79 & -1.51 & -2.58 & -13.43 &  6.53 &  72.88 &{\multirow{13}{*}{\rotatebox[origin=c]{0}{BTC}}} & DRA & 21.93 & 0.94 & 2.12 & 4.90 &  6.90 &  58.28 \\
&  PPO  & -59.58 & -1.64 & -2.59 & -14.12 &  \trd{5.55} &  67.37& &  PPO  &  19.68 & 0.90 & 2.01 & 4.39 &  6.83 &  58.30\\
\cmidrule(lr){2-8} \cmidrule(lr){10-16}

& CRP & -64.79 & -1.43 & -2.59 & -12.98 &  6.92 &  72.88& &
CRP &19.61 & 0.85 & 1.79 & 4.62 &  6.44 &  58.30  \\ 
& DQN &-55.85 &  -1.21 & -2.14 & -13.72 &  6.09 &  \trd{66.09}&& DQN & -82.13 & -12.03 & -4.94 & -51.81 &  \fst{1.77} &  82.37\\ 
\cmidrule(lr){2-8} \cmidrule(lr){10-16}

& MACD &   -82.85 & -3.04 & -4.12 & -22.67 &  6.28 &  88.70 && MACD &  -26.28 & -0.17 & -0.31 & -9.44 &  5.91 &  60.17\\ 
& IV &-71.00 & -1.59 & -2.73 & -14.31 &  6.59 &  73.21&& IV & -17.26 & 0.01 & 0.02 & -4.26 &  5.80 &  63.25\\ 
\cmidrule(lr){2-8} \cmidrule(lr){10-16}

&EDQN & -64.79 & -1.51 & -2.58 & -13.43 &  6.53 &  72.88
&& EDQN &\trd{25.11} & \trd{0.98} & \trd{2.15} & \trd{5.88} &  6.80 &  59.28 \\
&SUNRISE &-66.75 & -1.74 & -2.80 & -14.45 &  6.27 &  74.32
&& SUNRISE &-7.91 & 0.37 & 0.76 & -2.06 &  6.87 &  63.36 \\
\cmidrule(lr){2-8} \cmidrule(lr){10-16}
&RAQR & -68.88 & -1.84 & -3.10 & -14.22 &  6.47 &  73.27
&& RAQR &2.84 & 0.57 & 1.10 & 0.71 &  5.19 &  51.19\\
\cmidrule(lr){2-8} \cmidrule(lr){10-16}

&WINOW & \fst{86.65} & \snd{2.01} & \snd{5.99} & \snd{31.71} &  \snd{5.46} &  \snd{34.93} 
&& WINOW &-34.03 & -0.23 & -0.56 & -8.62 &  7.09 &  55.30 \\
& EHFT & -69.32 & -2.08 & -3.26 & -17.31 & 6.05 &  73.61  && EHFT & \snd{33.44} & \snd{1.42} & \snd{2.73} & \snd{10.78} &  \snd{4.29} &  \snd{42.61} \\
& MHFT & \trd{-50.42} & \trd{-0.81} & \trd{-1.48} & \trd{-9.10} &  6.47 &  67.95 && MHFT & -25.85 & -0.32 & -0.57 & -7.27 &  \trd{4.46} &  \trd{47.97} \\
\cmidrule(lr){2-8} \cmidrule(lr){10-16}
& FineFT & \snd{83.70} & \fst{2.55} & \fst{11.19} & \fst{101.17} &  \fst{3.57} &  \fst{15.49}  && FineFT & \fst{76.90} & \fst{2.13} & \fst{7.44} & \fst{39.51} &  \snd{4.29} &  \fst{23.51}\\
\bottomrule

{\multirow{13}{*}{\rotatebox[origin=c]{0}{DOT}}} &  DRA &  -65.15 & -1.63 & -2.63 & -15.21 &  6.20 &  73.36 &{\multirow{13}{*}{\rotatebox[origin=c]{0}{ETH}}} & DRA & -46.56 & -0.77 & -1.44 & -10.86 &  6.18 &  63.20\\
&  PPO  & -65.65 & -1.63 & -2.64 & -15.18 &  6.27 &  73.80&& PPO  & -46.08 & -0.74 & -1.39 & -10.68 &  6.23 & 63.01
\\
\cmidrule(lr){2-8} \cmidrule(lr){10-16}
 & CRP & -73.13 & -4.12 & -4.09 & -27.08 &  \trd{3.93} &  75.58& & CRP & -46.06 & -0.95 & -1.69 & -11.36 &  5.88 &  62.96 \\ 
& DQN & -64.89 & -1.35 & -2.40 & -13.66 &  6.84 &  73.42  && DQN & -73.15 & -3.94 & -3.88 & -27.31 & \trd{3.89} &  75.48\\ 
\cmidrule(lr){2-8} \cmidrule(lr){10-16}

& MACD & \trd{26.07} & \trd{1.10} & \trd{2.72} & 6.24 &  7.74 &  60.01& & MACD &  -73.93 & -2.78 & -3.49 & -21.52 &  5.30 &  80.85 \\ 
& IV & -66.24 & -1.67 & -2.70 & -15.32 &  6.26 &  73.80 && IV &  -50.53 & -1.04 & -1.82 & -12.29 &  5.87 &  63.87\\ 
\cmidrule(lr){2-8} \cmidrule(lr){10-16}

&EDQN & -65.65 & -1.63 & -2.64 & -15.18 &  6.27 &  73.80
&& EDQN &-46.00 & -0.73 & -1.39 & -10.65 &  6.23 &  62.96 \\
&SUNRISE &-59.18 & -1.02 & -1.84 & -12.51 &  6.94 &  73.60
&& SUNRISE &-81.22 & -4.45 & -4.57 & -23.85 &  4.48 &  83.36 \\
\cmidrule(lr){2-8} \cmidrule(lr){10-16}
&RAQR & -4.45 & -4.47 & -1.76 & -45.61 & \fst{0.10} &  \fst{4.72}
&& RAQR &-5.94 & -4.10 & -2.65 & -13.75 &  \fst{0.21} &  \fst{6.08}\\
\cmidrule(lr){2-8} \cmidrule(lr){10-16}

&WINOW& -81.86 & -4.17 & -4.63 & -26.72 &  4.85 &  83.35
&&WINOW& \trd{26.36} & \trd{1.06} & \snd{3.55} & \trd{6.98} &  6.59 & \trd{37.55}\\
 & EHFT & \snd{89.00} & \snd{1.95} & \snd{4.09} & \snd{19.78} &  6.28 &  57.18 
 && EHFT & \fst{48.72} & \snd{1.49} & \trd{3.33} & \snd{10.66} &  5.65 &  48.33\\
& MHFT  & \fst{220.44} & \fst{3.48} & \fst{10.47} & \fst{65.02} &  5.10 &  \trd{32.44} && MHFT & -75.16 & -2.21 & -3.26 & -15.37 &  6.32 &  81.79 \\
\cmidrule(lr){2-8} \cmidrule(lr){10-16}
& FineFT & 18.39 & 1.08 & 2.50 & \trd{14.46} &  \snd{2.64} &  \snd{21.87} && FineFT & \snd{39.05} & \fst{1.74} & \fst{5.78} & \fst{14.45} &  \snd{2.96} &  \snd{16.96}\\
\bottomrule
\end{tabular}
}

\vspace{-0.5cm}
\end{table*}
\section{Experiment}
\subsection{Experiments Setup}
\noindent\textbf{Datasets.} To comprehensively evaluate the algorithm, testing is conducted on 4 most mainstream cryptocurrencies over a period exceeding 2 years, covering a variety of market conditions, including bull, bear, and volatile markets, as summarized in Table {\ref{tab:dataset}} and further elaborated in Appendix~\ref{Appendx:sec:Dataset}. For the dataset split, we use data from the last 6 months for testing, the penultimate 6 months for validation, and the remaining 1 year for training on all 4 datasets. We first train the EDQN on the training dataset, then segment and label the validation dataset for performance filter \& boundary identification (VAE training) and tune the hyper-parameter (risk threshold $\tau$, time decay $\gamma$, and window length $L$) for heuristic routing on the validation dataset. The final model is tested in the test dataset.

\noindent\textbf{Evaluation Metrics.} We compare FineFT with 12 state-of-the-art baseline models using six widely-used financial metrics, as outlined in previous works~\cite{zhang2024finagent, qin2024earnhft, sun2024trademaster, zong2024macrohft}. These metrics include one profit metric—total return rate (TR)—three risk-adjusted profit metrics: annual Sharpe ratio (ASR), annual Calmar ratio (CR), and annual Sortino ratio (ASoR), as well as two risk metrics: maximum drawdown (MDD) and annual volatility (AVOL). Detailed definitions and formulas for these metrics can be found in the Appendix~\ref{Appendx:sec:evaluation_metrics}.

\begin{table}[!bt]
\setlength\tabcolsep{4pt}
  \begin{center}
  \caption{Convergence efficiency and performance analysis comparison of the efficient ensemble with selective updates. \fst{Pink} and \snd{green} demonstrate the \fst{best} and \snd{second-best} results.}
  \label{table:ablation_SUN}
  \vspace{-0.4cm}
    \begin{tabular}{cccccccc}
    \toprule
    Meth& CS$\downarrow$ & Dyn & RS$\uparrow$&Meth& CS$\downarrow$ & Dyn & RS$\uparrow$ \\ 
    \midrule
     \multirow{5}{*}{EP} & \multirow{5}{*}{\snd{71712}} & 1 &2.86&\multirow{5}{*}{ER} & \multirow{5}{*}{112320} & 1 &2.86 \\
     &  & 2 &2.14& & & 2 &2.15 \\
     & & 3 &0.50& & & 3 &0.48 \\
     & & 4 &4.98& & & 4 &4.98 \\
     &  & 5 &9.14& & &5 &9.14 \\
     \midrule
     \multirow{5}{*}{FP} &  \multirow{5}{*}{\fst{66528}} & 1 &\fst{45.29}&\multirow{5}{*}{FwoP} &  \multirow{5}{*}{79488} & 1 &
     \snd{37.34} \\
     &  & 2 &\fst{18.21}& & & 2 &\snd{16.18} \\
     &  & 3 &\snd{4.77}& & & 3 &\fst{6.39} \\
     && 4 &\snd{24.48}& & & 4 &\fst{25.53} \\
     &  & 5 &\snd{40.23}& & & 5 &\fst{46.17} \\
     \bottomrule
    \end{tabular}
  \end{center}
  \vspace{-0.8cm}
\end{table}

\noindent\textbf{Training Setup.}
All experiments were conducted on a server equipped with four NVIDIA RTX 4090 GPUs and an AMD Ryzen Threadripper PRO 5995WX CPU, with a total computation time of 6 hours. For a fair comparison, we applied the same hyperparameters used in FineFT (if included) to the baseline models. The transaction cost was set at 0.02\%, aligned with Binance's regulations, and a leverage factor of 5 was utilized. Further details regarding the trading and learning parameters are provided in Appendix~\ref{Appendx:sec:experiment_setting}.

\noindent\textbf{Baselines}
To evaluate FineFT comprehensively, we selected 12 baselines to compare with FineFT: 2 policy-based RL algorithms, PPO~\cite{schulman2017proximal} and DRA~\cite{briola2021deep}, 2 value-based RL algorithms, DQN~\cite{mnih2015human} and CRP~\cite{zhu2022quantitative}, 2 widely used rule-based trading strategies, MACD~\cite{krug2022enforcing} and IV~\cite{chordia2002order}, 2 ensemble RL algorithms: ensemble DQN (EDQN)~\cite{anschel2017averaged} and SUNRISE~\cite{lee2021sunrise}, 1 distributional RL algorithm: RAQR~\cite{dabney2018distributional}, and 3 hierarchical RL algorithms: WINOW~\cite{zhang2000regularized}, EHFT~\cite{qin2024earnhft} and MHFT~\cite{zong2024macrohft}. The Appendix~\ref{Appendx:sec:baselines_discription} provides more details about the baselines.

\vspace{-0.3cm}
\subsection{Comparison with Baselines}
According to Table~\ref{tab:performance_tex}, our method achieves the lowest risk in all 4 datasets, the highest risk-adjusted profit in 3 datasets, and the highest profit in 2 datasets. Value-based methods (CRP and DQN) perform well when the market trend is stable and the gap between the validation and the test dataset is small. CRP performs better than DQN, demonstrating the importance of training stability in a high-stochastic environment. The policy-based methods (DRA and PPO) generally outperform DQN. However, they fail to surpass the Constant Rebalanced Portfolio (CRP), suggesting that policy-based approaches may converge prematurely to suboptimal policies in non-stationary environments, thereby failing to identify the global optimum.  Yet the difference between the DRA and PPO is not evident; demonstrating LSTM does not help much in state representation under a stochastic environment. Rule-based methods (MACD and IV) rely heavily on the similarity between the validation dataset and test dataset since we pick the hyperparameter using the result from the validation dataset, and the overall performance is extremely sensitive to those hyperparameters. Ensemble RL methods (EDQN and SUNRISE)  and distributional RL (RAQR) demonstrate stable performance while effectively mitigating risk due to reduced variance. Hierarchical RL (WINOW, EHFT and MHFT) achieve significantly higher profit performance compared to the aforementioned methods. Moreover, their performance is less dependent on the similarity between validation and test datasets, highlighting their robustness in diverse scenarios. However, these hierarchical methods exhibit high maximum drawdowns, attributed to a lack of clear understanding of their capability boundaries. FineFT retains its performance under familiar market states, profiting like the previous hierarchical RL methods, while it becomes conservative when it confronts unfamiliar market states and, therefore, avoids huge losses. More behavioural analysis concerning FineFT and 12 baselines can be found in Appendix~\ref{Appendx:sec:behavior_result}. We also report the statistical significance of the main experimental results in Appendix~\ref{Appendx:sec:main_exp_statics} using daily metrics, showing our method significantly outperforms all baselines. Additional experiments on other future markets, such as corn and ES futures, is provided in Appendix~\ref{Appendx:sec:more_dataset_results}, where FineFT consistently achieves the best profitability and effective risk control.
\begin{figure}[!t]
    \centering
    \includegraphics[width=\linewidth]{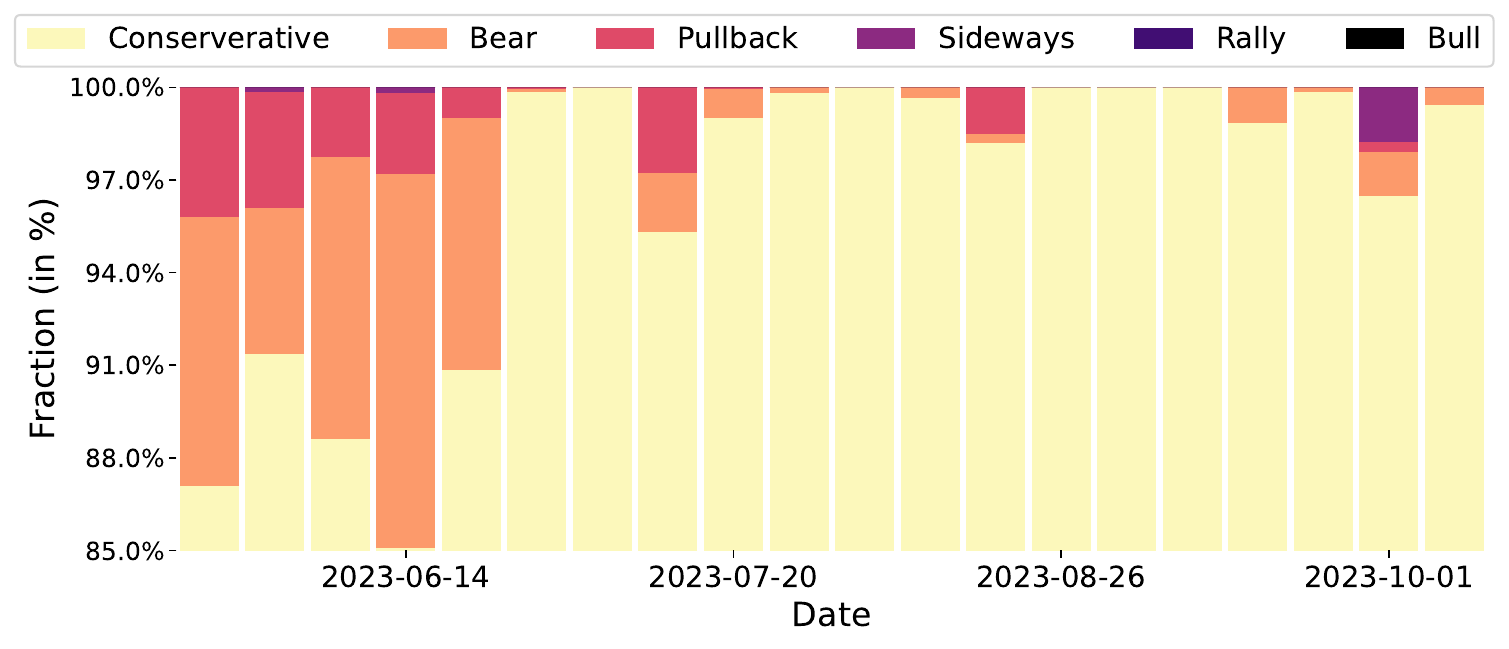}
    \vspace{-0.7cm}
    \caption{Routing result of VAEs along time during the test phase. The x-axis represents the timestamp. The y-axis represents the proportion of different market dynamics. }
    \vspace{-0.3cm}
    \label{fig:ablation_routing_result}
\end{figure}
\vspace{-0.3cm}
\subsection{Ablation}
In this section, we conduct ablation experiments to demonstrate the effectiveness of the efficient ensemble with selective updates and risk-aware heuristic routing on BTCUSDT. More results on other datasets can be found in Appendix~\ref{Appendx:sec:ab_study_ra_routing} and~\ref{Appendx:sec:ab_study_selective_update}. The influence of important hyperparameters (e.g., the number of agents, drawdown threshold) on experimental outcomes is analyzed in Appendix~\ref{Appendx:sec:hyper_influnce}.

\noindent
\textbf{The Effectiveness of Efficient Ensemble with Selective Update.} To demonstrate the improved efficiency and performance in the training strategy pool given by the efficient ensemble with selective update, we conduct an ablation study on whether to use ensemble learning or pre-train on 5 dynamics. We utilize the convergence steps (CS) to demonstrate the training efficiency and reward sum (RS) to demonstrate the converged performance. Besides FineFT with pre-training (the original training method for strategy pool in FineFT, FP), there are other 3 baselines to compare the performance: 1) EHFT with prioritized episode selection (EP), where each agent is trained independently and does not share any transitions but data chunk for each agent's training is sampled with different preference based on market trend for each agent. 2) EHFT with random dataset selection (ER), where each agent is trained independently, and data is randomly sampled for every agent. 3) FineFT without pre-training (FwoP), where agents are not trained equally before selective updates. The result shows that FP and FwoP demonstrate the best converged-result while EP and FP converge within the fewest steps. Overall, the FineFT with pre-training (FP) demonstrates the best performance with the least number of steps required to converge.
\begin{figure}[!t]
    \centering
    \includegraphics[width=\linewidth]{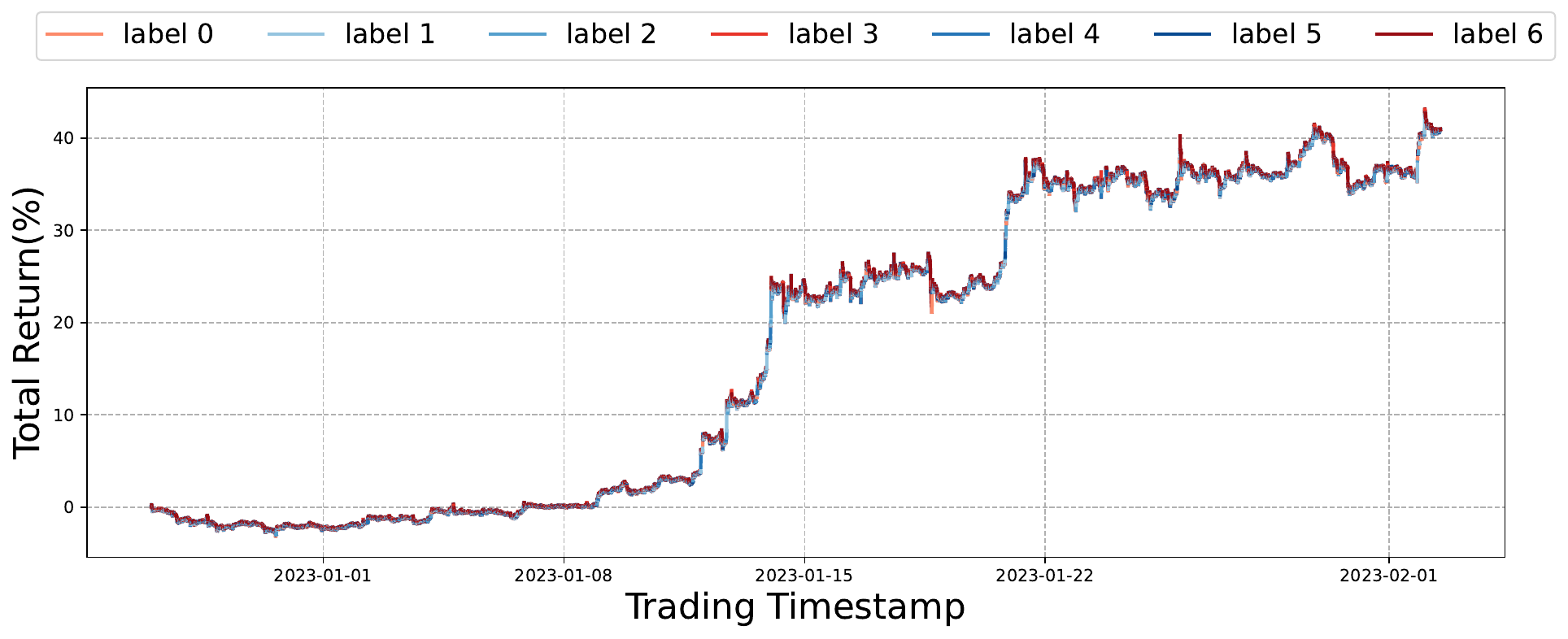}
    \caption{BTC Return Curve: Highlighting the relationship between updated agent indices and market dynamics, with the x-axis representing time, the y-axis showing cumulative returns, and colors indicating updated agent indices.}
    \vspace{-0.5cm}
    \label{fig:update_index_dynamic}
\end{figure}
\vspace{-0.2cm}
\begin{figure}[!thb]
    \centering
    \includegraphics[width=\linewidth]{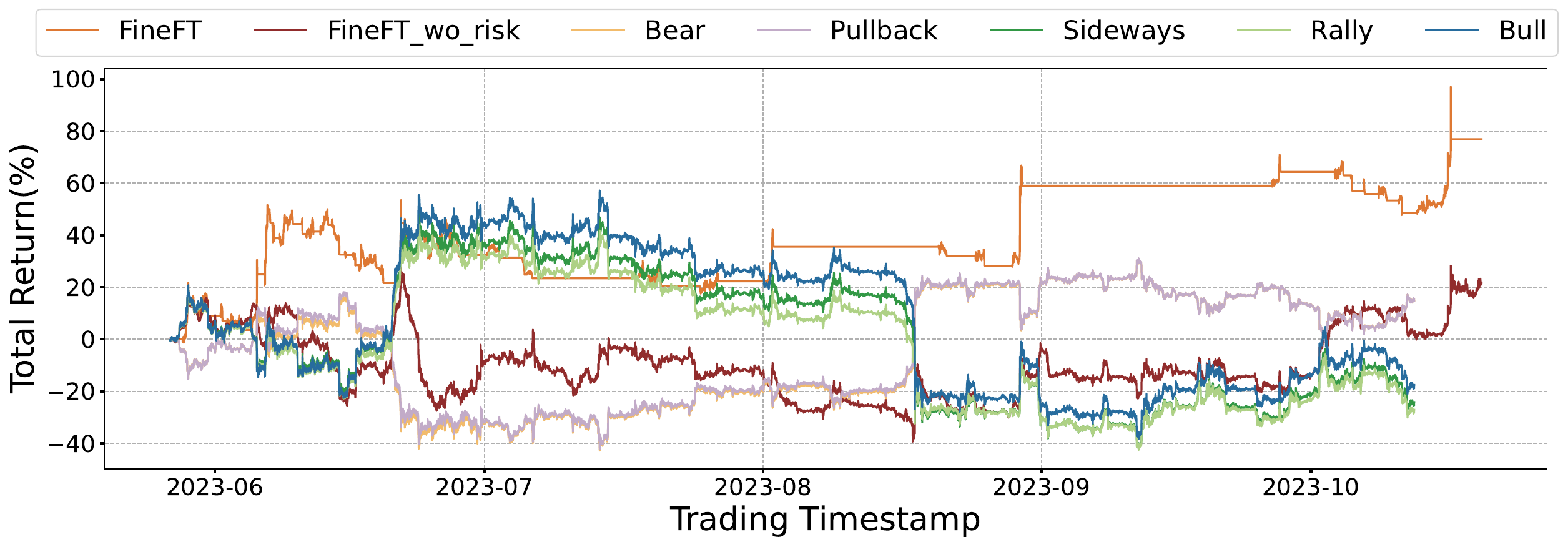}
    \vspace{-0.6cm}
    \caption{Risk-aware routing performance. The x-axis is the timestamp; the y-axis shows FineFT and subagents' returns.}
    \label{fig:performance_comparision_ablation_routing}
\end{figure}

\vspace{-0.3cm}
\noindent
\textbf{The Effectiveness of Risk-Aware Heuristic Routing.}
We examine the effectiveness of the risk-aware routing mechanism by analyzing how the routing results evolve and the performance comparison among the FineFT, FineFT without the risk threshold (FineFT\_wo\_risk), and each agent from the strategy pool. Figure~\ref{fig:ablation_routing_result} shows how the routing result evolves. In most periods, the conservative policy dominates the routing outcome, and we can see an increase in the proportion of conservative policy as the timestamp is further away from the train and valid dataset.
Figure~\ref{fig:performance_comparision_ablation_routing} is the performance comparison of the routing. It shows that FineFT is the best, while FineFT\_wo\_risk beats any other single agent. Agents for a bear market and pullback market are very similar, making a profit in July when the price dropped, while the sideways, rally, and bull market made profits from August to October when the price increased. The FineFT takes advantage of both sides and closes its position in a volatile dynamic, where market state representations are not met in the validation dataset to avoid potential loss.

\subsection{Case Study}
In this section, we first visualize the relationship between market dynamics and the corresponding agent being updated based on selective updates. Then, we present a scenario where the VAE model successfully detects a significant change in market conditions through its rejection mechanism. This rejection coincided with a fundamental event, leading to a sudden shift in price dynamics.

\begin{figure}[!thb]
    \centering
    \includegraphics[width=\linewidth]{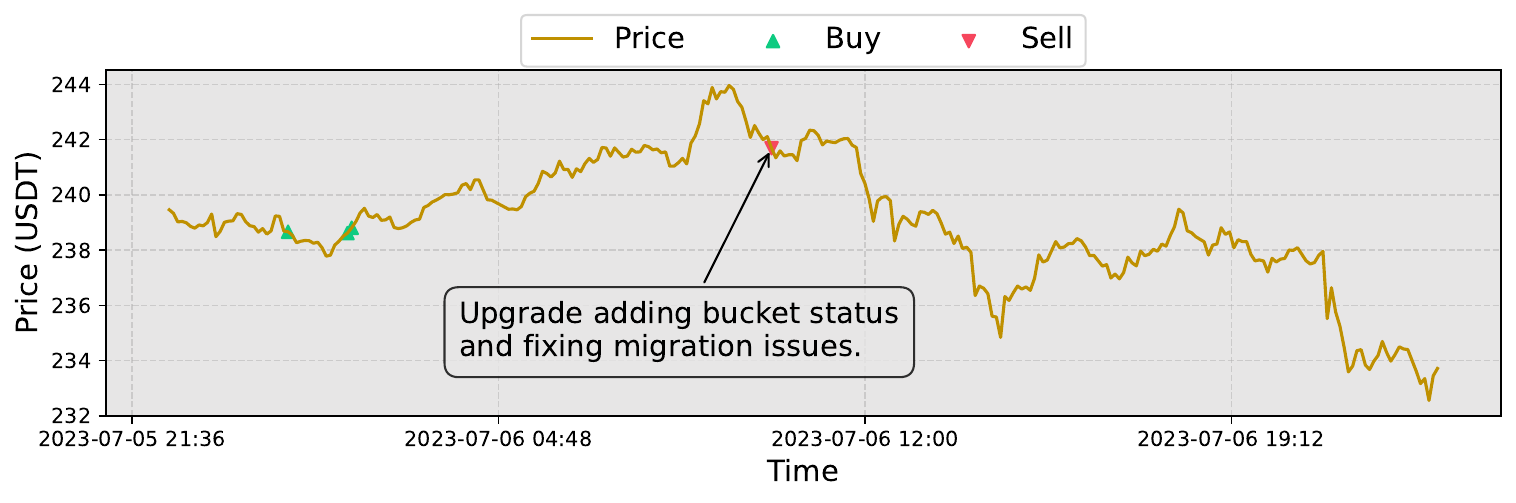}
    \caption{OOD detection with VAE. The X-axis represents the timestamp, and the y-axis is the price of BNB. The red and green symbol shows the buy and sell signal from FineFT.}
    \label{fig:vae_triggered}
\end{figure}
\noindent
\textbf{Visualisation of the market dynamic and selected index.} To validate the proposed selective update with the neighbour mechanism, a visualisation between the market dynamics and updated index is designed to demonstrate that agents do not learn randomly but instead specialise in handling distinct market dynamics. As shown in Figure~\ref{fig:update_index_dynamic}, we utilize 7 agents for selective update with neighbours. Each shows its speciality in the dynamic: agents corresponding to labels 4 and 5 primarily update during sudden price surges, label 7 handles significant price drops, labels 0 and 1 focus on minor fluctuations, and labels 2 and 3 manage larger market movements. This clear alignment between agents and market dynamics validates the selective update with neighbors mechanism's ability to promote specialization and avoid random learning.

\noindent
\textbf{Detecting Market Anomalies with VAEs.} Figure~\ref{fig:vae_triggered} illustrates using a VAE to handle unseen market state representations during test time, focusing on BNB's price movements in July 2023. Despite no explicit closing signals from the selected agent, FineFT proactively closed a position upon encountering unfamiliar states, indicated by a large mean VAE loss over a rolling window. This action avoided losses from subsequent price declines, coinciding with a macro event in BNB: an on-chain upgrade aimed at enhancing bucket status management, including introducing a new status classification and resolving previous migration-related issues.
\section{Conclusion}
This paper proposes FineFT, a novel three-stage ensemble RL approach for handling high stochasticity and risk for unseen markets in futures trading. First, an ETD error is computed to update the learner selectively to improve data efficiency and performance. Then, the ensemble is back-tested on various dynamics modelled by VAEs. Finally, we utilize risk-aware heuristic routing to avoid potential loss caused by epistemic uncertainty. Extensive experiments show FineFT's high profitability and strong risk management. Visualization and trading cases demonstrate that selective updates enable automated market segmentation, and the risk-aware routing mechanism effectively detects macroeconomic events.
Ablation studies demonstrate their effectiveness in performance.

\clearpage
\section*{Acknowledgments}
Shuo Sun is supported by the Guangzhou-HKUST(GZ) Joint Funding Program (No. 2024A03J0630).
\bibliographystyle{ACM-Reference-Format}
\bibliography{fin}

\clearpage
\appendix

\onecolumn
\twocolumn

\section*{APPENDIX}
\section{Notation Indication for this paper}
\label{app:notations}
We present a summary table that defines key financial notations (which means they appear at least twice in this paper) and elucidates their conceptual alignment with the reinforcement learning (RL) framework. Unless stated otherwise, all notations used throughout the main text and appendix adhere to those specified in Tables~\ref{Appendix:tab:notation_definition_reference} and~\ref{Appendix:tab:long_notation_definition_reference}.

\begin{table*}[h]
\centering
\caption{Definition \& Notation Reference. Sym. represents symbol and Desc. represent description.}
\label{Appendix:tab:notation_definition_reference}
\begin{tabular}{l ll l l l}

\toprule
\textbf{Sym.} & \textbf{Desc.} & \textbf{Sym.} & \textbf{Desc.} & \textbf{Sym.} & \textbf{Desc.} \\
\midrule
IV & order book imbalance volume & MACD & double moving average based strategy & LOB & limit order book \\
$B_t$ & LOB snapshot at time $t$ & $\Delta_i$ & remaining trading volume after level i & $\kappa$ &commission fee rate\\
$p^{a/b_i}_t$ &level $i$ ask/bid price in LOB at time $t$ & $q^{a/b_i}_t$ &level $i$ ask/bid quantity in LOB at time $t$ & $T_{\tau_0:\tau_t}$ & trades from time $\tau_0$ to $\tau_t$  \\
$T_t$ & single trade at time $t$ & $p_t$ & $T_t$'s executed price& $q_t$ &  $T_t$'s executed quantity\\
$c_t$ &  $T_t$'s executed side & $y_t$ &  market state, technical indicators at time $t$ & $H_t$ & position at time $t$\\
$M_t$ &  Mark price at time $t$ & $L_t$ & leverage at time $t$ & $O_t$ & market order loss at time $t$\\
$F_{ft}$ &  funding fee at time $t$ & $F_{rt}$ & funding rate at time $t$ & $V_t$ & margine balance at time $t$\\
$S$ &  state space & $A$ & action space & $r$ & reward \\
$S_t$ &  $(y_t,H_t)$, state at time $t$ &$a_t$ & action at time $t$ & $r_t$ &  reward at $t$,$V_t-V_{t-1}$, pnl\\
$P$ &  transition function for state &$Z$ & set of market dynamics & $P_z$ &  transition function for $Z$\\
$\mathcal{H}$ &  huber loss &$\mathcal{W}$ & selective update weight matrix & $\mathbf{d}_{\mathcal{W}}$ &  diagonal element of $\mathcal{W}$\\
$O_{\mathrm{KL}}$ &  KL divergence for optimal supervisor &$O_{\mathrm{KL}}^T$ & tranpose of $O_{\mathrm{KL}}$ & $M^v$ &  trained VAE models\\
$Y_i$ & set of $y_t$ for $z_i$ &$z_i$ & market dynamic $i$ & $M^{v_i}$ &  trained VAE for $z_i$\\
EMA & exponential moving average &$R^i$ & reference scores for $z_i$ & $R$ & all $R_i$ \\
$Y^u_T$ & $y_{T-u}$ to $y_{T}$, market state window &$F_{R^i}(b)$ & ecdf for $R^i$ given value b & $R_t^i$ & time $t$ state score \\
$Re_{T_t}^{iu}$ & ema score for $y_t$ in $Y^u_T$ & $Re_{T}^{iu}$ & used score for $y_T$ & $\pi^i$ &selected policy for $z_i$\\
$\Pi$ & all selected policy & $\mu_d$ & decoder output mean & $\sigma^2_d$ &decoder output variance\\
$\pi^c$ & conservative policy & $\mu$ & encoder output mean & $\sigma^2$ &encoder output variance\\
$tsne(y_t)$ & tsne embedding of $y_t$ & $npv_t$ & nominal position value at time $t$ & $E_p$ &expected executed price\\
$E_t$ & executed value & $E_{po}$ &open executed price& $E_{pc}$ &close executed price\\
$f_{cd}$ & funding count down & $P_p$ &position pool& $H_{max}$ &max holding position\\
$rT_i$ & profit made by $i^{th}$ trade & $P_l$ &leverage pool& $l_{max}$ &max leverage\\
\bottomrule
\end{tabular}
\end{table*}

\begin{table*}[h]
\centering
\caption{RL notation reference requiring a longer explanation.}
\label{Appendix:tab:long_notation_definition_reference}
\begin{tabular}{l ll l}
\toprule
\textbf{Sym.} & \textbf{Desc.} & \textbf{Sym.} & \textbf{Desc.} \\
\midrule

$Q(s, \cdot; \theta)$ &  Action values for Q network &$Q^*(s, \cdot)$ &  Action values for optimal policy \\
\bottomrule
\end{tabular}
\end{table*}
While Tables~\ref{Appendix:tab:notation_definition_reference} and~\ref{Appendix:tab:long_notation_definition_reference} provide a comprehensive overview of the financial and RL notations used throughout this paper, we highlight several symbol differences that may be easily overlooked:

\begin{itemize}
    \item \textbf{p,q:} Variables such as $p_t$, $q_t$, and $c_t$ refer to executed trade attributes—i.e., the price, quantity, and direction of individual transactions at time $t$. In contrast, variables like $p^{a/b_i}_t$ and $q^{a/b_i}_t$ refer to the state of the limit order book (LOB) at a given time, indexed by level $i$ and side (ask/bid). The superscript $a/b_i$ is used to differentiate LOB levels and sides, which do not appear in trade-related symbols.
    \item \textbf{r / R:} $r$ and $r_t$ denote raw rewards (margin account differences) at time $t$, while $R^i$ and $R^i_t$ refer to reference-based state scores, and $R$ represents the entire collection of such scores.
    \item \textbf{O:} $O_t$ denotes market order-related losses at timestamp $t$, whereas $O_{\mathrm{KL}}$ and its transpose $O^T_{\mathrm{KL}}$ are used as KL-divergence terms in representation learning or supervision.
    \item \textbf{$\mu$ / $\sigma^2$:} $\mu$ and $\sigma^2$ denote the encoder output mean and variance, while $\mu_d$ and $\sigma^2_d$ refer to decoder outputs. These symbols appear frequently in components of variational autoencoders.
    \item \textbf{y / Y:} $y_t$ denotes the market feature vector at time $t$, $Y_T^u$ is a sliding window with length $u$ of recent $y_t$ (until $y_T$)'s for calculating recent window scores, and $Y_i$ refers to the full state set associated with regime $z_i$ in the valid dataset.
    \item \textbf{P:} $P$ denotes the global state transition function, whereas $P_z$ is a dynamic-specific transition conditioned on a market regime $z$.
\end{itemize}
To avoid potential confusion, we clarify the terminology used in the main paper: all expressions such as $y_t$, market state (representation), state representation, and technical indicators refer to the same concept. Additionally, the term "IV" in this work does not denote implied volatility, as commonly used in options trading, but rather stands for imbalance volume—a technical indicator that captures the disparity between buying and selling pressure.
\section{Difficulties for RL in Finance Market \& Contributions' Originality}
\label{Appx_sec:Difficulties_for_RL_in_Finance}
Utilising RL is difficult in the financial market for two main reasons. The first is state-action, and the return pair is volatile (corresponding to motivation 1 in the main paper), and the unseen state representation in the previous history (corresponding to motivation 2 in the main paper). We first do a thorough description about how we get Figure~\ref {fig: motivation} and describe these two difficulties in detail, accompanied by visualisations, and then further discuss the technical and engineering contributions of this work. We will provide a conclusion regarding the source of motivation and the contributions of this work from both technical and engineering perspectives.
\subsection{Visualization}
\label{sec:app_motivation_visualization}
We begin by constructing a time-series dataset from raw market data, where each sample corresponds to a fixed-length market snapshot spaced at 5-minute intervals. For each time slice, we extract a high-dimensional market state vector $X \in \mathbb{R}^{N \times m}$ and compute a corresponding forward-looking return rate $y \in \mathbb{R}^{N \times 1}$, defined as the log-difference between the mark price at the current timestamp and that observed exactly 5 minutes later, resulting in aligned triplets of (timestamp, state, return). To visualize the structure of market states and explore uncertainty, we apply t-distributed Stochastic Neighbor Embedding (t-SNE) to the state matrix $X$, reducing the original feature space to a one-dimensional latent representation $z \in \mathbb{R}^{N \times 1}$. We adopt the standard implementation from \texttt{sklearn.manifold.TSNE} with default configuration and random state 42: \texttt{n\_components=1}, \texttt{perplexity=30.0}, \texttt{early\_exaggeration=12.0}, \texttt{learning\_rate='auto'}, \texttt{n\_iter=1000}, \texttt{n\_iter\_without\_progress=300}, \texttt{init='pca'},
\texttt{min\_grad\_norm=1e-7}, \texttt{metric='euclidean'},  \texttt{method='barnes\_hut'}, and \texttt{angle=0.5}. To facilitate clearer visualisation and isolate local behaviours, we partition the dataset into contiguous non-overlapping chunks, each plotted individually with reduced point density. For each chunk, we compute two complementary forms of uncertainty: \textit{epistemic uncertainty}, estimated as the standard deviation across the feature dimensions of state vectors within the chunk, and \textit{aleatoric uncertainty}, computed by first binning the chunk’s market states (after dimensionality reduction) into quantile-based groups and then measuring the standard deviation of return rates within each bin; the bin with the highest return variance is used as the representative aleatoric uncertainty of the chunk. Across all chunks, we identify those with the highest epistemic and aleatoric uncertainty, respectively, and select them for visualisation. For visualisation, we construct a time-aligned scatter plot where the x-axis denotes timestamps (spaced every 5 minutes), the y-axis corresponds to the 1D t-SNE embedding $tsne(y_t)$, and the colour of each point encodes the associated return rate. Due to the heavy-tailed nature of raw returns, we apply percentile-based clipping at the 1st and 99th percentiles to suppress outliers and enhance visual contrast. This visualisation captures the temporal evolution and geometric structure of market states while highlighting regions with high predictive uncertainty. It is also worth noting that even the the difference between Figure~\ref{Appendix:fig:au} and~\ref{Appendix:fig:eu}
and Figure~\ref{fig: motivation} is the colour representating the return rate.

\subsection{Unstable State Representation}
\begin{figure}
    \centering
    \includegraphics[width=\linewidth]{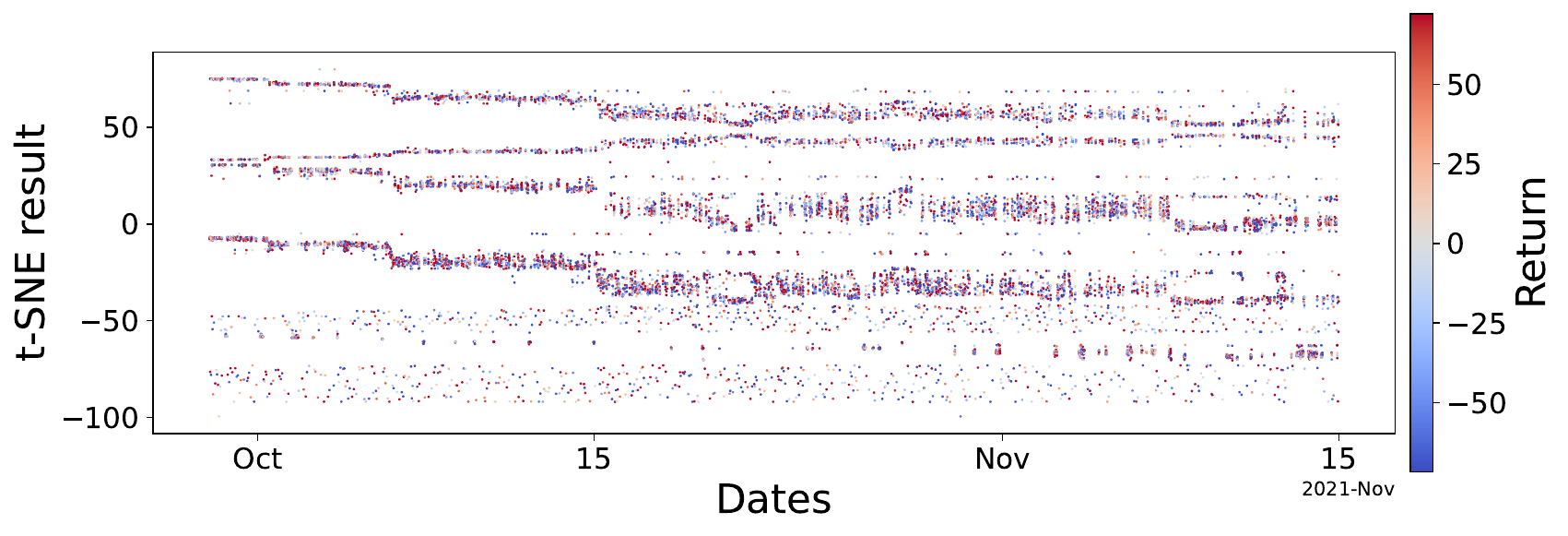}
    \caption{TSNE plot for market state under volatile market: the x-axis shows the timestamp while the y-axis shows the result of market state representation in 1-dimension. The color represents the future price movement.}
\label{Appendix:fig:au}
\end{figure}

\begin{figure}
    \centering
    \includegraphics[width=\linewidth]{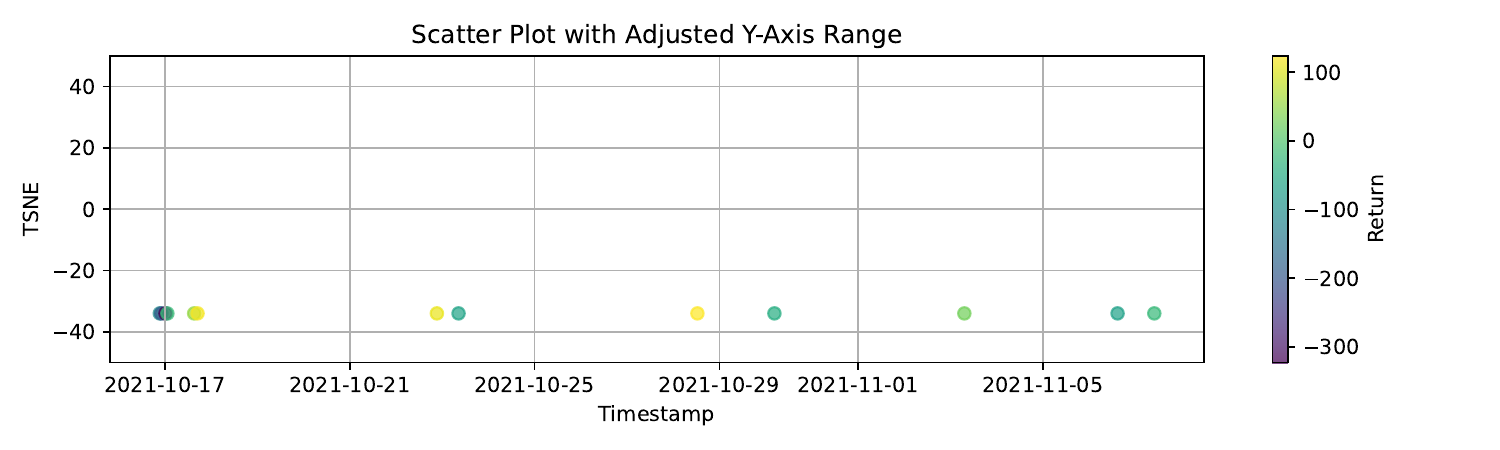}
    \caption{Extraction of a specific y value from Figure~\ref{Appendix:fig:au}.}
\label{Appendix:fig:au_specific}
\end{figure}

The correlation between the state-action pair and the reward and next state is extremely unstable, indicating that the state representation lacks sufficient expressiveness to fulfil the Markov property. Here we use a TSNE plot to demonstrate this point. As shown in Figure~\ref{Appendix:fig:au}, it shows a TSNE plot for a volatile market. We can see that although there are specific y-values where the data points are clustered, e.g., around 0, 100, and 30, the colours of data points from the same cluster are highly mixed, indicating that even with very similar market state representations, the future price movement is highly stochastic. Although the description appears to involve aleatoric uncertainty, further analysis within a smaller range can reveal the true outlook. As shown in Figure~\ref{Appendix:fig:au_specific}, we can see that the returns of a single y value across different periods are not in an identical independent distribution. Around the middle of October, this signal presents a bear market signal since the data points represent a dark blue colour. Yet from the end of October to the end of November, it appears to be a bull signal, showing a green to yellow colour. 

After the beginning of November, the state representation appeared to be a pullback signal that showed a green-to-blue color. Notice that multiple data points exist in each phase, especially in the middle phase, making it statically significant to reject the assumption that they all come from the same distribution and are sampled independently. The correlation of the returns under the similar market state representation is higher when their timestamps are close to each other. This observation coincides with a commonly used trick in quantitative trading: rolling-window training, which retrains the deployed model periodically to get a better result. Previous work on daily-level trading does not work on a large scale (whether a higher frequency or a longer period for training). Therefore, the RL agent could not encounter enough repeated market state representation to cause the convergence problem or observe the agent's policy change dramatically after each training epoch and could not converge at all. They pick one that performs the best under the valid dataset, making the final result highly correlated with the similarity between the valid dataset and test dataset (This coincides with non-hierarchical/Ensemble RL's experiment results in our paper). Limited works talk about the uncertainty in financial data, and even those limited works view this phenomenon shown in Figure~\ref{Appendix:fig:au} as aleatoric uncertainty and, therefore, do not pose a proper solution to it without sacrificing profitability. Our work is the first to analyze a single market representation and see how the return fluctuates over time as shown in Figure~\ref{Appendix:fig:au_specific} and point out that it is not aleatoric uncertainty (at least at high-frequency situations, where our paper's experiments are built) and could be solved by using time correlations. 
The source could be analysised from 2 perspectives. From an RL perspective, the it shows that the environment's dynamic are actively changing. From a financial perspective, the volitale (or more generally: a single-dynamic) market represents stable fundamentals. Then we can view trading as a near-zero sum game. Those who keep losing money in this market will absolutely change their policies to avoid keep losing (usually it will take several days because it requires some time to ensure them that the current policy could not make any money under current market conditions). It is their changing policies that make the dynamic keep changing in a rather slow speed and therefore making the state representation unstable.


\subsection{Constantly occurring distribution shift}
\begin{figure}
    \centering
    \includegraphics[width=\linewidth]{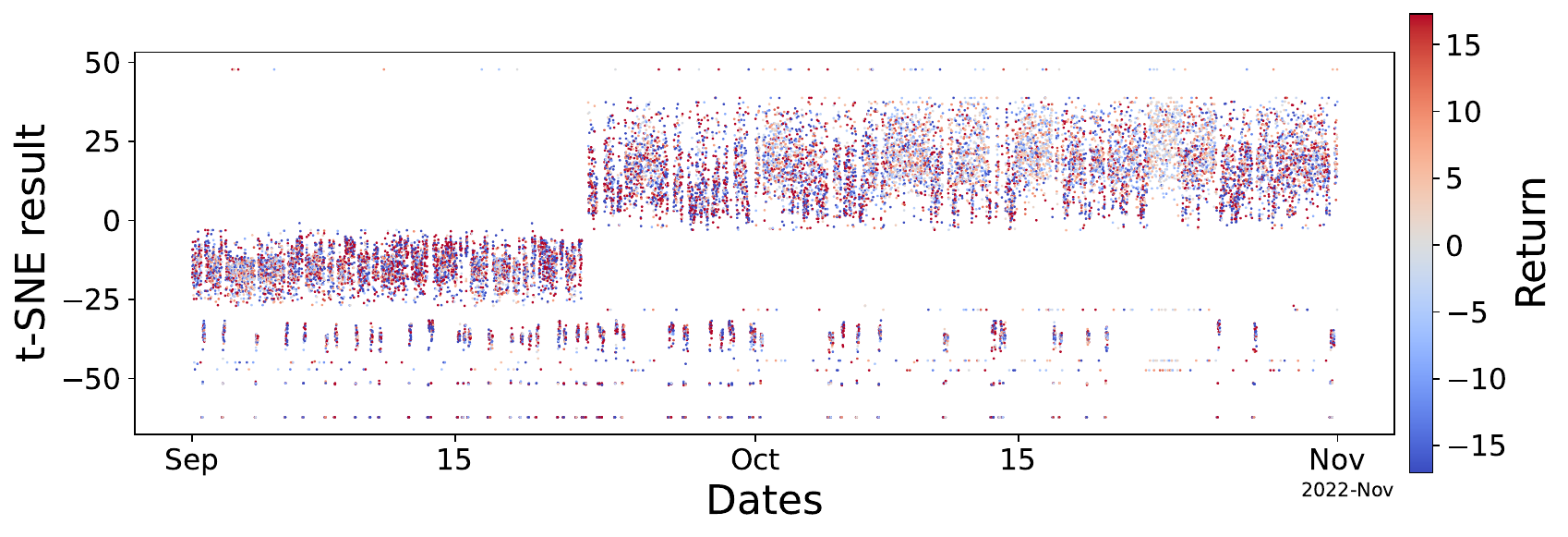}
\caption{TSNE plot for sudden dynamic change. Same setting as Figure~\ref{Appendix:fig:au}}
\label{Appendix:fig:eu}
\end{figure}

We also observe that the state representation could be suddenly changed, and therefore, RL agents have the risk of executing action under an unseen market state. As the TSNE plot shown in Figure~\ref{Appendix:fig:eu}, we can see a sudden change in y value around Oct $21^{st}$. This indicates a significant distribution shift in market state representation at that time. This epistemic uncertainty makes it dangerous for RL agents to be deployed in the real world without any self-awareness concerning their capability boundaries—some fundamental changes cause this. (Here in this picture, around $20^{th}$ September in 2022, the U.S. Consumer Price Index (CPI) data released showed a lower-than-expected increase, indicating a deceleration in inflation. This boosts people to invest in risky assets, such as Bitcoin, making the price go high). Since our market state does not include fundamental information, the state representation could not directly capture this information. Still, when new investors put more money into BTC, it severely changes the distribution of the technical indicators, which our market state is built on.
\subsection{Contributions' Originality}
From the technical perspective, there are mainly 3 contributions:
\begin{itemize}
    \item we propose a selective update based on an ensemble TD error matrix to handle the convergence problem caused by a non-stationary environment in training RL. It largely reduces the convergence steps and boosts the performance under various dynamics. 
    \item We utilize the losses from VAEs as the routing indicator and the OOD detector. It reduces the computational burden while achieving OOD detection and routing simultaneously, reducing the maximum drawdown in the final result while maintaining profitability while previous works focus on utilizing the embedding of VAE (or VQVAE) to improve generalization.
    \item Extensive experiments on various trading pairs in a high-fiedility environment have demonstrated the advantage of FineFT, which achieves high profitability and essentially reduces the downside risk compared with the current SOTA algorithms.
\end{itemize}
The motivation of Contribution 1 from the observation in EarnHFT~\cite{qin2024earnhft}, where we find out that the performance under different dynamics is somewhat conflicting. This paper further analyzes its reason and determines the first difficulty described in Appendix A.1. Inspired by the QRDQN~\cite{dabney2018distributional}, which handles the aleatoric uncertainty (please notice that here we do not conclude the difficulty as aleatoric uncertainty but more like an open-ended multi-dynamic problem), we propose the selective update with ensemble TD error matrices, to get an ensemble of agents where for each dynamic, at least one of the agents could perform well. 

The motivation for Contribution 2 and its difficulty are based on common financial sense. Though it is common knowledge that the distribution shift happens fast in the financial world, no one has done a TSNE plot to show it and analyze why or how fast the distributional shift happens. Our paper is the first to propose using the OOD detection method in RL for fintech to reduce risk.

From the engineering perspective, there are mainly 2 contributions:
\begin{itemize}
    \item We are the first to implement a high-fidelity, data-driven simulation environment for trading perpetual (a special kind of futures, which we will discuss in detail in Appendix C) in Binance, including quick calculation of maintenance margin and available margin, configurable leverage, position, and capital choices, trading cost calculation like market order slippage, funding fee, and commission fee, with a record of margin balance history, position history, and current trade maximum drawdown.
    \item we develop a corresponding data preprocess pipeline for this high-fidelity environment, including configurable down-scale frequency, timestamp alignment, and construction of thousands of technical indicators partially inspired by qlib~\cite{yang2020qlib} and kaggle and different methods to select the proper technical indicators to construct the market state representations.
\end{itemize}

In conclusion, we present two observations using the TSNE plot under different scenarios to illustrate the difficulties of training RL agents for quantitative trading. We then develop the corresponding methods to solve these two problems. We also develop a high-fidelity environment for RL and the corresponding data preprocess pipeline to benefit the RL for fintech communities.

\section{More detailed Related Work}
\label{sec:app_related_work}
\subsection{Technical Indicator Construction in Fintech}
State representation and pairing networks are very important in RL for Fintech because they determine the construction of the underlying MDP. A stable transition will make the training much easier and converge faster.
There are two main technical approaches to addressing this issue: a) pure raw data paired with sophisticated networks and b) well-constructed technical indicators paired with simple networks (MLP and its variants). For path a), these methods either want to integrate as much information from as many sources (mainly in text) as possible and align them at the token level~\cite{yu2024finmem,zhang2024finagent,yang2024finrobot,yu2024fincon} using LLM or follow the path of time series, utilizing either decomposition techniques (different periods~\cite{wu2022timesnet}, trends and volatility~\cite{zeng2023transformers}, and their combination~\cite{wang2024timemixer}) or accelerating or narrowing down the scope of the attention (QKV mechanism~\cite{zhou2021informer}, input-patchifying~\cite{zhang2022patchformer}, etc) or their combination~\cite{wang2021deeptrader}. The advantage of this path is that a) there is no need to spend much time processing raw data. b) the upper potential test result might be higher due to the diversity of information sources. c) the explainability of the LLM might be good. The defects of this path are that a) the result is extremely unstable and easy to be out of distribution. b) It is hard to align these experiences with expertise in finance. The second path is constructing high-quality technical indicators closely related to prices' future movement. Usually, these technical indicators have much better explainability for economists and are directly aligned with financial expertise. The operators calculating these technical indicators are usually simpler, making the delay in real-time trading smaller and, therefore, widely deployed in the real industry. Ever since~\cite{jiang2017deep}, it has been the most widely used setting in RL for fintech, especially for portfolio management. Recently, works combining the two paths have gained attention: ~\cite{sun2022deepscalper}
uses prediction of future trends and volatility as the auxiliary task to enhance the generalization ability of the learned state representation.~\cite{ding2018investor} uses the predicted rank of different technical indicators as the state input of the RL agent.~\cite{niu2023imm} uses the predicted trends with different window length as the input of the RL agent and utilize the trading signal generated by the predicted result as imitation target.
\subsection{RL for Fintech}
\cite{gu2023margin} develop an RL environment for portfolio management with margin, which utilizes a position fusion model and a margin maintenance detection module to handle the new trading setting. Here in our work, the action space is the target position, and we directly encode the previous position into the state, so the first module is unnecessary. We use a mask in the environment to demonstrate which action cannot be taken to substitute the margin maintenance detection module. Our trading setting differs from this work, where their environment is for portfolio management (handling multiple assets simultaneously), and ours is for algorithm trading (focusing on a single asset). \cite{chen2023mtrader} is the first paper to utilize VAE-related techniques in trading tasks. It utilizes a VQVAE~\cite{van2017neural} to replace the traditional technical indicators. It argues that the stable representation of the code book makes the RL agent perform better under out-of-distribution situations. Surprisingly \textbf{there have not been any works about using the OOD detection technique to reduce the risk caused by the epistemic uncertainty in RL for Fintech}.
\subsection{Uncertainty in RL \& their Solutions}
There are two kinds of uncertainty in RL: aleatoric uncertainty and epistemic uncertainty.\cite{ghosh2021generalization} utilize ensemble to solve the epistemic uncertainty caused by the difference between training and testing contexts. Yet the optimization of Bayesian RL can not handle the aleatoric uncertainty because the posterior relies only on the state. Then, it uses a simple ensemble method where it only chooses the most probable action among all the different policies learned from different contexts. Distributional RL optimizes the policy based on people's preferences towards aleatoric uncertainty.
C51\cite{bellemare2017distributional}, QR-DQN\cite{dabney2018distributional}, IQN\cite{dabney2018implicit} and FPQ\cite{yang2019fully} are the most popular distributional RL algorithms, where C51 use bins to model the value distribution, QR-DQN uses a network to simulate the quantiles of the action value distribution. IQN uses a network incorporating the quantiles as the input to simulate the whole distribution. Their utility and effectiveness have been tested in both \cite {bodnar2019quantile} single agent and \cite{qiu2021rmix} multi-agent situations. Though the \cite{bodnar2019quantile} claims that they utilize a risk-sensitive policy after training the overall distribution of the policy by using specific quantiles of the q-values distribution during inference. Yet since the update rules are based on the chosen arg max of the expectation of the next action value, this method only provides a sub-optimal solution for different risk measures after only training.

In this paper, we utilize a heuristic method to replace the 
Bayesian RL in~\cite{ghosh2021generalization} for faster training, because Bayesian RL is extremely computationally burdensome and hard to train because it needs sampling or approximate inference of the posterior distribution of model parameters. Though this heuristic method is not perfect~\cite{zhang2021understanding}
\subsection{RL for dynamics environment}
\cite{lee2020context} propose CaDM, using a context-aware network, whose auxiliary task is to predict the next state based on the previous state, action, and reward or the previous state based on the future state and current action and reward, to generate a context variable used as a part of the state in RL to handle the dynamics change. \cite{chen2021offline} use a dynamics pool to handle the change, using a context variable to determine which dynamics model to use and update.\cite{xue2023state} use the context encoder to help the agent make decisions and use the optimal state distribution consistent with the optimal state in another environment to boost the training efficiency. 
\cite{ghosh2021generalization} demonstrates the difficulties for generalization of RL policies.
\subsection{Ensemble and MoE RL}
Ensemble RL is utilized in many scenarios to achieve stable performance in exchange for higher computational costs.~\cite{lee2021sunrise} transition-based update to form a curriculum process for ensemble RL. By reducing the learning rate on the transitions where each learner's estimation is extremely different, the ensemble prioritizes the update of the data whose aleatoric uncertainty is small. AlphaMix~\cite{sun2023mastering} and AlphaMix+~\cite{sun2023prudex} are largely based on these intuitions and are designed for portfolio management tasks and, therefore, not included in the baselines for comparison. However, after analyzing the motivation with the unstable representations and our following analysis, the turnover of these methods will not be high. \textbf{Our method significantly differs from these methods regarding motivation and methodology.} First, from the motivation perspective, the previous ensemble RL focused on utilizing different learners to capture different samples (transitions) and use the standard deviation of their estimation as a sort of uncertainty metric. At the same time, FineFT wants to construct an ensemble where the upper bound of (the performance of the best learn in the ensemble) the ensemble under various dynamics is high, so it is more like RL dealing with a non-stationary environment. From the method perspective, previous methods derive a conservative policy by computing a lower bound of q-value for each action and choosing the action that performs the best under these pessimistic estimations. The diversity of the agent is guaranteed by sampling independent transitions, which is extremely computationally burdensome. Their efficiency is mostly improved by reducing the freedom of updates for each learner, like low-rank parameter metrics. However, these methods are criticized in~\cite{ghasemipour2022so}. Our method, on the other hand, maintains learners' independence. We keep the ensemble's diversity and improve efficiency by carefully selecting learners given the data. More specifically, we prioritize updates for the learner with the strongest performance on the current dynamic. Generating new data (transitions) is the most costly part of RL training, especially in RL for fintech, where the environments are mostly written in Python, leading to lower running efficiency. The previous methods throw away some transitions (transitions with high aleatoric uncertainty), which wastes resources. And from Figure~\ref{Appendix:fig:au_specific}, you can see that if the replay buffer is large enough, or an on-policy base RL algorithm for ensemble RL is utilized, it is much more likely that either the ensemble RL does not learn much under limited transitions because high aleatoric uncertainty is prevalent in most of the states or stick to a single dynamic (the most recent experience) and can not perform well on other dynamics. Our method also relies on the VAE result for routing, while other methods directly utilize the aggregation of the outputs from the ensemble. Recent works also show utilizing MoE in RL~\cite{obando2024mixtures} to solve large-scale problems. The use of soft MoE enables the performance boost, yet the use of MoE is largely based on the image state input and transformer base network structure. Here, we focus on tabular state input and MLP network structures because previous experience using transformer-based models has demonstrated poor performance. Our method first trains multiple RL learners and then establishes the prior knowledge about these learners concerning their results on various dynamics, which we can specifically judge using VAEs. The multiple RL learners take experts' positions, and the VAEs correspond to the gating mechanism in the mixture of experts, which is a special kind of ensemble learning. Therefore, our method is an ensemble RL and MoE framework.

\section{Financial Bases \& RL Environment}
\label{Appendix:sec:problem_formulation}
In this section, we first discuss the relationship between traditional and perpetual futures. Then, we introduce more financial concepts regarding futures trading, which will be used in the actual paper implementation. Finally, we introduce the RL environment we designed.

\subsection{Futures \& Perpetual}
Futures are contracts obligating an exchange of an asset at a predetermined date for current cash, which is standardized on this predetermined date. For example, on $15^{th}$ Aug 2024, if you buy a BTC future whose delivery day is $1^{st}$ Jan 2025, the exchange delivers the predetermined price \$65000. It means that you should pay the exchange \$65000 on $15^{th}$ Aug 2024, and no matter what price the BTC will be, The exchange will give you a BTC on $1^{st}$ Jan 2025. Normally, the price deviation of the futures and the spot will decay as the delivery day approaches. It was first developed for commodities like oil and beans because of their consistent need, and it gradually expanded to the stock index and crypto. It is a risk management tool, but unlike options, which serve more as an exchange right than an obligation. Perpetuals are a unique type of futures contract characterized by a delivery date set infinitely far in the future. Since there will be no delivery at all, the leverage for this asset can be extremely high. However, the perpetual price needs to align with the spot price as an extension of the no-arbitrage pricing principle. Therefore, the exchange uses the funding fee mechanism to balance the perpetual and spot prices, which will be discussed in Appendix C.2.

\subsection{Financial Bases}
The main paper describes futures trading in an extremely simplified form due to space and self-consistency constraints. Here, we first briefly explain futures trading and the role that exchanges and traders play in the process. Then, I reintroduce the concepts of mark price, leverage, margin balance, and open losses. We introduce new concepts such as funding fees, nominal position value, conditions of opening a position or changing leverage, transaction cost, different margins, wallet balance, and available balance. Finally, we utilize an example of opening a new position with a specific leverage to demonstrate the overall trading process. 

\noindent\textbf{Brief Futures Trading.}
A simplified version of futures trading is that traders do not have enough money to open the target position. Therefore, they borrow money from the exchange to open a full position. Yet the exchange will not risk their own money on the volatile asset, so they ask the traders to leave a portion of the deposit in exchange for the borrowed money (normally if the leverage is low (e.g., 2 $\times$ in the stock market), the borrowed money amount is less than or equal to the deposit. However, in distributed exchange, the borrowed amount of money could be hundreds of times more than the original deposit). The exchanges keep track of traders' trading records. Suppose they find out traders lost almost as much as their original deposit. In that case, the exchange will automatically close all traders' positions (what we call liquidation) and get back their original lending money, leaving the traders almost nothing. Next, we will introduce some concepts used to explicitly calculate the traders' profit and loss and how the exchanges profit from the traders' trading process.

\noindent\textbf{Mark Price} refers to the reference price of a future given by the exchange, used to calculate the nominal position value, which is the estimated value for the position. Mark price at time $t$ denoted as $M_{t}$. Furthermore, the nominal value of the position at time $t$ is defined as $npv_t=M_{t}\times H_t$. This nominal position value (NPV) can be viewed as the true price of risky assets in futures trading. Mark price is extremely important to exchange because it is the price for the exchange to calculate the position and judge whether the account should be liquidated. It should not fluctuate too much since a reference price with high volatility will easily cause the trader with high leverage to liquidate. In Binance, they claim to use an average price of the corresponding spot price, the average price for the best ask and best bid in the current LOB, and other reference prices given by other distributed exchanges like OBX, Bybit, and CoinBase to prevent market manipulations.  

\noindent\textbf{Profit \& Loss} (PnL) refers to the profit or loss made by trading. There are 2 kinds of PnL in futures trading: realized PnL and unrealized PnL. Realized PnL refers to PnL conducted, trading costs like transaction costs, slippage for closing position, and funding fee. Unrealized PnL refers to PnL that has not been conducted, such as the nominal position value fluctuations caused by the mark price change (the current nominal position value minus the nominal position value when opening the position).

\noindent\textbf{Margin} refers to the least money to be kept in exchange for opening or maintaining a position to help cover the credit risk for borrowing the money. This margin refers to the deposit we explain that traders have to keep the money amount in exchange for borrowing money. There are two kinds of margins: initial margin and maintenance margin. The initial margin is the deposit amount traders will deliver when opening a new position. The leverage and the nominal position value determine it. The maintenance margin is the money required to keep the current position, which is determined by the exchange and the current nominal position value. In Binance, a quick-maintenance-calculation table is delivered for each perpetual to calculate the maintenance margin. Since the maintenance calculation involves liquidation, the upper bound of the leverage for calculating the initial margin and portion of the maintenance margin to the nominal position value is increasing regarding the nominal position value. The logic here is that the liquidation, i.e., closing all the positions at a broken price, will cause the mark price to move to the opposite side of the current position, causing the traders losing more money. (For example, if you long 10 coins at mark price 10\$ (here we take the market order's execution as the mark price for simple explanations for margin mechanism consideration, the actual execution in the ), yet you only have 1 dollar, the exchange will lend you 99 dollars for you to finish the trade. Still, once the mark price drops to \$ 9.95 (whose nominal position value is \$ 99.5, meaning currently, your only loss \$ 0.5), though your deposit still left 50\%, the exchange won't let you continuously take the risk, and directly close your position (liquidation), and since you are longing the position, closing position meaning selling your holdings. Selling at the broken price (an extremely low price for quick execution) will cause the order to be conducted in a market-order manner. Therefore, the price will go down, causing the average conducted price to be around \$ 9.94, so after the liquidation, you will only have \$ 0.4. The exchange does not want to undertake the risk of market slippage caused by liquidation therefore, the higher the nominal position value, the larger portion of the margins should be kept in the exchange) This market slippage loss should be taken by traders instead of the broker, and the more nominal position value the trader has, the more likely the slippage will be larger. Therefore, the exchange will set a lower upper bond for leverage and a higher portion of maintenance margin to reduce its own risk. The maintenance margin calculation is 
\begin{equation}
\label{Appendix:eq:main_margine_calculation}
    M_m=kNPV_t-j
\end{equation}
for a given nominal position value. For example, for BTCUSDT perpetual, if the nominal position value is under \$ 50000, then $k=0.004$ and $j=0$. if the nominal positive lue is under \$ 500000 yet higher than \$ 50000, then $k=0.005$ and $j=50$. So that is why liquidation is so dangerous because even if no slippage happens, you at most will have 0.4\% left for your nominal position, saying the leverage is 25 $\times$. If you open a position after liquidation, we will only have 10\% of your original money, meaning your PnL is -90\%. We provide the quick margin calculate ($k$ and $j$) for different tickers and different NPV bound in Table~\ref{Appendix:tab:main_margin}

\noindent\textbf{Funding fee} refers to the interest charged when leveraging funds or assets. It is settled every 4 or 8 hours based on the specific trading pair and calculated as $F^f_t=f_r V^n_t$, where the exchange determines funding rate $f_r$ based on the price difference between the spot and futures markets. This mechanism is designed specifically for perpetual to keep the alignment with spot prices. You can view this as a trading cost(or bonus). The institution is that if the perpetual's price is severely higher than the spot price, meaning that there are too many long-position holders in the market. Therefore, for a fixed-length period, the exchange will punish that long-position holder by deducting a small amount of money from their account and using that money to reward those short-position holders in the market. The exchange does not make a profit from this process (at least Binance emphasizes this point several times on its website). You can also view it as an interest rate problem between long-position and short-position holders. There are too many long-position holders, meaning that those short-position holders are lending USDT to those long-position holders to open such large positions with leverage. Lending the money comes with interest, directed from the borrower's account to the lender's account periodically. If the perpetual price is smaller than the spot price, there are too many short-position holders, and the funding rate becomes negative. The long-position holder will be rewarded with money deducted from the short-position holders. If we view this as an interest rate problem, it becomes that the long-position holders are lending volatile coins (determined by the perpetual; if the perpetual is BTCUSDT, then here is the BTC) to the short-position holders, and they get the interest periodically. If the future price is higher than the spot price, then this funding mechanism will benefit the short-position holder. The long-position holders will tend to close their position, and there will be more selling orders than buying orders, causing the futures' mark price to decrease. if the spot price exceeds the futures' mark price, this funding mechanism will benefit the long-position holder. The short-position holders will tend to close their position, and there will be more buying orders than selling orders, causing the futures' mark price to increase. Therefore, this mechanism causes the perpetual prices to converge to the spot price, maintaining the non-arbitrary-price principle.

\noindent\textbf{Balance.} There are 3 kinds of balance in futures trading, i.e., wallet balance, margin balance, and available balance. The wallet balance refers to the balance integrated with the realized PnL. Initially, if you do not have any positions and transfer your money into a futures account, all the money is in the wallet balance. Trading costs like funding fees, transaction costs, and the realized PnL caused by closing position will be directly added or deducted from the wallet balance. Margin balance refers to the wallet balance plus the unrealized PnL. You can view this as the net value in spot trading (the cash plus the risk asset current value). The margin balance is a key indicator to judge the condition of liquidation. If the margin balance is equal to or smaller than the maintenance margin, all the positions will be closed due to liquidation. During our formulation of MDP, we also utilize the difference in the margin balance between the timestamps as the reward. Available balance refers to the margin balance minus the initial margin for the current position. It reflects the amount of cash left to open new positions. If the avoidance is larger than the open loss plus the initial margin for opening a new position, the position could be opened successfully. It is worth noticing that the available balance could be negative, but it does not mean the account would be liquidated instantly.

\noindent\textbf{Market Order with Leverage} is to execute an order with a target volume instantly with a specific leverage. If the market order is to open a position, then an estimated execution price will be calculated. For a long position, the estimated executed price equals the best ask price times 1 plus 0.5 percent, while for a short position, the estimated execution price is just the best bid price. After the calculation, the exchange will judge whether you have enough available balance (this measurement is for preventing instant liquidation after the execution of the order). The execute value is calculated as Equation~\ref{equation:calculate_market_order}.
\begin{equation}
\label{equation:calculate_market_order}
E_t(Q_t) = \sum_{i} (p_t^{c_i} \times \min(q_t^{c_i}, \Delta_{i-1}))(1+\sigma)
\end{equation}
where $E$ is the execution value, $M$ is the trading volume, $R_{i-1}$ is the remaining quantity after level $i$ in LOB, $\sigma$ is the commission fee rate, and $q_t^i,p_t^i$ are the level $i$ price and quantity in LOB respectively.  Furthermore, the executed price $p_e$ for this order is $\frac{E_t(Q)}{(\sum_i \Delta_{i})}$. This is exactly like spot trading. We can further compute the execution price as $E_P(Q)=\frac{E_t(Q)}{Q}$. Suppose this order is to open a new position. The open loss is calculated as $O_l=Q(E_P-M)$, a subtraction of the executed price and nominal position value for the new position. 
The unrealized PnL is calculated as the $U_p=Q(M_t-E_{po})$ while the realized PnL is calculated as $R_p=Q(E_{pc}-E_{po})$, where $E_{po}$ is the executed price for opening the position and $E_{pc}$ is the executed price for closing the position, $M_t$ is the mark price.

\noindent\textbf{Specific Trading Process}
Initially, we initialize our balance with limited capital, the current mark price, the initial position, and leverage at timestamp $t$. Then, we pose an action (a specific target position with a specific target leverage). We first judge whether this is an open new or closed position. For opening position orders, we change the current leverage to the target leverage before calculating the execution price described in the Market Order with the Leverage. For closing position orders, we first close the position before changing the order. Then, we calculate the margin balance and check whether it is smaller than the maintenance margin. Liquidation will happen if the margin balance is smaller than the maintenance margin. The market order for closing the position will happen, and we must pay a higher transaction cost for the market order due to the liquidation. If no liquidation happens, the order can be executed, and we wait for a certain period. Then, we recalculate the margin based on the latest mark price and judge if liquidation is triggered. If triggered, we terminate the current environment; otherwise, we calculate the margin balance difference caused by the mark price change and utilize that difference as the reward for the timestamp. Then, we iter this process until the $t-1$, where t is the length of the provided data.

\subsection{RL Environment Design}
Though there already has been work on trading with margin~\cite{gu2023margin}, our environment focuses on trading one single asset with a much more realistic data-driven simulation and more flexible in trading settings. 
Our environment is more flexible regarding leverage choices (up to 125 times) and contains more details about commission fees and funding costs, which is unique in trading perpetuals in Crypto. Since there is no delivery day for perpetual futures, Binance creates a mechanism called funding to force the price of the futures to align with the price of the spot. The funding mechanism is that the exchange will first calculate the difference between the mark price of the perpetual and the spot price and, based on the difference, derive a funding fee. If the mark price is higher than the spot price, then the funding fee is positive; if the mark price is lower than the spot price, then the funding fee is negative. For every 8 hours, the exchange charges the position holder a funding fee, which is the product of the funding rate and the nominal position value. The exchange itself does not profit from these transactions; rather, the funding fee acts as a mechanism to balance the interests of position holders, effectively facilitating cash exchanges to meet loan interest requirements. When the mark price of the future exceeds the spot price, indicating a shortage of cash on the exchange, the funding fee is positive. In this scenario, traders with long (positive) positions are charged the funding fee, distributed to those holding short (negative) positions. Conversely, when the mark price of the future is lower than the spot price, indicating a shortage of volatile cryptocurrencies, the funding fee becomes negative. In this case, traders with short positions are charged, and the collected funds are distributed to those with long positions.).

\noindent\textbf{State} $S_{t}$ consists of 3 parts: technical indicators $y_{t}$ and position $H_t$ and funding count down $f_{cd} $. $y_t$, consisting of around 300 features, is denoted as the market state representation or market state in the paper. Because of the funding mechanism to maintain the price of the spot and perpetual, perpetual trading demonstrates a periodic property. Some traders will focus on making money through the funding fee, which will cause market dynamic changes. Since our experiment is conducted on a minute level, the funding countdown consists of 2 parts: the hour and minute of the countdown. We integrate these 2-time dimensions into the input embedding and integrate them with position and technical indicators.
\noindent\textbf{Dynamic} $Z_t$ at time $t$ refers to the macro-economic fundamentals like interest or inflation rate changes, which could not be directly reflected in the market state representations. This changes both the state transition matrix and the market state representation distribution, although market state representation distribution might overlap under different market dynamics.  

\noindent\textbf{Reward} $r_t$ refers to the margin balance change. It is calculated as $r_{t}=H_t(M_{t+1}-M_t)-O_t$, consisting of capital loss due to market order and value change for the position. It is worth noticing that although the reward does not directly contain leverage, it is determined by your current position $H_t$, and given a limited capital, the upper bound of the position $H_{t_{up}}$ is determined by the leverage. So, in the actual implementation, we will notice that although our environment supports a list of leverage choices, all of the experiments are conducted under the leverage of 5 because the leverage itself does not matter; what matters is the maximum position a trader can hold.

\noindent\textbf{Action} $a_t$ at time $t$ refers to the target position with corresponding leverage choice. Both position and leverage are chosen from pre-defined pools. Position pool $P_p$ is a finite set, defined as $\{-H_{max},-H+\frac{2H_{max}}{|P_p|-1},\cdots,0, \frac{2H_{max}}{|P_p|-1},..., H_{max}\}$ where $|P_p|$ represents the number of the position choices and $H_{max}$ is the maximum position (here we demand the number of the position choices must be odd to contain the zero position situations) Leverage pool $P_l$ is also finite set, defined as $\{1, 1+\frac{L_{max}-1}{|P_l|-1},..., L_{max}\}$, where $L_{max}$ refers to the largest leverage and $|P_l|$ refers to the number of leverage choices. We encode these two dims (position, leverage) into just one action, with a total number of $|A|=|P_L|\times(|P_p|-1)+1$ (since no leverage when the position is 0) elements. We directly choose from these $|A|$ elements, where each element represents a specific position with a specific leverage.

\section{Method}
\subsection{Proof of Convergence}
\label{Appendx:sec:Proof}
Since our method is largely based on q learning, we provide theoretical proof that, under a DP-MDP, where there are $m$ MDPs whose state space and action space are identical and finite, an ensemble with m elements can converge to the optimal action-value ensemble, where each element converges to the optimal action-value $Q^*$ under a specific MDP. Here, we first define the distance between two ensembles given a finite group of transitions and show this definition fulfills the properties of distance. Then, we show that the optimal ensemble is a fixed point regarding our update operator. Finally, we show that the operator is a contraction.

Since we are dealing with ensemble RL, it is not evident how the distance between the two ensembles should be defined. Integrating the whole transition space, including multiple dynamics and the elements from the ensemble, is necessary. Given $m$ dynamics $D=(D_1,\cdots, D_m)$ (MDPs with different transition matrices and the same state space and action space), we construct the transitions for each dynamics $T_a=(T^1,\cdots, T^m)$, where $T^i=((s_0,a_0,r_0,s_1), \cdots)$, indicates the transitions in dynamics $i$. We further denote the td error for the transition $T^j$ $i^{th}$ element in the ensemble $Q_{e1}$ as 
$L_{Q_{e1}^i}(T^j)=\frac{\sum_{s,a,r,s' \in \text{dynamic $j$}}|r+\gamma \argmax_{a} {Q_{e1}}^i(s',\cdot)-{Q_{e1}}^i(s,a)|}{|s,a,r,s' \in \text{dynamic $j$}|}$ 
We define the distance between two ensembles for the $ith$ dynamic as 
\begin{equation}
    d_i(Q_{e1},Q_{e2})=\argmax_{s,a \in \text{Dynamic i}}|Q_{e1}^j(s,a)-Q_{e2}^k(s,a)|
\end{equation}
where 
\begin{equation}
j=\argmin_j \sum L_{Q_{e1}^j}(T^i)
\end{equation}
and 
\begin{equation}
k=\argmin_k \sum L_{Q_{e2}^k}(T^i)
\end{equation}
Further, we conclude the distance between two ensembles for all dynamics as 
\begin{equation}
    d(Q_{e1},Q_{e2})=\argmax_{i}d_i(Q_{e1},Q_{e2})
\end{equation}
In conclusion, given two ensembles, we first compute the average TD errors among each dynamic for each ensemble and assign the corresponding element with the smallest average TD errors in the ensemble to this dynamic. Then, we find the state action combination that maximizes the subtraction of the Q value from the chosen element in different ensembles. This means that for 2 ensembles under one dynamic, we can get a value to measure the distance between the two ensembles. We choose the dynamic that maximizes the value and define this value as the distance between the two ensembles under $m$ dynamics, denoted as  $d(Q_{e1}, Q_{e2})$. 

It is easy to validate that the distance between two same ensembles is 0 (since the two ensembles are identical, the element assigned into each dynamic is the same, and therefore, the distance under each dynamic is 0) and $d(Q_{e1}, Q_{e2})=d(Q_{e2}, Q_{e1})$. Now we validate the triangle rule triangle inequality: given three ensembles $Q_{e1}, Q_{e2}, Q_{e3}$, we can get $d(Q_{e1}, Q_{e3})\leq |d(Q_{e1}, Q_{e2})+d(Q_{e3}, Q_{e2})|$.
Here is the process to prove this:
\begin{equation}
\label{eq:tri_ineq}
    \begin{aligned}
    & d(Q_{e1}, Q_{e3})=\argmax_{i} d^i(Q_{e1}, Q_{e3})=\\& \argmax_{i} \argmax_{s,a \in \text{Dynamic i}} |Q_{e1}^j(s,a)-Q_{e3}^l(s,a)| \\&\leq |Q_{e1}^j(s,a)-Q_{e2}^m(s,a)|+|Q_{e2}^m(s,a)-Q_{e3}^l(s,a)| \\&\leq \argmax_{d_1} \argmax_{s,a\in \text{Dynamic $d_1$},}|Q_{e1}^{j}(s,a)-Q_{e2}^{m}(s,a)|+\\&\argmax_{d_2}\argmax_{s,a\in \text{Dynamic $d_2$}}|Q_{e2}^{m}(s,a)-Q_{e3}^{l}(s,a)\\&=|d(Q_{e1}, Q_{e2})+d(Q_{e2}, Q_{e3})|
    \end{aligned}
\end{equation}
As described in equation~\ref{eq:tri_ineq}. We first go through the definition and break it into scalar levels. Then, we apply the normal triangle inequality and magnify it by picking the dynamic, state, and action combination that maximizes each element to form the final inequality.

Assume that we have $m$ elements in our ensemble, $Q_e=(Q_e^1,\cdots, Q_e^m)$ and the ranking of average td error from $m$ dynamic for $m$ elements in the ensemble does not overlap, which means that for each dynamic $D_i$, the average td error of transitions from $D_i$ for element j in the ensemble is calculated as $L_{ij}=L_{Q_e^j}(T^i)$. The rank of the $L_{ij}$ across each $i$ for dynamic or each $j$ for the ensemble element does not overlap, which indicates that each element of the initial ensemble has the best q estimation across each dynamic.
We now define the operator H for transitions $(x,a,r,b)$ and aggregated transition matrix $T(y|x,a)$, we have
\begin{equation}
    \begin{aligned}
    \label{eq:def_operator}
    HQ_e(x,a)=(Q_e^1,\cdots,T(y|x,a)(r(x,a,y)+&\\ \gamma
    \argmax_b Q_e^i(y,b)),\cdots,Q_e^m)
    \end{aligned}
\end{equation}
where 
\begin{equation}
    \begin{aligned}
    \label{eq:def_index}
i=\argmin_{i}|T(y|x,a)(r(x,a,y)+ &\\ \gamma \argmax_b Q_e^i(y,b))-Q_e^i(x,a)|
    \end{aligned}
\end{equation}
In short, equation~\ref{eq:def_operator} and~\ref{eq:def_index} indicate in DP-MDP scenarios, we only update the index with the smallest average TD errors for given transitions. It helps the ensemble of the RL focus on one specific dynamic and helps the convergence of each element. 

Now, we show that the optimal ensemble $Q_e^*$ is a fixed point for operator $H$ for all transitions, where $Q_{e^*}=(Q_{e^*}^1,\cdots,Q_{e^*}^m)$ and $Q_{e^*}^i(x,a)=T(y|x,a)(r(x,a,y)+\gamma \argmax_b Q_{e^*}^i(y,b))$ for transitions from $i^{th}$ dynamic. 
Without loss of generality, Assuming given a bunch of transitions from $D_{i}$, we have
\begin{equation}
\begin{aligned}
HQ_{e^*}(x, a) =(Q_{e^*}^1,\cdots T_i(y|x,a)(r(x,a,y)+&\\\gamma \argmax_b Q_{e^*}^i(y,b)),\cdots, Q_{e^*}^m)&\\= (Q_{e^*}^1,\cdots,Q_{e^*}^m)=Q_{e^*}(x, a) 
\end{aligned}
\end{equation}
Since it holds for each $i$, the operator $H$ is a contraction of all the transitions from all the dynamics.

The averaged TD error determines the chosen index for updating, and the rest is the same with q-learning. So, we first prove that the update is a contraction under a single dynamic since no index change happens. 

Since the rank of the $L_{ij}$ across each $i$ for dynamic or each $j$ for the ensemble element does not overlap, we only need to prove that the averaged td errors do not decrease more on transitions from dynamic $j$ than on the transitions from dynamic $i$ if for trained index. ~\cite{ma2024discerning} provides proof of the monotonically decreasing property of TD error under certain assumptions. we denote $Q^0$ as the q value initial chosen index before any update and $Q^1$ as the q value that has been updated once. Without loss of generality, we denote the chosen index to be $i$ and the transition matrix as $T_i(y|x, a)$, $TD_{s_{ij}}$ represents the subtraction of the averaged td error reduction regarding dynamic $i$ and $j$,
where we require for any $\tau=(s, a,s')$, By the conclusion from ~\cite{ma2024discerning}, and the assumption that the TD error reduces more for transition with higher probability under one specific dynamic: which is for any $\tau_1,\tau_2$, we have $((T_i(\tau_1)-T_i(\tau_2))(L_i(\tau_1)-L_i(\tau_2)-(L_{i+1}(\tau_1)-L_{i+1}(\tau_2)))>0$, where$L_i$ refers to the td errors after $i^{th}$ update.
we also assume that $T_j(\tau)=T_i(\tau)+c(a-T_i(\tau))$ where $0 \leq c\leq 1$ and a is a constant to make sure the normalization is guaranteed.

$TD_{s_{ij}}$ is calculated as 
\begin{equation}
    \begin{aligned}
&TD_{s_{ij}}=\sum_{(s,a,s')}(T_i(s,a,s')|r+\gamma\argmax_{a'} Q^1(s',a')-Q^1(s,a)|\\&-T_j(s,a,s')|r+\gamma\argmax_{a'} Q^1(s',a')-Q^1(s,a)|-\\&(T_i(s,a,s')|r+\gamma\argmax_{a'} Q^0(s',a')-Q^0(s,a)|
 )-\\&T_j(s,a,s')|r+\gamma\argmax_{a'} Q^0(s',a')-Q^0(s,a)|)\\&=\sum_{(s,a,s')}((T_i(s,a,s')-T_j(s,a,s'))(|r+\gamma\argmax_{a'} Q^1(s',a')-\\&Q^1(s,a)|-|r+\gamma\argmax_{a'} Q^0(s',a')-Q^0(s,a)|)\\&=\sum_{(s,a,s')}((T_i(s,a,s')-T_j(s,a,s'))(L_1-L_0)\\&=\sum_{(s,a,s')}cT_i(s,a,s')(L_1-L_0)<0
 \end{aligned}
\end{equation}
Therefore, the TD loss reduction for different dynamics is not as large as that of the originally chosen index. Therefore, the chosen index remains the same, and the bellman operator is a contraction~\cite{jaakkola1993convergence}; therefore, for every index, it is a contraction, making it a global contraction on the whole ensemble, and its upper bond regarding the index is also a contraction. Therefore, we can say that our algorithm works in multi-dynamic environments.

In real deployment, we made 3 adjustments so that the algorithms could be better and more easily deployed in the real world. 1) since there is no clear division of the transitions from different dynamics for a long trajectory, it is somewhat difficult to ensure that the sampled transitions come from the same dynamic, so we make the experience replay buffer smaller. Since smaller experience replay buffers mean that the sampled transitions are made by timestamps close to each other, they have a better chance of sharing the same dynamic. 2) we utilize a selective update with a neighbor mechanism, forcing the indexes close to each other to be updated with similar transitions. This will help if the index number is larger than the number of the dynamics. 3) We add a demonstration value loss function from~\cite{qin2024earnhft} for faster convergence. 4) we pre-train each agent with the same transitions so that the assumption that the ranking of average td error from $m$ dynamic for $m$ elements in the ensemble does not overlap can be more easily fulfilled.

\section{Experiment}

\begin{table}[!b]
\setlength\tabcolsep{4pt}
  \begin{center}
  \caption{Quick maintenance margin calculation table.}
  \label{Appendix:tab:main_margin}
    \begin{tabular}{cccc}
    \toprule
    Perpetual& NPV Bound & k & j\\ 
    \midrule
     \multirow{5}{*}{BNBUSDT} &   10000 &0.005&0\\
     &  50000 &0.006& 10 \\
     &  100000 &0.01& 210 \\
     &  500000 &0.02& 1210 \\
     &  2000000 &0.05& 16210 \\
     \midrule
     \multirow{3}{*}{BTCUSDT} &   50000 &0.004&0\\
     &  500000 &0.005& 50 \\
     &  10000000 &0.01& 2550 \\
     \midrule
     \multirow{4}{*}{DOTUSDT} &   10000 &0.0065&0\\
     &  50000 &0.01&35 \\
     &  500000 &0.02& 535 \\
     &  2000000 &0.05& 15535 \\
     \midrule
     \multirow{3}{*}{ETHUSDT} &   50000 &0.004&0\\
     &  500000 &0.005& 50 \\
     &  10000000 &0.0065& 800 \\
     \bottomrule
    \end{tabular}
  \end{center}
\end{table}
\begin{table}[!t]
\setlength\tabcolsep{5pt}
\small
\centering
\renewcommand{\arraystretch}{1.2}
\caption{Chronological splits for each dataset. All dates are in YY/MM/DD format.}
\label{Appx:tab:dataset_train_valid_test_split}
\vspace{-0.3cm}
\begin{tabular}{lccc}
\toprule
\textbf{Dataset} & \textbf{Train Range} & \textbf{Valid Range} & \textbf{Test Range} \\
\midrule
BNB/USDT & 22/01/01 -- 22/12/31 & 23/01/01 -- 23/05/31 & 23/06/01 -- 24/01/01 \\
BTC/USDT & 22/01/01 -- 22/12/31 & 23/01/01 -- 23/05/31 & 23/06/01 -- 24/01/01 \\
DOT/USDT & 22/01/01 -- 22/12/31 & 23/01/01 -- 23/05/31 & 23/06/01 -- 24/01/01 \\
ETH/USDT & 22/01/01 -- 22/12/31 & 23/01/01 -- 23/05/31 & 23/06/01 -- 24/01/01 \\
\bottomrule
\end{tabular}
\vspace{-0.5cm}
\end{table}

\subsection{Dataset}
\label{Appendx:sec:Dataset}
We begin by characterizing each trading pair, followed by a detailed description of the dataset partitioning into training, validation, and test splits. We examine the temporal dynamics of the training set and conduct a comparative analysis of the validation and test sets. This analysis reveals the shortcomings of non-MoE approaches and substantiates our selection of specific time intervals to ensure robust backtesting. Subsequently, we provide a rationale for selecting the cryptocurrency market as the primary experimental environment. Finally, we investigate the generalisation capability of FineFT by presenting empirical evidence across a variety of market conditions.

Here, we first provide the training, validation and testing dataset split in Table~\ref{Appx:tab:dataset_train_valid_test_split}.
\begin{itemize}
    \item BNBUSDT perpetual is a special trading pair in Binance. As a native utility token for Binance, BNB's liquidity is not that high, but its trading condition, like the maintenance margin rate, benefits traders. The depth of the LOB might be just $\frac{1}{10}$ of BTCUSDT perpetual, and therefore, the price fluctuates severely and is therefore pursued by radical traders. Though the trading liquidity is not that good, the maintenance margin calculation setting is somewhat favorable in Binance because it is a Binance native coin, as shown in Table~\ref{Appendix:tab:main_margin} 
    
    \item BTCUSDT perpetual is one of the most famous trading pairs with the highest liquidity. It allows for leverage of up to 125 $\times$. The order book's depth is large, so the price tends to be less volatile. The maintenance margin rate is the second smallest as its liquidity is very good.

    \item DOTUSDT perpetual is one of the niche trading pairs with low liquidity. The order book is shallow, so the price is extremely volatile. The maintenance margin rate is the worst compared among these 4 perpetuals.

    \item ETHUSDT perpetual is one of the most famous trading pairs with the highest liquidity. It allows for leverage of up to 125 $\times$. The order book's depth is large, but the price also tends to be volatile. The maintenance margin rate is the smallest as many orders are being conducted.
\end{itemize}

\begin{figure}[ht]
    \centering
    \includegraphics[width=\linewidth]{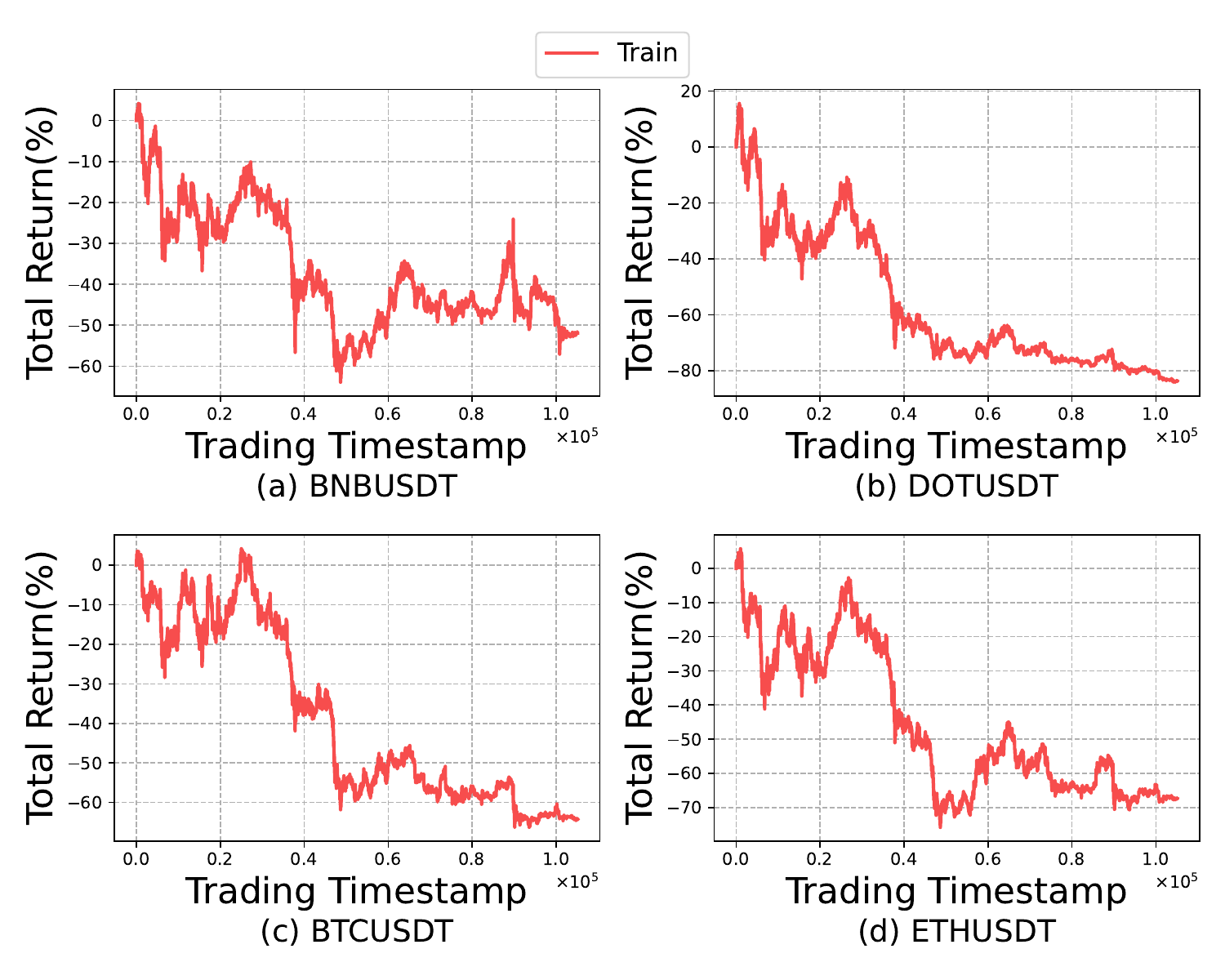}
    \caption{Training Dataset. The Y-axis is the buy \& hold return rate, and the X-axis is the timestamp.}
    \label{Appendix:fig:train_dataset}
\end{figure}

\begin{figure}[ht]
    \centering
    \includegraphics[width=\linewidth]{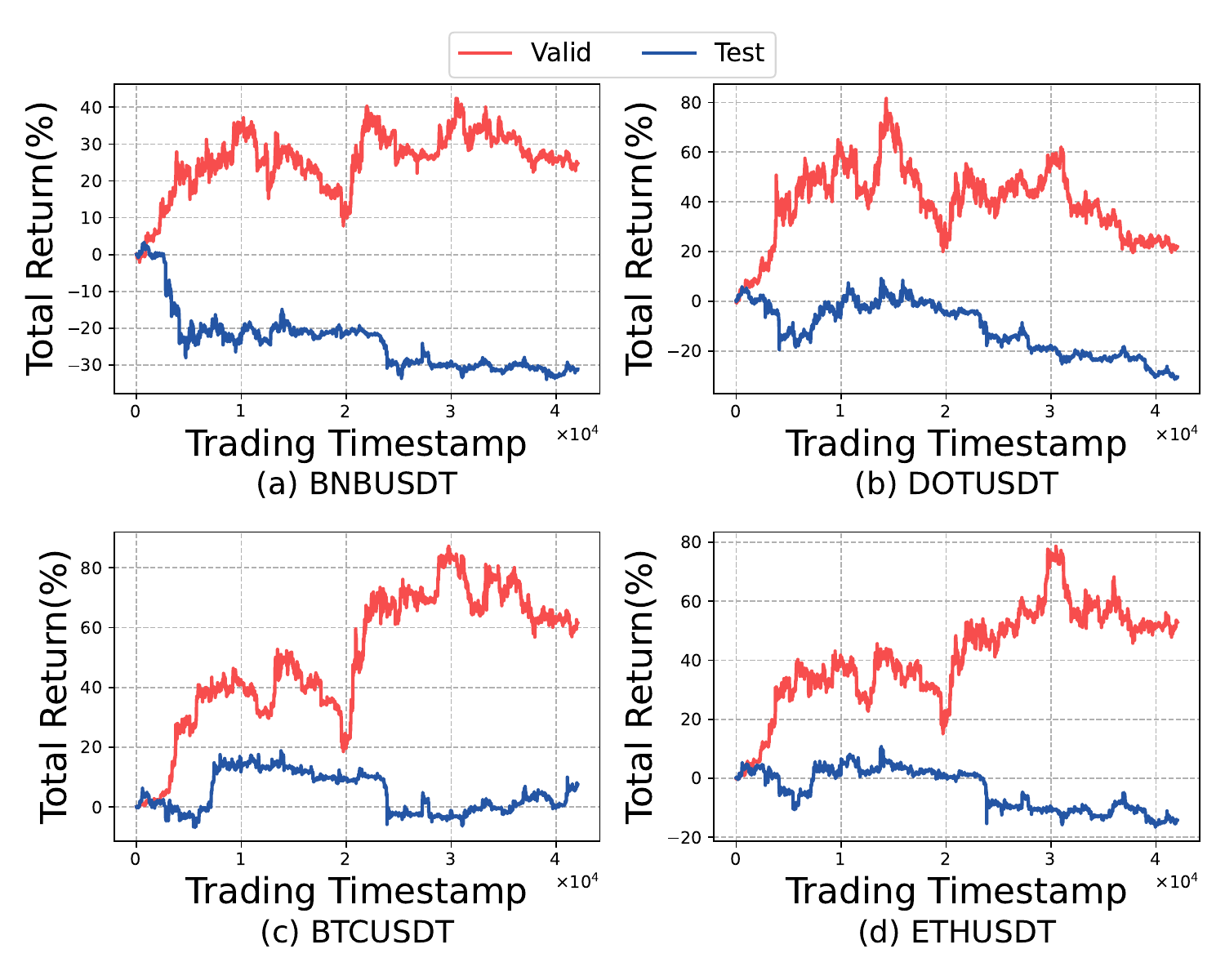}
    \caption{Valid \& Test Dataset. The Y-axis is the buy \& hold return rate, and the X-axis is the timestamp.}
    \label{Appendix:fig:valid_test_dataset}
\end{figure}

Here, we present the market trend for differnt trading pair in the training phase in Figure~\ref{Appendix:fig:train_dataset} and in the validation and testing phase in Figure~\ref{Appendix:fig:valid_test_dataset}. We observe that the training sets of most assets coincide with market drawdown phases, particularly during the aftermath of the mining ban and the latter half of the Bitcoin halving cycle, largely corresponding to the year 2022. This explains the pronounced negative returns in the training periods. In contrast, the validation sets generally exhibit higher volatility than the test sets, providing a more challenging environment for model selection. Both the validation and test periods contain asset pairs with upward and downward trends, ensuring that the full dataset spans diverse market regimes. This setup enables us to assess the robustness and adaptability of the agent across varying market conditions, thereby enhancing the comprehensiveness and realism of the backtesting evaluation.  Furthermore, each trading pair in our dataset comprises tens of thousands of intraday trading timestamps. This extensive granularity renders our backtesting results statistically more robust than conventional daily backtests, which typically contain only several hundred data points per year. Although our test period may appear relatively short in calendar time, the volume of high-frequency data substantially enhances the statistical significance of our evaluation. Our backtesting contains over 66 times more data points than a typical 3-year daily backtest, thereby offering significantly greater statistical power and reliability in performance evaluation. As a result, we do not report results over multiple random seeds. Although we do not report results across multiple random seeds on each test set, we instead leverage the high temporal resolution of our backtests to assess statistical significance. Specifically, we aggregate high-frequency returns into daily-level metrics—return, Sharpe ratio, and maximum drawdown—using 288 intraday points per trading day, which aligns with the typical granularity of daily financial data. Each trading day across all test datasets is treated as an independent evaluation instance, allowing us to compute statistical significance over a large number of downsampled yet meaningful observations. This approach provides a principled alternative to seed-based evaluation, enabling robust significance testing while preserving the fine-grained characteristics of high-frequency trading environments. The corresponding staticas is provided in Appendix~\ref{Appendx:sec:main_exp_statics}.

\noindent\textbf{Data Preoprocessing}. We collect high-frequency trading data from Tardis, including trades, quotes, derivatives tickers, and order book snapshots. All raw data streams are first downsampled to a uniform 5-minute frequency. We then construct base features that can be derived directly from single-timestamp observations, such as trade volume, mid-price, bid-ask spread, and implied volatility (where applicable).

Following this, we perform a timestamp-based join to merge features across the different data sources. The resulting feature vectors are aligned chronologically, sorted by timestamp, and concatenated to form the initial time-series feature matrix. We then compute rolling-window features over time (e.g., moving averages, realized volatility, and order flow imbalance) to incorporate short-term temporal dynamics.

For each target variable—defined as forward return or realized volatility over varying horizons (ranging from 5 minutes to 24 hours)—we train an XGBoost regression model to estimate its value from the constructed features. Feature importance scores derived from the trained models are used for initial feature filtering: any feature with an importance score above 0.01 for a given target is retained. Among the retained features, we compute pairwise Pearson correlation coefficients. For any pair of features whose absolute correlation exceeds 0.7, only the feature with the higher importance score is kept, ensuring redundancy is reduced while preserving predictive power.

This procedure yields a decorrelated feature subset for each target. We then combine all selected feature sets across targets, sorted in a priority order based on predictive relevance: (1) short-horizon return, (2) long-horizon return, (3) short-horizon volatility, and (4) long-horizon volatility. A second round of correlation filtering is applied across the combined set, using the same correlation threshold (0.7) and priority ordering to retain the most informative features. The final feature set is then used to construct the training data for each trading pair individually.

\noindent\textbf{Reason for Selecting Crypto and Discussion about the Generalization of FineFT}. This paper focuses on Crypto markets for the following reasons: 1) \textbf{Crypto is fit for technical analysis.} Crypto, as a kind of currency, lacks direct use value to its holders, compared with traditional markets like stock or certain foreign exchange (for stock, the holder can receive dividends whose value is based on the expectation and actual performance of the company; for foreign exchange, its holder can directly consume in the corresponding countries). This property makes it a perfect land for technical analysis, causing the proportion of retail investors in the crypto market to be almost the highest across all markets. Because the prices of things with the use value will fluctuate with the use value (even in an efficient market, the price of a company's stock will still fluctuate because of changes in dividends due to performance changes in the company itself). However, the cryptocurrency market seems to suffer less from this information flood (although it is true that some of them have already been taking up real services), causing most of the traders in Crypto to focus on technical analysis and somewhat make prices fluctuate more regularly.
2) \textbf{Crypto's data is easy to get.} This paper focuses on trading a single asset with \textbf{high leverages}. Though high leverage is tempting for radical traders, this mechanism puts the exchange or broker into a more risky situation (we talk about this in detail in Appendix C) and, therefore, is not common in the centralized market (for example, the leverage limit for most broker is 2 $\times$ for stock, yet in Binance, the leverage can go up to 125 $\times$.). Two common assets allow for high leverages: foreign exchange and Crypto. Yet, compared with crypto, historical fine-grained trading data for foreign exchange is extremely difficult to obtain. 
Therefore, This paper focuses on the Crypto. 

However, this does not mean the FineFT framework is limited to the scope of the crypto futures market with high leverage. Stochasticity, which corresponds to motivation 1 in this paper and is increased by the high leverage, is pervasive across all 
financial markets. Since the leverage is not that high in another market, the FineFT framework, though it can do better, may not be able to surpass other algorithms that much. The 
Unseen market representation, which corresponds to motivation 2 in this paper, is more common in other markets, such as the stock market, due to changes in underlying fundamentals like the release of earnings reports. 
It will cause a more frequent OOD triggering and, therefore, might need to do the rolling experiment more frequently. To further assess the generalization capability of FineFT across heterogeneous market environments, we perform additional experiments on corn futures and E-mini S\&P 500 (ES) futures. The results, presented in Appendix~\ref{Appendx:sec:more_dataset_results}, compare FineFT against DQN and a buy-and-hold baseline.

\subsection{Evaluation Metrics}
\label{Appendx:sec:evaluation_metrics}
\begin{itemize}
    \item \textbf{Total Return (TR)} is the overall return rate of the test period, defined as \( TR = \frac{V_{t}-V_{1}}{V_{1}} \), where \({V}_{t}\) is the final margin balance and \({V}_{1}\) is the initial margin balance.
    \item \textbf{Annual Volatility (AVOL)} is the variance of the annual return defined as \(\sigma[\mathbf{ret}] \times \sqrt{m}\) to measure the volatility risk, where \(\mathbf{ret}=(ret_1, ret_2,...,ret_t)\) is a vector of daily return, $\sigma[.]$ is the variances, and m is 365.
    \item \textbf{Maximum Drawdown (MDD)} measures the largest loss from any peak to show the downside risk of the strategy.
    \item \textbf{Annual Sharpe Ratio (ASR)} is the profit adjusted by volatility risk, defined as: \( SR =\frac{{E}[\mathbf{ret} ]}{\sigma[\mathbf{ret}]} \times \sqrt{m}\), where \(E[\mathbf{ret}]\) is the expectation of daily return.
    \item \textbf{Annual Calmar Ratio (ACR)} is defined as \(ACR = \frac{E[\mathbf{ret}]}{MDD} \times m \), measuring profit adjusted by downside risk.
    \item \textbf{Annual Sortino Ratio (ASoR)} applies downside deviation as the risk measure. It is defined as: \(ASoR = \frac{E[\mathbf{ret}]\times \sqrt{m}}{DD}\), where downside deviation is the standard deviation (DD) of the negative daily return rates.
    
\end{itemize}

\subsection{Baselines}
\label{Appendx:sec:baselines_discription}
\begin{figure}
    \centering
    \includegraphics[width=\linewidth]{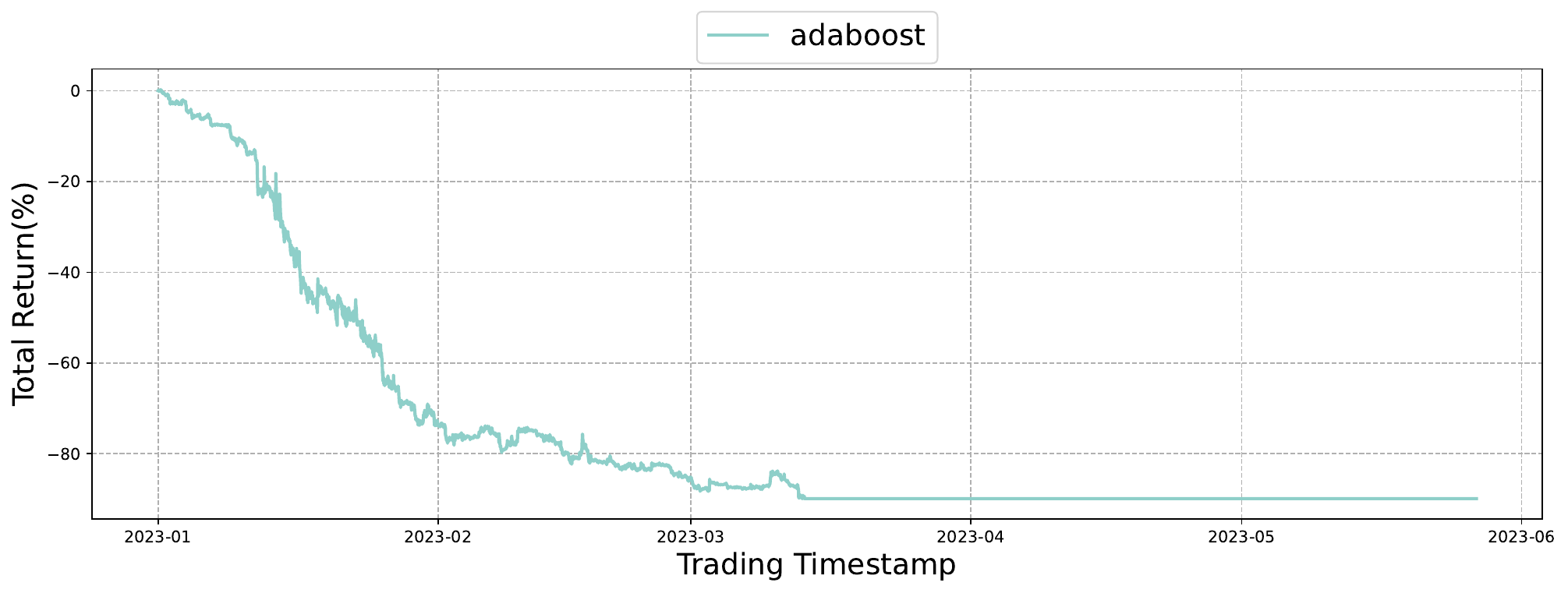}
    \caption{Example to show that the trading frequency of the supervised learning method is too high and therefore could not make profits in high-frequency trading.}
    \label{Appendix:fig:adaboost_not_work}
\end{figure}

Though not mentioned explicitly in this paper, our experiment is conducted under a minute-level setting. This frequency fails supervised learning because it only focuses on a short-term return. Without considering the influence of the current target position in the future, the agents tend to trade frequently, causing a tremendous loss in the commission fee (which is also magnified because of the high leverage). As shown in Figure~\ref{Appendix:fig:adaboost_not_work}, we use the result of AdaBoost to demonstrate the failure of supervised learning on such a task and justify the baseline choices.

\begin{itemize}
    \item \textbf{PPO} \cite{schulman2017proximal} utilizes importance sampling and generalized advantage estimation for RL.
    \item \textbf{DRA} \cite{briola2021deep} utilizes the LSTM for actor and critic to improve the state representation ability of PPO. 
    \item \textbf{DQN} \cite{mnih2015human} incorporates experience replay, multi-layer perceptrons, and soft backup to the target network to off-policy Q-learning for higher data efficiency. 
    \item \textbf{CRP} \cite{zhu2022quantitative} uses a random perturbed target frequency for a backup target network in DDQN.
    \item \textbf{IV} \cite{chordia2002order} reflects the imbalance between selling and buying pressure in LOB.
    \item \textbf{MACD} \cite{krug2022enforcing} utilize EMA from different time scales
    for profitable trading signals.
    \item \textbf{RAQR}~\cite{bernhard2019addressing} trains a QRDQN~\cite{dabney2018distributional} to handle the aleatoric uncertainty in the RL and utilizes a risk-averse policy to avoid the worst case. 
    \item \textbf{EDQN}~\cite{carta2021multi} utilizes different DQNs to handle different markets.
    \item \textbf{SUNRISE}~\cite{lee2021sunrise} views the deviation of learners as the aleatoric uncertainty and decreases the learning rate of the uncertain samples.
    \item \textbf{EHFT} \cite{qin2024earnhft} utilizes a high-level agent to select from an RL strategy pool trained on diverse trends.
    \item \textbf{MHFT} \cite{zong2024macrohft} is a hierarchical RL utilizing a Q-mix-like~\cite{rashid2020monotonic} network to route diverse agents trained on different trends and volatilities.
\end{itemize}

\begin{table*}[!th]
\centering
\renewcommand{\arraystretch}{1.2}
\caption{Training parameter setting. Here we use the soft update for the target network for value-based methods except for CDQNRP because the random perturbation of the frequency to hard update the target network is one of its contributions. }
\label{Appendix:tab:baselines_parameter_setting}
  \resizebox{0.95\textwidth}{!}{
  
    \begin{tabular}{lccccccccc|}
    \toprule
    \textbf{Algorithm} & \textbf{Buffer Size}  & \textbf{Batch Size}  & \textbf{lr}  & \textbf{$\alpha$} & \textbf{Window}&  \textbf{Entropy} &\textbf{$\tau$}&\textbf{$\beta$}&\textbf{Dynamic} \\ \midrule
    MacroHFT & $10^{6}$  & 512 & 2.5e-4 &128 & 864 &-&0.005&0.5&6\\
    EarnHFT & $10^{6}$  & 512 & 2.5e-4 &128 &- &-&0.005&-&5\\
    DDQN & $10^{6}$  & 512 & 2.5e-4 &-&-&-&0.005&-&- \\
    CDQNRP & $10^{6}$  & 512 & 2.5e-4 &- &3600&- &-&-&-\\
    PPO &512 & 64  & 1e-5 &-  &-&0.01&-&-&-\\
    DRA &512  & 64 & 1e-5 &-  &600&0.01&-&-&-\\ 
    \bottomrule
    \end{tabular}
}
\end{table*}

\subsection{Experiment Settings}
\label{Appendx:sec:experiment_setting}
For the trading setting, the commission fee rate is 0.02\% for all the datasets following Binance's policy. The maintenance margin is calculated as Equation~\ref{Appendix:eq:main_margine_calculation} and Table~\ref{Appendix:tab:main_margin}. For BNBUSDT, the maximum holding number is 100. For BTCUSDT, the maximum holding number is 8. For DOTUSDT, the maximum holding number is 6000. For ETHUSDT, the maximum holding number is 160. The position choice is 9 for trading pairs, indicating we split the position choice quarterly on both the long and short sides with a zero position. leverage choice is 5. The long estimated rate is 0.0005, and the short estimated rate is 0 to calculate open losses. The initial wallet balance is \$ 100000 with an empty position. All the trading settings follow the rules of Binance futures trading.

The replay buffer size is $1e6$.Adam is the optimizer for all the algorithms for training RL hyper-parameters. For FineFT, the number of hidden nodes for each independent DQN learner is 128. The ensemble size is 7, the neighbor range is 1, and the learning rate decays from $5e-3$ to $1e-4$ with $2e4$ steps. $\epsilon$ for $\epsilon$-greedy decays from 1 to 0.1 with $1e5$ steps and the soft update $\tau$ is 0.005. The $\alpha$ for optimal value supervisor decays from $256$ to $0$ with $5e5$ steps. The epoch number for pre-train is 2. For risk-aware heuristic routing, we utilize optuna~\cite{akiba2019optuna} to tune the parameter on the valid dataset before the inference on the test dataset. The parameters for the high-level agents are shown in Table~\ref{Appendix:tab:high_level_para}.
For other baselines hyper-parameters, we conclude them in Table~\ref{Appendix:tab:baselines_parameter_setting}, where $\beta$ indicates the memory loss coefficient, $\alpha$ is the optimal value supervisor coefficient, $\tau$ is the soft backup for DQN-based algorithms except CDQNRP, which utilises a random perpetuation backup. 
\begin{table*}[htb]
\setlength\tabcolsep{4pt}
  \begin{center}
\caption{Hyperparameter for risk-aware heuristic routing. $\gamma$, L, and $\tau$ refer to the decay factor for calculating the sliding window score, the length of the window and the risk-aware threshold.}
\label{Appendix:tab:high_level_para}
\begin{tabular}{cccccccccccccccc}
  \toprule
  Dataset & $\gamma$ & L &  $\tau$ & Dataset & $\gamma$ & L &  $\tau$ &  Dataset & $\gamma$ & L &  $\tau$ & Dataset & $\gamma$ & L &  $\tau$\\ 
  \midrule
  BNBUSDT & 0.94 & 134 & 0.37 & BTCUSDT  & 0.99 & 60 & 0.50 &DOTUSDT  & 0.97 & 63 & 0.76 & ETHUSDT & 0.99 & 240 & 0.50\\ 
  \bottomrule
\end{tabular}
\end{center}
\end{table*}

Under the specified parameter configuration, inference efficiency remains highly practical for real-time deployment. Leveraging PyTorch's vmap operation, joint inference involving five VAE modules and one DQN completes in approximately 0.78 milliseconds on an NVIDIA RTX 4090 GPU. This latency is only marginally higher than that of a standalone DQN model (0.28 milliseconds), indicating the computational overhead introduced by multiple VAEs is minimal. For comparison, a standard multi-layer perceptron (MLP) requires approximately 0.2 milliseconds per inference. Given that our trading policy operates on a 5-minute interval, all aforementioned latencies are orders of magnitude smaller than the decision frequency, and thus have a negligible impact on overall system responsiveness and real-time applicability.

\subsection{More Experiments Results}

\subsubsection{Results on More Datasets}
\label{Appendx:sec:more_dataset_results}
To further evaluate the generalisation capability of FineFT, we extend our experiments to two representative traditional financial futures: corn futures (as a commodity future) and ES futures (as an equity index future). These datasets, sampled at a daily frequency, exhibit distinct market dynamics compared to those originally used for training FineFT, thereby providing a rigorous test of its adaptability across heterogeneous financial environments. The dataset partitions for training, validation, and testing are detailed in Table~\ref{tab:dataset_splits_appendix}. These results are not included in the main paper, as the relatively limited number of data points (on the order of several thousand) is substantially smaller than that of the crypto datasets used in our main experiments (which contain nearly one million data points), and thus offers less statistical power. Nevertheless, we include these datasets to illustrate the generalizability of FineFT across diverse asset classes and market structures. Table~\ref{tab:performance_two_datasets} reports the performance of FineFT across key metrics, demonstrating its strong capability in achieving high returns while maintaining effective risk control.

\begin{table*}[!thb]
\centering
\caption{Train/Valid/Test splits for additional datasets. The dataset is downloaded via Yahoo Finance.}
\label{tab:dataset_splits_appendix}
\renewcommand{\arraystretch}{1.2}
\resizebox{0.95\textwidth}{!}{
\begin{tabular}{ccccc  ccccc}
\toprule
Dataset & Split & Start & End & Length & Dataset & Split & Start & End & Length \\
\midrule
\multirow{3}{*}{Corn}
& Train & 2001-06-01 & 2013-04-29 & 2994 
& \multirow{3}{*}{ES}
& Train & 2001-06-20 & 2013-03-11 & 2963 \\
& Valid & 2013-04-30 & 2018-01-31 & 1197
&       & Valid & 2013-03-12 & 2017-11-28 & 1185 \\
& Test  & 2018-02-01 & 2022-11-03 & 1197
&       & Test  & 2017-11-29 & 2022-08-19 & 1185 \\
\bottomrule
\end{tabular}
}
\end{table*}

\begin{table*}[!thb]
\centering
\caption{Performance comparison of 3 represented algorithms for ES and corn,\fst{pink} result shows the best performance.}
\label{tab:performance_two_datasets}
\renewcommand{\arraystretch}{1.2}
\resizebox{0.95\textwidth}{!}{
\begin{tabular}{cccccccccc}
\toprule
Dataset & Strategy & TR (\%) & SR & MDD (\%) & Dataset & Strategy & TR (\%) & SR & MDD (\%) \\
\midrule
\multirow{3}{*}{Corn} 
& Buy and Hold ×5 & 440.71 & 1.04 & 91.93 
& \multirow{3}{*}{ES} 
& Buy and Hold ×5 & 288.96 &  0.88 & 90.64 \\
\cmidrule(lr){2-5} \cmidrule(lr){7-10}
& DQN             & 503.17 & 1.27 & 47.37 
&                 & DQN             &316.30 & 1.18 & 35.31 \\
\cmidrule(lr){2-5} \cmidrule(lr){7-10}
& FineFT          &\fst{ 730.86} & \fst{1.74} & \fst{30.47} 
&                 & FineFT          & \fst{374.70}	 & \fst{1.27} & \fst{28.74} \\
\bottomrule
\end{tabular}
}
\end{table*}

\subsubsection{Significance Testing of Main Experiments}
\label{Appendx:sec:main_exp_statics}
As detailed in Appendix~\ref{Appendx:sec:Dataset}, we aggregate daily backtesting results across all test datasets to construct a unified set of evaluation instances for each performance metric. Specifically, for each calendar day, we compute the aggregated metric (e.g., mean return, Sharpe ratio, or maximum drawdown) by collecting the corresponding daily statistics from all trading pairs. This yields a series of daily evaluation points, each representing the model's performance on a specific day across multiple instruments.

To assess the statistical significance of our method’s advantage over baselines, we conduct pairwise comparisons using the Wilcoxon Signed-Rank Test. This non-parametric test evaluates whether the distribution of paired differences between our method and each baseline consistently favors one method over the other, without assuming normality. For a given metric, let ${x_i},y_i$ denote the aggregated daily values of our method and a baseline, respectively. The test computes the signed ranks of the non-zero differences 
$x_i-y_i$ and assesses whether the sum of positive ranks significantly exceeds (or falls below) that of the negatives. This enables robust evaluation of statistical superiority under distributional uncertainty and heavy-tailed characteristics typical of financial returns. The results are summarized in Table~\ref{Appendix:tab:wilcoxon_test_row}. We observe that for both return and Sharpe ratio, all p-values fall below the conventional 5\% significance threshold, indicating that our method consistently and significantly outperforms all baselines in these two metrics. Notably, for maximum drawdown, the p-values are exceedingly small—below 0.005\% in all pairwise comparisons—demonstrating that our approach achieves statistically superior drawdown control across the board. These results collectively validate the robustness and effectiveness of our method from multiple performance perspectives.

\begin{table}[h!]
\centering
\caption{Wilcoxon Signed-Rank Test p-values (\%) Across Metrics}
\label{Appendix:tab:wilcoxon_test_row}
\begin{tabular}{lccc}
\toprule
\textbf{Compared with} & \textbf{Return (>)} & \textbf{Sharpe (>)} & \textbf{MaxDrawdown (<)} \\
\midrule
EarnHFT         & 4.92  & 1.27  & 0.00 \\
MacroHFT        & 3.87  & 4.90  & 0.00 \\
AverageDQN      & 2.12  & 1.55  & 0.00 \\
CDQN-rp         & 0.73 & 1.88  & 0.00 \\
DQN             & 0.00 & 0.00 & 0.00 \\
DRA             & 1.53  & 1.57  & 0.00 \\
ImbalanceVolume & 1.38  & 0.82  & 0.00 \\
MACD            & 0.00 & 0.00 & 0.00 \\
NCQRDQN         & 3.98  & 3.84  & 0.00 \\
PPO             & 1.53  & 1.58  & 0.00 \\
SunriseDQN      & 0.02 & 0.90 & 0.00 \\
\bottomrule
\end{tabular}
\end{table}
\subsubsection{Behavior Analysis}
\label{Appendx:sec:behavior_result}
Besides the traditional financial metrics in the main paper, we deeply analyze the trading behaviour. Here are the definitions of these behavior-related metrics.
\begin{itemize}
    \item \textbf{Turnover (TO)} refers to the ratio between the number of shares traded and the maximum position. It measures how frequently the asset is traded. A higher turnover will result in a PnL further away from the buy-and-hold/sell-and-hold PnL for an asset, making the result more statistically significant and robust under a different dynamic. It is calculated as $TO=\frac{\sum_t |H_t-H_{t+1}|}{H_{max}}$, where $H_{max}$ refers to the maximum position a trader can hold and $H_i$ refers to the position holds at time $t$.
    \item \textbf{Total Trading Number (TTN)} refers to the number of times a trader closes its position. Together with TO, we can get a broad understanding of how radical a trader is. Once it opens a position, it tends to open a full position and just a half position. It is calculated as $TTN=\sum_t 1_{H_t \neq 0 \text{ and } H_{t+1} = 0}$. Through this, we can segment the overall return into returns for each single trade $rT_T=(rT_1, rT_2, \cdots, rT_{TTN})$, where the $rT_i$ refers to the profit made in the $i^{th}$ trade.
    \item \textbf{Trading Times (TT)} refers to the number of times a trader changes its position. It is calculated as $TT=\sum_t 1_{H_t \neq H_{t+1}}$.
    \item \textbf{Wining Rate (WR)} refers to the ratios between the number of profitable trades and the TTN, calculated as $WR = \frac{\left|\{rT_i \in rT_T \mid rT_i > 0\}\right|}{|rT_T|}$.
    \item \textbf{Reward-Risk Ratio (RRR)} refers to the ratio between the overall profit and the overall loss, calculated as $RRR=\frac{\sum_i\{ rT_i \in rT_T \mid rT_i > 0\}}{\sum_i\{ rT_i \in rT_T \mid rT_i < 0\}}$.
    \item \textbf{Averaged Reward-Risk Ratio (ARR)} refers to the ratio between the averaged profit and the averaged loss per trade, calculated as $ARR=\frac{ \frac{\sum_i\{ rT_i \in rT_T \mid rT_i > 0\}}{\left| \{ rT_i \in rT_T \mid rT_i > 0\}\right|}}{\frac{\sum_i\{ rT_i \in rT_T \mid rT_i < 0\}}{\left| \{ rT_i \in rT_T \mid rT_i < 0\}\right|}}$.
\end{itemize}

\begin{table*}[!thb]
\centering
\caption{Behavior analysis on 4 Crypto markets with 12 baselines  including 4 plain, 3 hierarchical, 2 ensemble, 1 distributional RL, and 2 rule-based methods. \fst{Pink}, \snd{green}, and \trd{blue} results show the \fst{best}, \snd{second-best}, and \trd{third-best} results.}
\label{Appendix:tab:behavior_tex}
\renewcommand{\arraystretch}{1.2}
  \resizebox{0.95\textwidth}{!}{
\begin{tabular}{cccccccccccccccc}
\toprule
\multicolumn{2}{ c  }{} & \multicolumn{3}{ c  }{Behavior} & \multicolumn{3}{ c }{Behavior-Adjusted Profit}  &\multicolumn{2}{ c  }{} & \multicolumn{3}{ c  }{Behavior} & \multicolumn{3}{ c  }{Behavior-Adjusted Profit} 
\\ 
\midrule
Market   & Model & TO$\uparrow$ & TTN$\uparrow$ & TT$\uparrow$ & WR(\%)$\uparrow$ & RRR$\uparrow$ & ARR$\uparrow$ &Market   & Model & TO$\uparrow$ & TTN$\uparrow$ & TT$\uparrow$ & WR(\%)$\uparrow$ & RRR(\%)$\uparrow$ & ARR(\%)$\uparrow$ \\
\midrule
{\multirow{13}{*}{\rotatebox[origin=c]{0}{BNB}}} & DRA & 0.50 & 1 & 1 & 0.00 &  0.00 &  0.00 &{\multirow{13}{*}{\rotatebox[origin=c]{0}{BTC}}} & DRA & 13.25 & 1 & 51 & \fst{100.00} &  -&  - \\
&  PPO  & 2.50 & 1 & 9 & 0.00 &  0.00 &  0.00& &  PPO  & 0.75 & 1 & 1 & \fst{100.00} &  -& -\\
\cmidrule(lr){2-8} \cmidrule(lr){10-16}

& CDQNRP & 0.75 & 1 & 1 & 0.00 &  0.00 &  0.00& &
CDQNRP &0.75 & 1 & 1 & \fst{100.00} &  - & - \\ 
& DQN &156.75 & 28 & 337 & 44.44 &  37.84 &  47.30&& DQN & \fst{4119.75} & 1 & \fst{16436} & 0.00 &  0.00 &  0.00\\ 
\cmidrule(lr){2-8} \cmidrule(lr){10-16}

& MACD &  \snd{867.75} & \snd{598} & \snd{1233} & 24.62 &  \trd{76.69} &  \fst{234.78}&& MACD & 317.00 & \snd{159} & 352 & 20.13 &  86.74 &  \fst{344.25}\\ 
& IV &1.00 & 0.00 & 2.00 & 0.00 &  0.00 &  0.00&& IV & \trd{331.50} & 64 & \trd{754} & 40.62 &  \trd{87.55} &  \trd{127.96}\\ 
\cmidrule(lr){2-8} \cmidrule(lr){10-16}
&EDQN & 0.25 & 1 & 1 & 0.00 &  0.00 &  0.00
&& EDQN &53.25 & 36 & 71 & 61.11 &  121.31 &  77.20 \\
&SUNRISE &136.00 & 1 & 541 & 0.00 &  0.00 &  0.00
&& SUNRISE &802.50 & 129 & 1135 & 46.51 &  94.91 &  109.15 \\
\cmidrule(lr){2-8} \cmidrule(lr){10-16}
&QRDQN &23.50 & 8 & 80 & 25.00 &  7.60 &  22.79
&& QRDQN &5.50 & 6 & 11 & 50.00 &  106.27 &  106.27\\
\cmidrule(lr){2-8} \cmidrule(lr){10-16}

&WINOW & 543.00 & 275 & 569 & 31.27 &  135.82 &  298.48
&& WINOW &1232.50 & 500 & 1694 & 24.40 &  85.55 &  265.06 \\
& EarnHFT & \fst{1874.50} & \fst{921.00} & \fst{1948.00} & \snd{51.90} &  \snd{78.76} &  \trd{72.99} && EarnHFT & \snd{784.50} & \fst{345} & \snd{1063} & \snd{49.57} &  \snd{108.44} &  110.34 \\
& MacroHFT & \trd{187.00} & 9 & 492 & \fst{55.56} &  35.48 &  28.38&& MacroHFT & 174.75 & 11 & 641 & \trd{45.45} &  64.70 &  77.64\\
\cmidrule(lr){2-8} \cmidrule(lr){10-16}
& FineFT & 183.50 & \trd{88.00} &\trd{256.00} & \trd{46.59} &  \fst{186.10} &  \snd{213.34} && FineFT & 190.00 & \trd{95} & 251 & 40.00 &  \fst{158.47} & \snd{ 233.53}\\
\bottomrule

{\multirow{13}{*}{\rotatebox[origin=c]{0}{DOT}}} &  DRA &3.50 & 1 & 13 & 0.00 &  0.00 &  0.00 &{\multirow{13}{*}{\rotatebox[origin=c]{0}{ETH}}} & DRA & 7.25 & 1 & 27 & 0.00 &  0.00 &  0.00\\
&  PPO  & 0.50 & 1 & 1 & 0.00 &  0.00 &  0.00&& PPO  & 2.25 & 1 & 7 & 0.00 &  0.00 &  0.00\\
\cmidrule(lr){2-8} \cmidrule(lr){10-16}
 & CDQNRP &\fst{503.50} & 1 & \fst{1143} & 0.00 &  0.00 &  0.00& & CDQNRP & 0.50 & 1 & 1 & 0.00 &  0.00 &  0.00 \\ 
& DQN & \trd{116.00} & \trd{31} & \trd{323} & \trd{54.84} &  \trd{58.58} & \trd{48.24}  && DQN & \fst{3482.75} & 1 & \fst{13609} & 0.00 &  0.00 &  0.00\\ 
\cmidrule(lr){2-8} \cmidrule(lr){10-16}

& MACD & \snd{245.00} & \fst{123} & 245 & 33.33 & \snd{107.09} &  \fst{214.17}& & MACD &  \snd{204.75} & \snd{162} &\trd{368} & 19.75 &  \trd{57.18} &  \fst{232.29}\\ 
& IV & 1.00 & 1 & 1 & 0.00 &  0.00 &  0.00&& IV &56.25 & 17 & 108 & \fst{52.94} &  52.40 &  46.58\\ 
\cmidrule(lr){2-8} \cmidrule(lr){10-16}
&EDQN & 0.25 & 1 & 1 & 0.00 &  0.00 &  0.00
&& EDQN &0.50 & 1 & 1 & 0.00 &  0.00 &  0.00 \\
&SUNRISE &99.00 & 19 & 263 & 42.11 &  50.92 &  70.02
&& SUNRISE &1073.50 & 497 & 3421 & 46.28 &  61.45 &  71.07\\
\cmidrule(lr){2-8} \cmidrule(lr){10-16}
&QRDQN & 16.00 & 17 & 33 & 17.65 &  5.71 &  24.73
&& QRDQN & 29.00 & 49 & 102 & 14.29 &  11.84 &  69.33\\
\cmidrule(lr){2-8} \cmidrule(lr){10-16}

&WINOW & 3379.00 & 1711 & 3517 & 36.47 &  72.96 &  127.09
&& WINOW &655.50 & 328 & 657 & 39.94 &  108.15 &  162.64\\
 & EarnHFT & 3.00 & 1 & 9 & \fst{100.00} &  - &  -&& EarnHFT & 193.00 & \trd{39} & 282 &\snd{ 51.28} &  \fst{208.34} &  \snd{197.92}\\
& MacroHFT  & 114.75 &\snd{38} & \snd{379} & \snd{73.68} &  \fst{536.23} &  \snd{191.51} && MacroHFT & \snd{426.50} & \fst{324} & \snd{1474} & 45.99 &  49.89 &  58.26\\
\cmidrule(lr){2-8} \cmidrule(lr){10-16}
& FineFT & 1.00 & 2 & 2 & 50.00 &  - &  -&& FineFT & 57.50 & 29 & 79 & \trd{48.28} &  \snd{183.22} &  \trd{183.22}\\
\bottomrule
\end{tabular}
}

\end{table*}

\begin{figure}[!th]
\centering
\vspace{-0.2cm}
\begin{subfigure}[t]{0.48\textwidth}
    \centering
    \includegraphics[width=\linewidth]{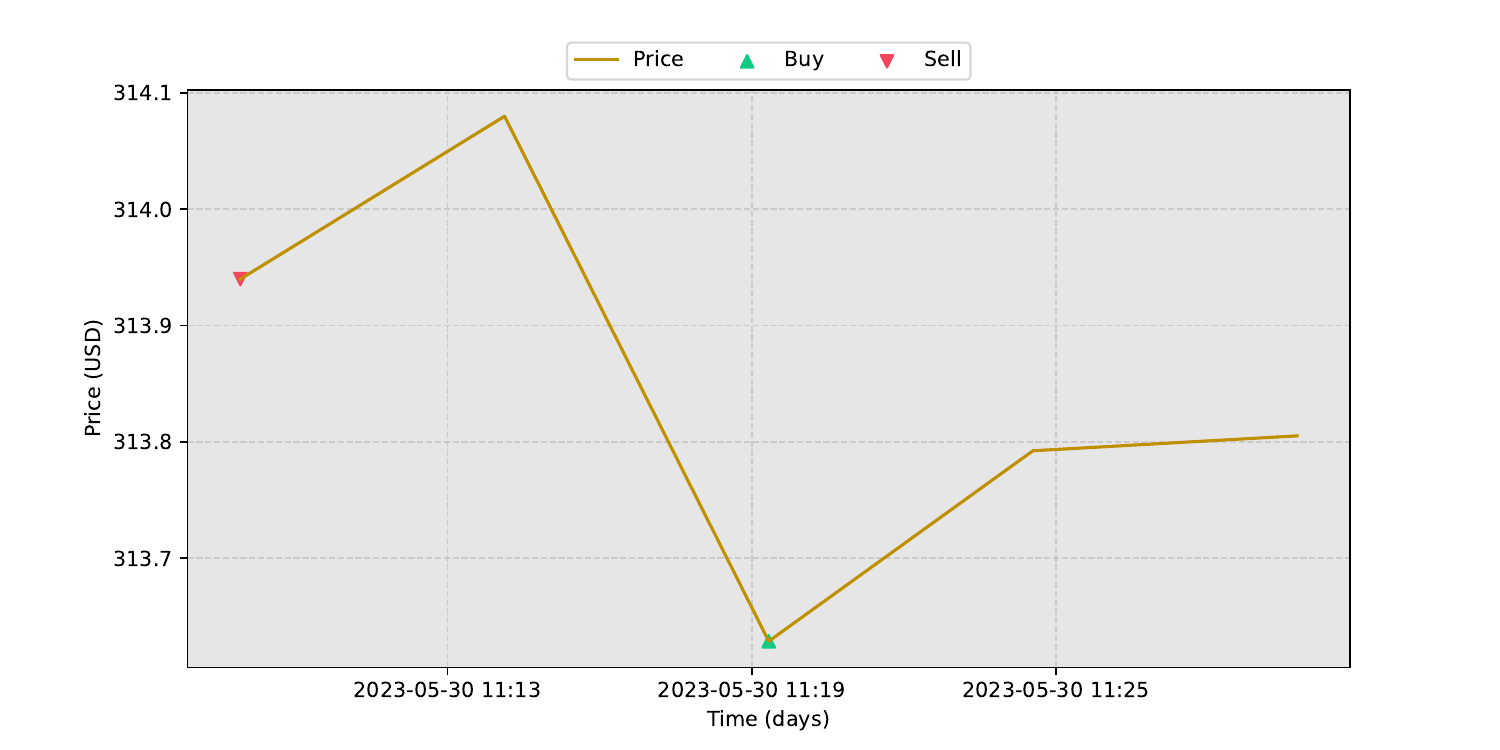}
    \caption{BNB trading example}
    \label{Appendix:fig:BNB_trade}
\end{subfigure}
\hfill
\begin{subfigure}[t]{0.48\textwidth}
    \centering
    \includegraphics[width=\linewidth]{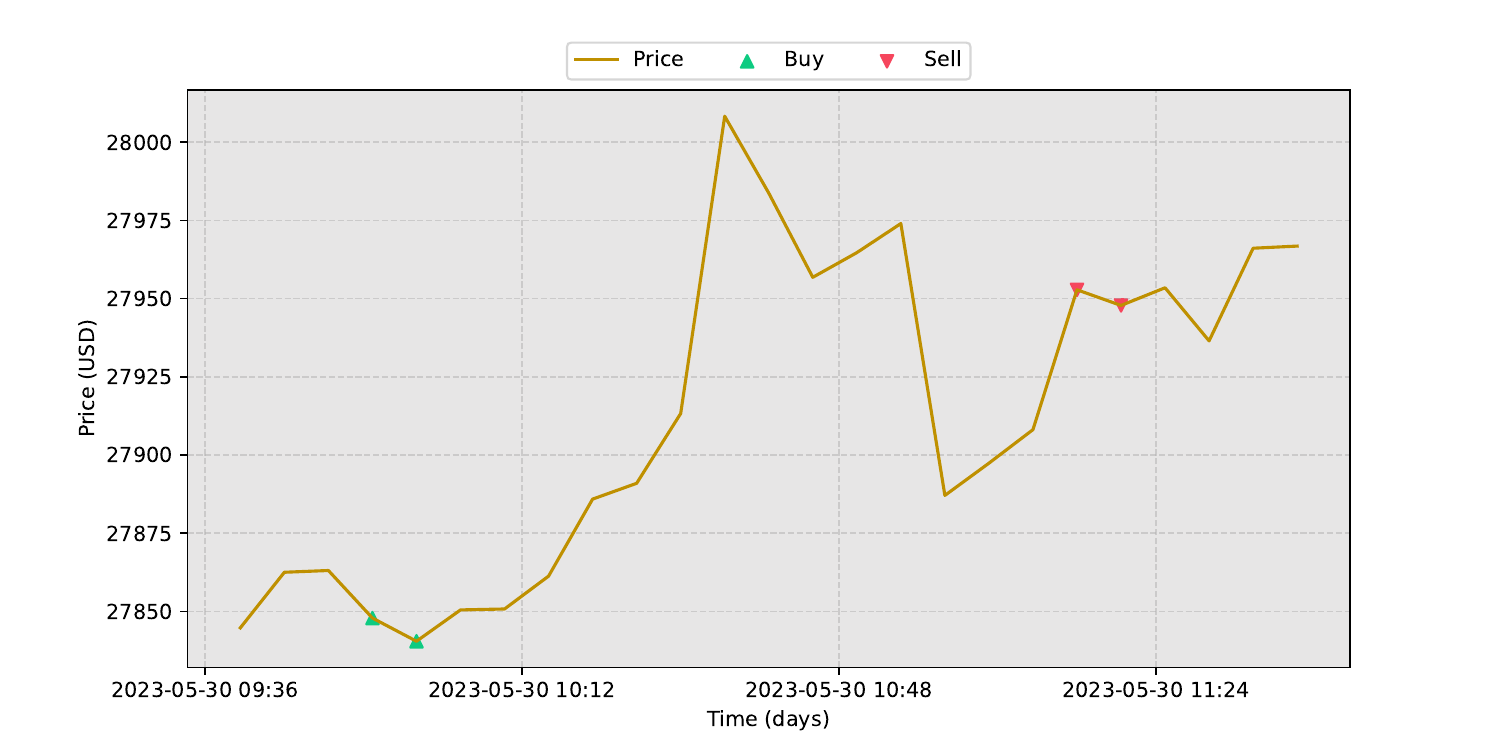}
    \caption{BTC trading example}
    \label{Appendix:fig:BTC_trade}
\end{subfigure}

\vspace{0.3cm}

\begin{subfigure}[t]{0.48\textwidth}
    \centering
    \includegraphics[width=\linewidth]{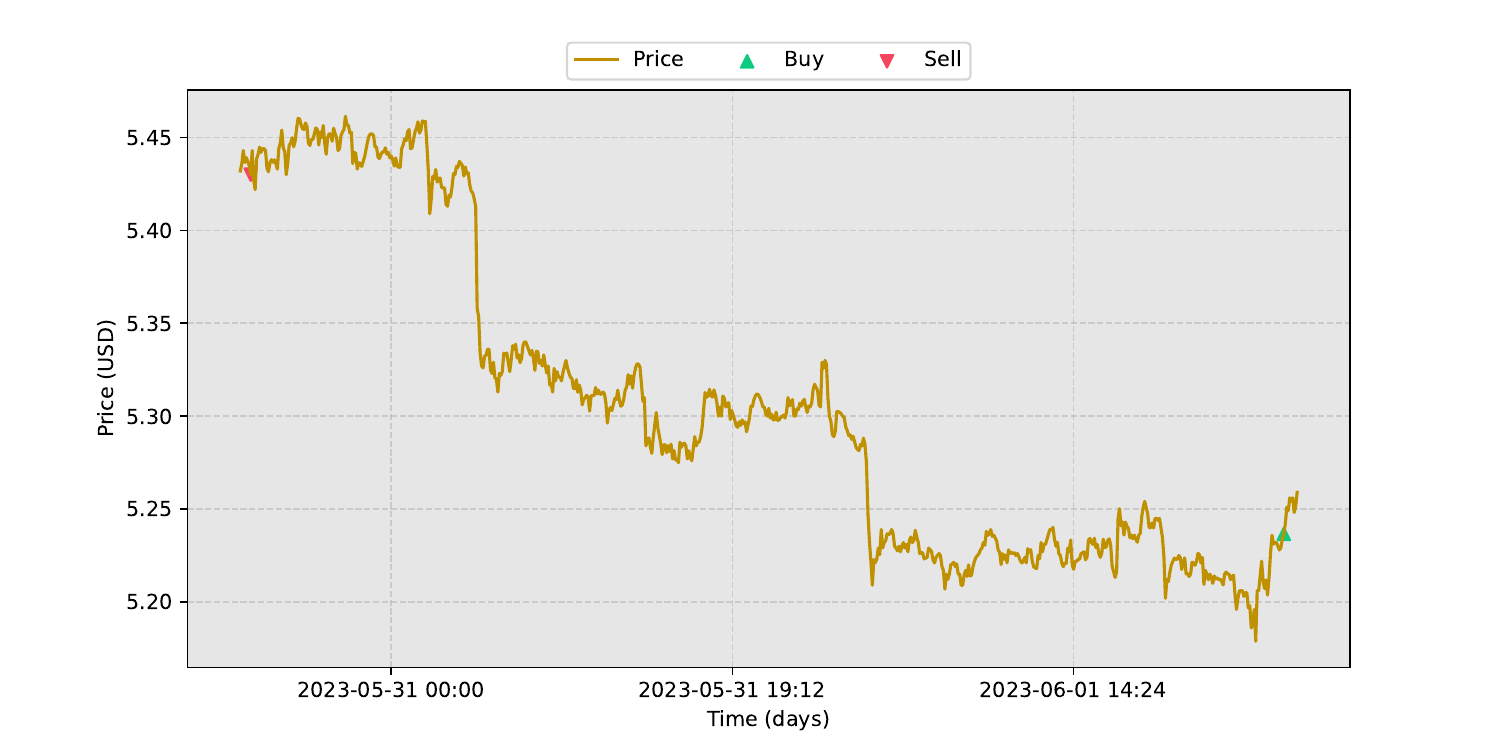}
    \caption{DOT trading example}
    \label{Appendix:fig:DOT_trade}
\end{subfigure}
\hfill
\begin{subfigure}[t]{0.48\textwidth}
    \centering
    \includegraphics[width=\linewidth]{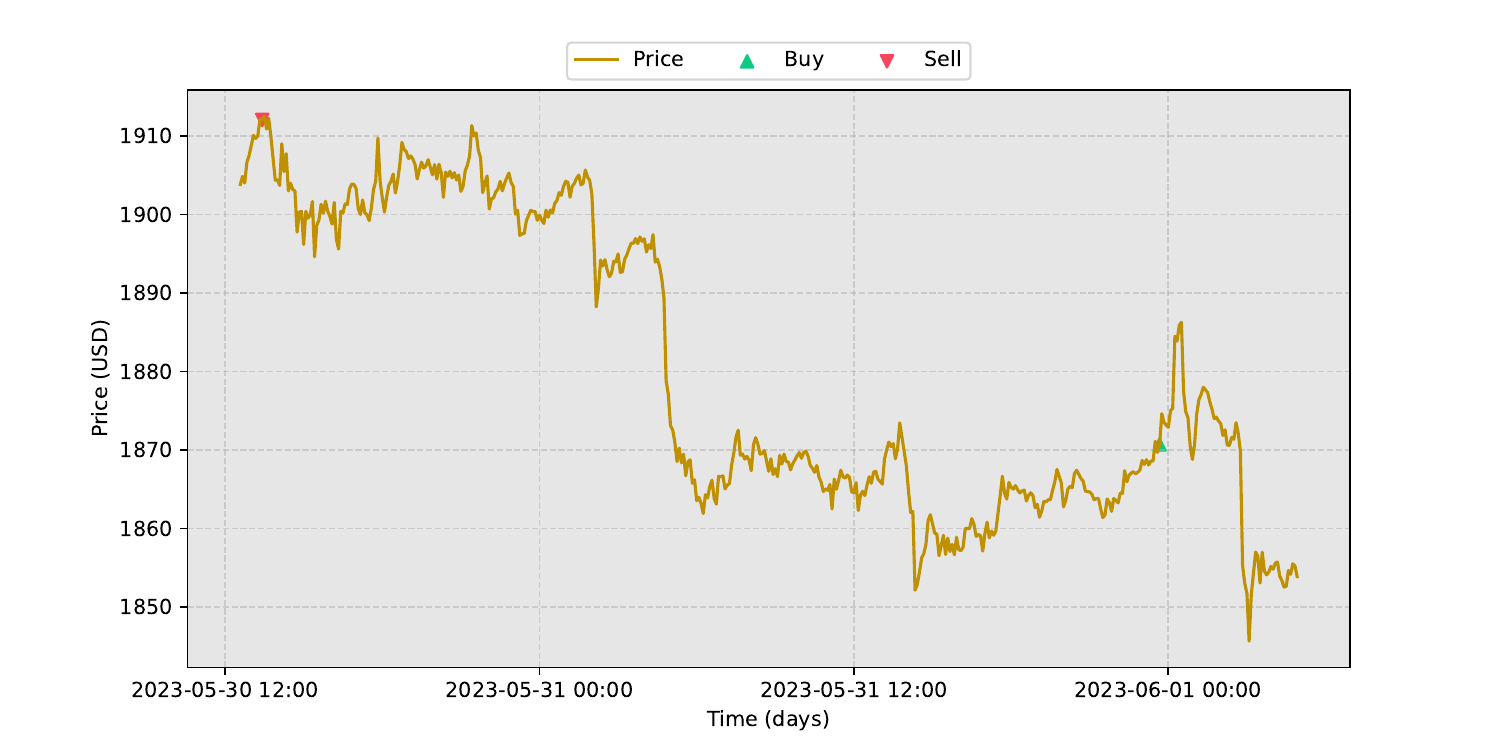}
    \caption{ETH trading example}
    \label{Appendix:fig:ETH_trade}
\end{subfigure}

\caption{Selected trading examples for multiple crypto assets.}
\label{Appendix:fig:good_trades_grid}
\end{figure}

As shown in Table~\ref{Appendix:tab:behavior_tex}, we can see that except for DOTUSDT, FineFT achieves stable behavior-adjusted profit like the risk-reward ratio is pretty stable. During the test phase of DOTUSDT, Polkadot announced its update on the Polkadot 2.0 plan, therefore changing the fundamentals of the perpetual. The caused market state representation is 
\begin{table}[htb]
\setlength\tabcolsep{4pt}
  \begin{center}
  \caption{Ablation Study of Selective Update with Neighbor for DOTUSDT. \fst{Pink}, \snd{green}, and \trd{blue} demonstrate the \fst{best}, \snd{second-best}, and \trd{third-best} results. Meth stands for method, and EP, ER refers to EarnHFT with and without prioritised episode selection, FP and FwoP mean FineFT's selective update with and without pretrain. Dyn means the corresponding market dynamics, and CS means the steps needed to converge, RS refers to the reward sum.}
  \label{Appendix:tab:ablation_SUN_DOT}
    \begin{tabular}{cccccccc}
    \toprule
    Meth& CS$\downarrow$ & Dyn & RS$\uparrow$&Meth& CS$\downarrow$ & Dyn & RS$\uparrow$ \\ 
    \midrule
     \multirow{5}{*}{EP} & \multirow{5}{*}{\snd{20736}} & 1 &\trd{1.48}&\multirow{5}{*}{ER} & \multirow{5}{*}{\fst{18144}} & 1 &\trd{1.48} \\
     &  & 2 &\trd{0.76}& & & 2 &\trd{0.76} \\
     & & 3 &\trd{0.12}& & & 3 &\trd{0.12} \\
     & & 4 &\trd{0.75}& & & 4 &\trd{0.75} \\
     &  & 5 &\trd{1.37}& & &5 &\trd{1.37} \\
     \midrule
     \multirow{5}{*}{FP} &  \multirow{5}{*}{85536} & 1 &\snd{13.71}&\multirow{5}{*}{FwoP} &  \multirow{5}{*}{\trd{74304}} & 1 &
     \fst{13.74} \\
     &  & 2 &\snd{6.97}& & & 2 &\fst{7.07} \\
     &  & 3 &\fst{2.45}& & & 3 &\snd{1.86} \\
     && 4 &\fst{7.16}& & & 4 &\snd{7.01} \\
     &  & 5 &\snd{12.35}& & & 5 & \fst{12.65} \\
     \bottomrule
    \end{tabular}
  \end{center}
  
\end{table}
We also provide more trading examples for each trading pair as shown in Figure~\ref{Appendix:fig:BNB_trade}, Figure~\ref{Appendix:fig:BTC_trade}, Figure~\ref{Appendix:fig:DOT_trade}, and Figure~\ref{Appendix:fig:ETH_trade}.

\begin{table}[htb]
\setlength\tabcolsep{4pt}
  \begin{center}
  \caption{Ablation Study of Selective Update with Neighbor for BNBUSDT. Pink, green, and blue demonstrate the best, second-best, and third-best results. Meth stands for method, and EP, ER refers to EarnHFT with and without prioritised episode selection, FP and FwoP mean FineFT's selective update with and without pretrain. Dyn means the corresponding market dynamics, and CS means the steps needed to converge, RS refers to the reward sum.}
  \label{Appendix:tab:ablation_SUN_BNB}
    \begin{tabular}{cccccccc}
    \toprule
    Meth& CS$\downarrow$ & Dyn & RS$\uparrow$&Meth& CS$\downarrow$ & Dyn & RS$\uparrow$ \\ 
    \midrule
     \multirow{5}{*}{EP} & \multirow{5}{*}{541728} & 1 &\fst{4.57}&\multirow{5}{*}{ER} & \multirow{5}{*}{\trd{283392}} & 1 &\trd{4.11} \\
     &  & 2 &\fst{3.57}& & & 2 &2.93 \\
     & & 3 &\fst{0.75}& & & 3 &\trd{0.26} \\
     & & 4 &\snd{3.25}& & & 4 &\trd{3.24} \\
     &  & 5 &\snd{5.21}& & &5 &\fst{6.56} \\
     \midrule
     \multirow{5}{*}{FP} &  \multirow{5}{*}{\fst{66528}} & 1 &4.09&\multirow{5}{*}{FwoP} &  \multirow{5}{*}{\snd{85536}} & 1 &
     \snd{4.21} \\
     &  & 2 &\snd{3.15}& & & 2 &\trd{3.10} \\
     &  & 3 &\snd{0.27}& & & 3 &\trd{0.26} \\
     && 4 &\fst{3.28}& & & 4 &3.23 \\
     &  & 5 &\trd{4.48}& & & 5 &4.42 \\
     \bottomrule
    \end{tabular}
  \end{center}
  
\end{table}

\subsubsection{Ablation Study Concerning Risk-Aware Heuristic Routing}
\label{Appendx:sec:ab_study_ra_routing}
\begin{figure}
    \centering
    \includegraphics[width=\linewidth]{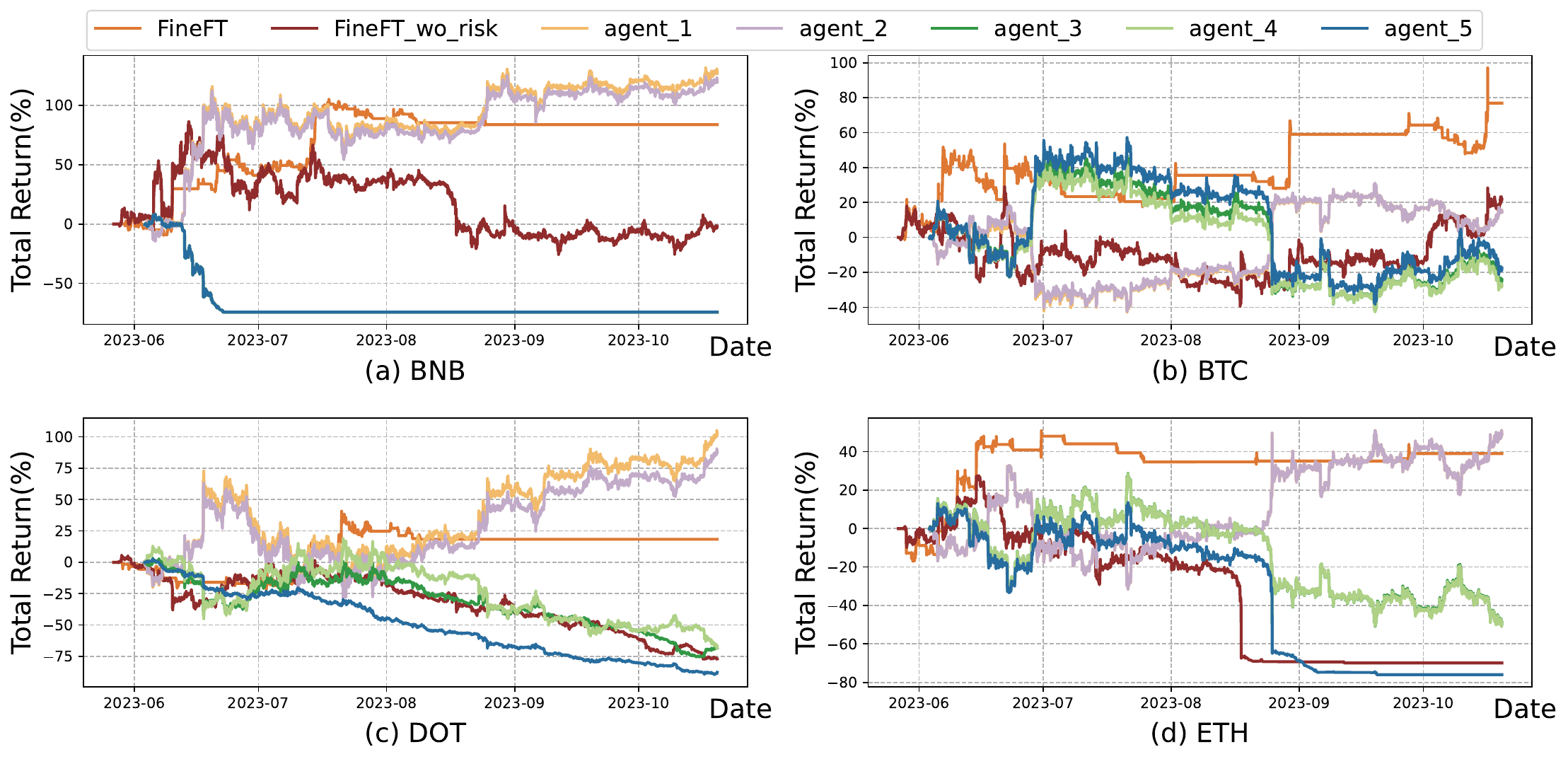}
    \caption{Performance comparison of different datasets}
    \label{Appendix:fig:ablation_routing_performance_comparision_all}
\end{figure}

Here, we demonstrate the performance comparison among the FineFT, FineFT without the risk threshold, which sets $\tau$ to zero, and each learner in the ensemble as shown in Figure~\ref{Appendix:fig:ablation_routing_performance_comparision_all}. We can see that in most cases, FineFT achieves the best or the second-best profitability, significantly reducing the maximum drawdown. FineFT without the risk threshold performs moderately compared with each learner in the ensemble. Notably, we observe that both FineFT and FineFT without a risk threshold initially perform the best during the trading process. However, when faced with unfamiliar market state representations, FineFT without a risk threshold tends to misjudge the current dynamic, leading to incorrect trading decisions and significant losses. We argue that the Mixture-of-Experts (MoE) framework is particularly suitable for complex financial modeling tasks, where the training dataset exhibits highly diverse and potentially conflicting market patterns that are difficult for a single agent to capture effectively. By distributing the representation burden across multiple specialized learners, MoE enhances the model’s expressivity and adaptability. However, this decomposition inevitably reduces the training data available to each expert, increasing the risk of overfitting and degrading generalization—especially under out-of-distribution (OOD) market states.

Our empirical results indicate that under such unseen states, the VAE-based uncertainty estimator tends to produce uniformly high losses across all experts, making their routing scores nearly indistinguishable. As a result, the gating network may oscillate frequently between experts. These switches, though algorithmically neutral, have real-world trading implications: they typically involve substantial portfolio rebalancing, leading to significant transaction costs. This explains why the variant without the risk threshold (i.e., $\tau=0$) sometimes underperforms even the best individual expert, due to excessive position shifts induced by unstable expert selection.

Furthermore, while FineFT generally outperforms single learners due to its ability to manage risk and uncertainty, there are scenarios where a single expert may achieve higher profitability. This occurs when the expert’s learned risk exposure happens to align well with the prevailing market regime—such as a bull- or bear-biased learner in a trend-consistent environment. In contrast, FineFT adopts a more conservative stance under high epistemic uncertainty, particularly in state representations with sparse training coverage. This design choice reflects a principled tradeoff: in a low-data regime (on the order of hundreds of thousands of samples), relying on aggressive generalisation is often detrimental, and limiting exposure to uncertain states is a necessary safeguard.

These observations jointly support our architectural decision: while the MoE framework introduces flexibility, its effectiveness under distribution shift hinges on the presence of a robust OOD-aware risk threshold. The $\tau$-based gating mechanism plays a critical role in stabilising expert selection and mitigating overreaction to ambiguous states, thereby improving robustness and reducing unnecessary trading frictions. Our ablation study experiment results (Figure~\ref{Appendix:fig:ablation_routing_performance_comparision_all}) validate this mechanism both quantitatively and qualitatively, demonstrating that the proposed risk-aware routing scheme achieves a better balance between adaptability and stability than any single learner or naive ensemble variant.
\subsubsection{Ablation Study Concerning Selective Update with Neighbor}
\label{Appendx:sec:ab_study_selective_update}
Here, we demonstrate the convergence steps and converged reward sum as listed in Table~\ref{Appendix:tab:ablation_SUN_BNB},~\ref{Appendix:tab:ablation_SUN_DOT},~\ref{Appendix:tab:ablation_SUN_ETH}. We can see that the FineFT with pre-training and FineFT without pre-training requires the least the steps to converge. The converged reward sum is the highest in the dataset ETHUSDT and DOTUSDT. For BNBUSDT, EP (EarnHFT with prioritised episode selection) achieves the best convergence result, yet it requires as much as 10 $\times$ data as the FP does, with a decrease of less than 10\% regarding the performance. So, in conclusion, FP requires the least number of steps to converge and achieves the best converged performance.

Compared to prior frameworks such as EarnHFT and direct MoE-style learning, our proposed Selective Update with Neighbors mechanism offers both faster convergence and better performance stability, particularly in environments with highly diverse market dynamics. We attribute this to several key differences in training principles and structural design.

First, in EarnHFT with prioritized episode sampling, transitions are sampled in a way that increases the likelihood of repeatedly observing episodes from the same market dynamic. While this enhances per-dynamic convergence by concentrating updates, it also isolates the learning of each dynamic. As a result, the overall convergence time becomes the sum of the convergence times across all market regimes, which grows linearly with the number of dynamics. In contrast, EarnHFT with random sampling lacks such specialization, leading to cross-dynamic contamination: transitions from one market regime may degrade performance under another, increasing learning variance and slowing convergence. This issue is particularly pronounced in settings where different dynamics demand mutually exclusive strategies, making the learner vulnerable to performance poisoning.

Second, FineFT without pretraining suffers from cold-start inefficiency, as each agent must individually acquire basic trading behavior without any shared prior. This significantly inflates the early training burden and increases the number of iterations required to reach competent performance, especially in low-signal or volatile markets.

Finally, unlike traditional Mixture-of-Experts architectures, which rely on a prior gating mechanism to select sub-networks for both inference and update (based on hard-coded state representations or classifier outputs), our Selective Update framework adopts a posterior-driven approach. Specifically, transitions are assigned to agents based on observed estimation error (i.e., agents with lower TD errors receive more weight), allowing the system to learn specialization organically from feedback. Once agents develop localized expertise through selective training, we then introduce a VAE-based uncertainty model to estimate each agent's domain of competence (i.e., its input prior). This two-stage process—posterior-driven specialisation followed by prior modelling—ensures stable convergence at each phase. In contrast, standard MoE methods couple inference and learning via fixed routing, which can lead to premature or unstable specialisation and hinder convergence due to noisy or inaccurate priors in the early stage.

Together, these insights explain the superior learning efficiency and robustness of our method. By encouraging focused and data-driven specialisation while minimising harmful cross-agent interference, Selective Update with Neighbours accelerates convergence and enables a more reliable division of labour among agents across varying market conditions.

\begin{table}[thb]
\setlength\tabcolsep{4pt}
  \begin{center}
   \caption{Ablation Study of Selective Update with Neighbor for ETHUSDT. Pink, green, and blue demonstrate the best, second-best, and third-best results. Meth stands for method, and EP, ER refers to EarnHFT with and without prioritised episode selection, FP and FwoP mean FineFT's selective update with and without pretrain. Dyn means the corresponding market dynamics, and CS means the steps needed to converge, RS refers to the reward sum.}
  \label{Appendix:tab:ablation_SUN_ETH}
    \begin{tabular}{cccccccc}
    \toprule
    Meth& CS$\downarrow$ & Dyn & RS$\uparrow$&Meth& CS$\downarrow$ & Dyn & RS$\uparrow$ \\ 
    \midrule
     \multirow{5}{*}{EP} & \multirow{5}{*}{\trd{76032}} & 1 &\trd{8.21}&\multirow{5}{*}{ER} & \multirow{5}{*}{\fst{25920}} & 1 &\trd{8.21} \\
     &  & 2 &\trd{4.65}& & & 2 &3.55 \\
     & & 3 &\trd{2.20}& & & 3 &2.10 \\
     & & 4 &\trd{13.62}& & & 4 &\trd{13.62} \\
     &  & 5 &\trd{17.01}& & &5 &\trd{17.01} \\
     \midrule
     \multirow{5}{*}{FP} &  \multirow{5}{*}{\snd{66528}} & 1 &\snd{47.55}&\multirow{5}{*}{FwoP} &  \multirow{5}{*}{86400} & 1 &
     \fst{48.91} \\
     &  & 2 &\fst{29.34}& & & 2 &\snd{26.62} \\
     &  & 3 &\fst{13.52}& & & 3 &\snd{8.68} \\
     && 4 & \snd{33.36}& & & 4 &\fst{35.48} \\
     &  & 5 &\fst{71.05}& & & 5 &\snd{48.72} \\
     \bottomrule
    \end{tabular}
  \end{center}
 
\end{table}

\begin{table}[!th]
\centering
\renewcommand{\arraystretch}{1.2}
\caption{Ablation of different numbers of agents and stop loss (drawdown threshold (DDT)).}
\label{tab:ablation_agents_stoploss}
\resizebox{0.5\textwidth}{!}{
\begin{tabular}{lcccc}
\toprule
\textbf{\#Agent} & \textbf{DDT} & \textbf{TR (\%)} & \textbf{SR} & \textbf{MDD (\%)} \\ \midrule
3 & 5\%  & 52.74 & 1.25 & 39.10 \\
5 & 5\%  & 76.90 & 2.13 & 23.51 \\
7 & 5\%  & 59.63 & 1.46 & 37.50 \\
5 & 1\%  & 53.82 & 1.38 & 22.96 \\
5 & 10\% & 71.32 & 1.64 & 41.85 \\
\bottomrule
\end{tabular}
}
\end{table}

\subsubsection{Number of agents and Drawdown setting}
\label{Appendx:sec:hyper_influnce}
Based on the results in Table~\ref{tab:ablation_agents_stoploss}, we observe that the model exhibits robust performance across a range of settings for the number of agents and drawdown thresholds.

In terms of the number of agents, performance remains strong for 3, 5, and 7 agents, with 5 agents yielding the best trade-off between diversification and stability. Fewer agents (e.g., 3) increase model confidence due to more data per agent but may underperform slightly due to limited specialization. More agents (e.g., 7) provide finer specialization but reduce per-agent data, slightly affecting stability. However, the results across these configurations show that the overall performance does not degrade drastically, indicating resilience to this hyperparameter.

As for the drawdown threshold, varying it between 1\%, 5\%, and 10\% results in predictable changes in the risk-return profile—lower thresholds reduce drawdown but also lower return, while higher thresholds increase return with more risk. The default 5\% setting demonstrates a good balance, but the consistent behavior across thresholds shows that the policy design remains effective and stable under different risk preferences.

Notably, the stop-loss threshold can be interpreted as a user-defined risk preference parameter, akin to the risk-aversion coefficient in classical mean-variance optimization~\cite{Markowitz1952}. The number of market dynamics further reflects a trade-off between routing complexity and agent specialization—more dynamics yield finer-grained specialization at the expense of data sparsity and potentially diminished VAE training efficacy, which may impact out-of-distribution detection performance.

These ablations confirm that our method is not overly sensitive to these hyperparameters, and the chosen configuration (5 agents, 5\% drawdown) reflects a well-validated, robust setting that generalizes well in practice.

\end{document}